\def\isarxiv{1} 

\ifdefined\isarxiv
\documentclass[11pt]{article}

\usepackage[numbers]{natbib}

\else

\documentclass{article} 
\usepackage{microtype} 
\usepackage{graphicx}
\usepackage{subfig}
\usepackage{hyperref}
\usepackage{icml2024}
\fi

\usepackage{amsmath}
\usepackage{amsthm}
\usepackage{amssymb}
\usepackage{algorithm}
\usepackage{algpseudocode}
\usepackage{grffile}
\usepackage{wrapfig,epsfig}
\usepackage{url}
\usepackage{xcolor}
\usepackage{epstopdf}

\usepackage{bbm}
\usepackage{dsfont}

\allowdisplaybreaks

\ifdefined\isarxiv

\usepackage{tikz}
\usepackage{hyperref}  
\hypersetup{colorlinks=true,citecolor=blue,linkcolor=blue} 
\usetikzlibrary{arrows}
\usepackage[margin=1in]{geometry}

\else

\usepackage{microtype}
\usepackage{hyperref}
\definecolor{mydarkblue}{rgb}{0,0.08,0.45}
\hypersetup{colorlinks=true, citecolor=mydarkblue,linkcolor=mydarkblue}

\fi

\theoremstyle{plain}
\newtheorem{theorem}{Theorem}[section]
\newtheorem{lemma}[theorem]{Lemma}
\newtheorem{definition}[theorem]{Definition}

\newtheorem{fact}[theorem]{Fact}

\newtheorem{claim}[theorem]{Claim}

\newcommand{\wh}{\widehat}

\newcommand{\R}{\mathbb{R}}

\renewcommand{\d}{\mathrm{d}}

\newcommand{\Tmat}{{\cal T}_{\mathrm{mat}}}

\DeclareMathOperator*{\Z}{\mathbb{Z}}

\DeclareMathOperator{\poly}{poly}

\DeclareMathOperator{\nnz}{nnz}

\DeclareMathOperator{\diag}{diag}

\DeclareMathOperator{\reg}{reg}

\DeclareMathOperator*{\argmin}{arg\,min}

\makeatletter
\newcommand*{\RN}[1]{\expandafter\@slowromancap\romannumeral #1@}
\makeatother

\usepackage{lineno}

\ifdefined\isarxiv

\else

\usepackage[textsize=tiny]{todonotes}

\icmltitlerunning{How to Inverting the Leverage Score Distribution?}

\fi

\begin{document}

\ifdefined\isarxiv

\date{}

\title{How to Inverting the Leverage Score Distribution?}
\author{
Zhihang Li\thanks{\texttt{lizhihangdll@gmail.com}. Huazhong Agricultural University.}
\and
Zhao Song\thanks{\texttt{zsong@adobe.com}. Adobe Research.}
\and
Weixin Wang\thanks{\texttt{wwang176@jh.edu}. JHU.}
\and 
Junze Yin\thanks{\texttt{junze@bu.edu}. Boston University.}
\and 
Zheng Yu\thanks{\texttt{yz388620@alibaba-inc.com}. Alibaba Inc.}
}

\else

\twocolumn[
\icmltitle{How to Inverting Leverage Score Distribution?}



\icmlsetsymbol{equal}{*}

\begin{icmlauthorlist}
\icmlauthor{Firstname1 Lastname1}{equal,yyy}
\icmlauthor{Firstname2 Lastname2}{equal,yyy,comp}
\icmlauthor{Firstname3 Lastname3}{comp}
\icmlauthor{Firstname4 Lastname4}{sch}
\icmlauthor{Firstname5 Lastname5}{yyy}
\icmlauthor{Firstname6 Lastname6}{sch,yyy,comp}
\icmlauthor{Firstname7 Lastname7}{comp}
\icmlauthor{Firstname8 Lastname8}{sch}
\icmlauthor{Firstname8 Lastname8}{yyy,comp}
\end{icmlauthorlist}

\icmlaffiliation{yyy}{Department of XXX, University of YYY, Location, Country}
\icmlaffiliation{comp}{Company Name, Location, Country}
\icmlaffiliation{sch}{School of ZZZ, Institute of WWW, Location, Country}

\icmlcorrespondingauthor{Firstname1 Lastname1}{first1.last1@xxx.edu}
\icmlcorrespondingauthor{Firstname2 Lastname2}{first2.last2@www.uk}

\icmlkeywords{Machine Learning, ICML}

\vskip 0.3in
]



\printAffiliationsAndNotice{\icmlEqualContribution} 

\fi

\ifdefined\isarxiv
\begin{titlepage}
  \maketitle
  \begin{abstract}

Leverage score is a fundamental problem in machine learning and theoretical computer science. It has extensive applications in regression analysis, randomized algorithms, and neural network inversion. Despite leverage scores are widely used as a tool, in this paper, we study a novel problem, namely the inverting leverage score problem. We analyze to invert the leverage score distributions back to recover model parameters. Specifically, given a leverage score $\sigma \in \mathbb{R}^n$, the matrix $A \in \mathbb{R}^{n \times d}$, and the vector $b \in \mathbb{R}^n$, we analyze the non-convex optimization problem of finding $x \in \mathbb{R}^d$ to minimize
\begin{align*}
    \| \diag( \sigma ) - I_n \circ (A(x) (A(x)^\top A(x) )^{-1} A(x)^\top )  \|_F
\end{align*}
where $A(x):= S(x)^{-1} A \in \mathbb{R}^{n \times d} $, $S(x) := \diag(s(x)) \in \mathbb{R}^{n \times n}$ and $s(x) : = A x -b \in \mathbb{R}^n$. Our theoretical studies include computing the gradient and Hessian, demonstrating that the Hessian matrix is positive definite and Lipschitz, and constructing first-order and second-order algorithms to solve this regression problem. Our work combines iterative shrinking and the induction hypothesis to ensure global convergence rates for the Newton method, as well as the properties of Lipschitz and strong convexity to guarantee the performance of gradient descent. This important study on inverting statistical leverage opens up numerous new applications in interpretation, data recovery, and security.

  \end{abstract}
  \thispagestyle{empty}
\end{titlepage}

{\hypersetup{linkcolor=black}
\tableofcontents
}
\newpage

\else

\begin{abstract}

Leverage score is a fundamental problem in machine learning and theoretical computer science. It has extensive applications in regression analysis, randomized algorithms, and neural network inversion. Despite leverage scores are widely used as a tool, in this paper, we study a novel problem, namely the inverting leverage score problem. We analyze to invert the leverage score distributions back to recover model parameters. Specifically, given a leverage score $\sigma \in \mathbb{R}^n$, the matrix $A \in \mathbb{R}^{n \times d}$, and the vector $b \in \mathbb{R}^n$, we analyze the non-convex optimization problem of finding $x \in \mathbb{R}^d$ to minimize
\begin{align*}
    \| \diag( \sigma ) - I_n \circ (A(x) (A(x)^\top A(x) )^{-1} A(x)^\top )  \|_F
\end{align*}
where $A(x):= S(x)^{-1} A \in \mathbb{R}^{n \times d} $, $S(x) := \diag(s(x)) \in \mathbb{R}^{n \times n}$ and $s(x) : = A x -b \in \mathbb{R}^n$. Our theoretical studies include computing the gradient and Hessian, demonstrating that the Hessian matrix is positive definite and Lipschitz, and constructing first-order and second-order algorithms to solve this regression problem. Our work combines iterative shrinking and the induction hypothesis to ensure global convergence rates for the Newton method, as well as the properties of Lipschitz and strong convexity to guarantee the performance of gradient descent. This important study on inverting statistical leverage opens up numerous new applications in interpretation, data recovery, and security.

\end{abstract}

\fi

\section{Introduction}

Leverage scores are an important concept in statistics and machine learning, with applications in areas like regression analysis \cite{cy21,akk+20,m18}, randomized matrix algorithms \cite{m11,dmmw12}, and more recently neural network inversion and adversarial attacks \cite{jrm+99,lkn99,zjp+20}. While leverage scores have been studied extensively and utilized in areas like randomized numerical linear algebra, prior work has not considered the problem of inverting or reversing a leverage score distribution back to the parameters that generated it. Most existing techniques simply use leverage scores as a tool for matrix approximations or improving algorithm efficiency. However, the ability to reconstruct parameters from leverage scores has significant implications. The novel concept explored in our work is leveraging the one-to-one mapping between parameters $x$ and leverage scores $\sigma$ to recover the hidden $x$ given only $\sigma$, which has not been shown before.
\begin{definition}[Inverting leverage score problem]\label{def:inverting_leverage_score}
    Given the matrix $A \in \R^{n \times d}$, the vector $b \in \R^n$, and the leverage score $\sigma \in \R^n$, we define $s(x) : = A x -b \in \R^n$, $S(x) := \diag(s(x)) \in \R^{n \times n}$, and $A(x):= S(x)^{-1} A \in \R^{n \times d} $.
    
    The goal of the inverting leverage score problem is to find the vector $x \in \R^d$, which satisfies
    \begin{align*}
        \min_{x \in \R^d } \| \diag( \sigma ) - I_n \circ (A(x) (A(x)^\top A(x) )^{-1} A(x)^\top )  \|_F.
    \end{align*}

\end{definition}

The importance of studying this inverting leverage score problem not only lies in gaining more insight into understanding the theoretical model interpretation but also helps with training data recovery and adversarial security. In some cases \cite{cle+19,zlh19}, sensitive private data used to train models could potentially be reconstructed from publicly available model information like leverage scores. Our inversion techniques make this feasibility analysis possible.
Regarding adversarial Security, systems relying on leverage score sampling may be vulnerable to attacks if adversaries can figure out how to accurately invert leverage scores back to model parameters. This could enable evasion or data theft. Proactively analyzing these risks is crucial.

\paragraph{Challenge and Contribution.}

Although the importance of studying this inverting leverage score problem is discussed, this leverage score inversion presents significant challenges. The problem is highly non-convex, with the leverage score mapping $\sigma$ being complex and nonlinear. Additionally, the high dimensionality of typical parameters $x$ poses scaling difficulties. We propose and analyze both first-order and second-order algorithms for tackling the inversion.

To tackle these obstacles, we make several theoretical contributions. First, we formulate and analyze this novel inversion task, providing gradient and Hessian computations. Our analysis decomposes the dense Hessian matrix into simpler constituent components mediated by the leverage score operator $\sigma(x)$. Next, we establish important optimization properties, including the positive definiteness and Lipschitz continuity of the Hessian matrix. These properties help to support the design of iterative methods.
Finally, leveraging the Hessian properties, we develop and analyze both first-order (gradient descent) and second-order (Newton method) algorithms for recovering parameters from leverage scores.

Our convergence rate analyses couple the strong convexity and smoothness guarantees from the positive definiteness and Lipschitz continuity of the Hessian matrix. For gradient descent, we bound the optimization error's decrease over iterations. For the Newton method, we utilize quadratic convergence rates inside shrinking neighborhoods surrounding the optima. Together, these results provide a rigorous foundation for effectively inverting statistical leverage, opening numerous applications, like training data recovery and adversarial security as discussed earlier.

\paragraph{Our Result.}

Our main result for studying the inverting leverage score problem (see Definition~\ref{def:inverting_leverage_score}) is as follows:

\begin{theorem}[Main result, informal version of Theorem~\ref{thm:gradient_formal}]\label{thm:gradient_informal}
    Let $L: \R^d \to \R$ denote the loss function. Let $n$ and $d$ be defined as in Definition~\ref{def:inverting_leverage_score} and $R > 0$. Let $\epsilon \in (0, 0.1)$. 
    
    Then, there exists a gradient descent algorithm (see Algorithm~\ref{alg:first_order}) that uses $O(\epsilon^{-1} \poly(nd) \exp(R^2))$ iterations, and each iteration takes $O(n^2d + d^{\omega})$ time to find $|L(x_T) - L(x^*)| \leq \epsilon$, where $x^*$ denotes the optimal solution for this problem and $x_T$ denotes the $T$-th iteration of the gradient descent.
\end{theorem}
Here $\omega$ is the exponent of matrix multiplication. Currently, $\omega \approx 2.37$ \cite{w12,lg14,aw21,dwz23,lg23,wxxz23}.

\begin{theorem}[Main result, informal version of Theorem~\ref{thm:newton_formal}]\label{thm:newton_informal}
    Let $L: \R^d \to \R$, $n$, $d$, $\epsilon$, and $R$ be defined identically as in Theorem~\ref{thm:gradient_informal}. 
    
    Then, there exists a newton-type greedy algorithm (see Algorithm~\ref{alg:second_order}) that uses $O(\log (1/\epsilon))$ iterations, and each iterations takes $O( n \cdot  ( \Tmat(d, n, n) + \Tmat(d, n, d)) + d^{\omega} )$ time to find $\| x_T - x^* \|_2 \leq \epsilon$, where $x^*$ denotes the optimal solution for this problem and $x_T$ denotes the $T$-th iteration of the Newton method.
\end{theorem}

Our first-order gradient descent method (Theorem~\ref{thm:gradient_informal}) has a low per-iteration cost but requires more total iterations to converge, while our second-order Newton method (Theorem~\ref{thm:newton_informal}) converges extremely quickly in fewer iterations due to quadratic convergence rates but has higher per-iteration complexity for the Hessian computation. Therefore, there is an inherent trade-off--gradient descent has simpler, cheaper iterations but needs more of them to reach accuracy $\epsilon$ compared to Newton, which has costlier iterations with faster convergence. Finally, our gradient descent algorithm returns $L(x_T)$ which is close to $L(x^*)$ and the Newton method returns an approximation of $\argmin_{x \in \R^d} L(x)$.

\paragraph{Notations.}

We let $[n] := \{1, 2, 3, \dots, n\}$. $\circ$ is a binary operation called the Hadamard product: $x \circ y \in \R^d$ is defined as $(x \circ y)_i := x_i \cdot y_i$. Also, we have $x^{\circ 2} = x \circ x$. For all $p \in \Z_+$, we define the $\ell_p$ norm of the vector $x$, denoted as $\|x\|_p$ to be equal to $\sqrt[p]{\sum_{i = 1}^d |x_i|^p}$. ${\bf 1}_n$ is the $n$-dimensional vector whose entries are all ones. $e_k$ is a vector whose $k$-th entry equals 1 and other entries are 0. When dealing with iterations, we use $x_t$ to denote the $t$-th iteration. In this paper, we only use the letters $t$ and $T$ for expressing iterations. $\langle x, y \rangle$ represents the inner product of the vectors $x$ and $y$. We define $\diag : \R^d \to \R^{d \times d}$ as $\diag(x)_{i, i} := x_i$ and $\diag(x)_{i, j} := 0$, for all $i \neq j$. We define $(A_{i, *})^\top \in \R^d$ to be the $i$-th row of $A$, and define $A_{*, j} \in \R^n$ to be the $j$-th column of $A$. We define the spectral norm and the Frobenius norm of $A$ as $\|A\| := \max_{x \in \R^d} \|Ax\|_2 / \|x\|_2$ with $\|x\|_2 \neq 0$ and $\|A\|_F := \sqrt{\sum_{i = 1}^{n}\sum_{j = 1}^{d}| A_{i, j} |^2}$, respectively. We use $x^*$ to denote the exact solution. 
$\nabla L$ and $\nabla^2 L$ denote the gradient and Hessian respectively. $\mathcal{T}_{\mathrm{mat}}(n, d, d)$ represents the running time of multiplying a $n \times d$ matrix with a $d \times d$ matrix. $\nnz(A)$ represents the number of non-zero entries of the matrix $A$.

\paragraph{Roadmap.}

In Section~\ref{sec:related_work}, we present the related work. In Section~\ref{sec:technique}, we give an overview of the techniques we use. In Section~\ref{sec:problem}, we formulate our inverting leverage score problem: we transform it into a simpler form and add a regularization term to it. In Section~\ref{sec:gradient_hessian}, we present our result of the gradient and Hessian computation. In Section~\ref{sec:property}, we analyze the properties of the Hessian matrix. In Section~\ref{sec:order_1}, we present the properties of the first-order method. In Section~\ref{sec:newton}, we introduce the properties of the Newton method. 
In Section~\ref{sec:conclusion}, we present our conclusion. 

\section{Related Work}
\label{sec:related_work}

\paragraph{Leverage Score}

Leverage scores have been studied in statistics for analyzing the linear regression models. The concept of statistical leverage refers to the influence of individual data points on the overall result in a linear regression \cite{ch86}. These ideas were brought into the field of numerical linear algebra and randomized matrix algorithms by the foundational work of \cite{dkm06,dmm06}. They showed that sampling matrix rows and columns according to their leverage scores allows efficient approximate solutions to problems like regression and fast matrix multiplication.  

Since then, leverage scores have been widely applied in randomized numerical linear algebra algorithms, such as computing approximate matrix factorizations, like CUR decomposition \cite{md09,swz17,swz19} and tensor CURT decomposition \cite{swz19}. To achieve this goal, rows and columns are selected based on their leverage scores, with probabilities corresponding to those scores \cite{dmm08}. Additionally, leverage scores can be applied to generate approximation algorithms for column subset selection \cite{bmd09,gs12,bdm14}, the design of deterministic algorithms \cite{pkb14}, the greedy algorithm \cite{abf+16,cm12,fgk11}. More recently, leverage scores have been applied to more areas, such as kernel methods \cite{ss02}, approximate factorizations and sampling methods \cite{ltos19,emm20,am15,clv17,mm17,lhc+20}, weighted low rank approximation \cite{syyz23}, matrix completion problem \cite{gsyz23}, the quantum algorithm for solving the linear regression, multiple regression, and ridge regression problems \cite{syz23}.

\paragraph{Inverting Problem}

Neural network inversion is an emerging research area with applications in model interpretation, adversarial attacks, and data recovery. Many recent works have studied different methods to invert specific types of neural networks \cite{cwd+18,db16,bwz+21,mv15,jrm+99,lkn99,zjp+20}.

\cite{mv15} introduced one of the first methods to invert CNN image classifiers to recover an approximation of the original input image. Their approach uses gradient-based optimization along with a natural image before finding an image that minimizes the loss between the target class score and the model's predicted class score. After that, \cite{db16,cwd+18} built upon this idea to improve the reconstruction quality for deeper networks and GAN generators respectively. \cite{jrm+99} studies different methods for reversing neural networks to find inputs that generate specific outputs. It shows how inversion can be useful for applications like query-based learning, analyzing sonar systems, assessing power grid security, and creating codebook vectors. \cite{lkn99} introduces a way to invert the trained networks by treating it as a mathematical optimization problem. \cite{zjp+20} proposes a new inversion attack technique that can successfully reconstruct faces from deep neural network models. It studies the link between a model's prediction accuracy and its vulnerability to such attacks.

\paragraph{Attack Problem}

With the growing integration of machine learning into security-critical systems, understanding model vulnerabilities to adversarial attacks has become imperative. Multiple attack strategies have been developed to fool neural networks at test time by introducing small perturbations to the input \cite{szs13}. For image classifiers, these tactics include methods such as the fast gradient sign method (FGSM) \cite{gss14} and the projected gradient attack (PGD) \cite{mms+17}, which modify pixel values to cause misclassification. The idea behind FGSM is to compute the gradient of the loss function with respect to the input data and then add or subtract a small perturbation to the input data in the direction that maximizes the loss, whereas PGD involves multiple iterations. It starts with an initial input and gradually updates it in the direction that maximizes the loss while staying within a bounded region. Defending against such attacks remains an open challenge. Adversarial training augments the training data with adversarial examples to increase robustness \cite{zyj+19}. However, this can degrade performance on clean images over time. Detection-based approaches have also been proposed, training models to identify adversarial inputs or using statistical tests to flag anomalies \cite{gmp+17,fcsg17}. However, maintaining high accuracy under adaptive attacks has proven difficult.

Protecting privacy in the face of increasingly sophisticated data analysis techniques is a critical challenge. Many machine learning models can inadvertently reveal sensitive personal information contained in the training data \cite{fjr15}. For example, membership inference attacks determine whether a given record was present in the model's training set \cite{ssss17}. Differential privacy has emerged as a principled technique to prevent such privacy violations by carefully calibrating noise into the learning algorithm \cite{dr14}. Specific approaches include output perturbation and objective perturbation \cite{lllw21}. However, substantial reductions in utility often accompany the privacy gains from differential privacy \cite{je19}.

\paragraph{Second-Order Method}

After the Newton method was introduced, it became one of the most fundamental and widely used root-finding algorithms due to its simplicity and quadratic convergence rate. There have been many theoretical modifications and enhancements being made to improve its performance. For example, \cite{b65} extends it to multiple dimensions for systems of nonlinear equations, \cite{nw99} develops various convergence acceleration techniques like line searches and trust regions to improve convergence issues for poor initial guesses, and more recently \cite{a00, jkl+20, bpsw21, szz21, hjs+22, lsz23} study the approximate newton method, namely using the approximate Hessian in the Newton method instead of the exact computation of Hessian.

Moreover, the second-order method has been utilized to solve a wide range of convex and non-convex optimization problems, including  
streaming algorithms \cite{syz23_sdp,bs23,lsz+23}, 
cutting plane methods \cite{jlsw20,lsw15}, 
attention optimization problem \cite{gsy23,gswy23}
linear programming \cite{sy21,gs22,jswz21,hlz23,b20,cls19}, 
semidefinite programming \cite{gs22,hjs+22,jkl+20,syz23_sdp}, 
empirical risk minimization \cite{qszz23,lsz19}, 
two-layer regression problem \cite{swy23,lswy23},
support vector machines \cite{gsz23}, 
and federated learning \cite{bsy23}.

\section{Technique Overview}
\label{sec:technique}

In this paper, we formally define and analyze the novel problem of inverting leverage score distributions to recover model parameters from known leverage scores, namely
\begin{align}\label{eq:inverting_leverage_score}
    \min_{x \in \R^d } \| \diag( \sigma ) - I_n \circ (A(x) (A(x)^\top A(x) )^{-1} A(x)^\top )  \|_F.
\end{align}

This regression problem is very complicated to analyze directly, so we define different functions into $s(x)$, $A(x)$, $\sigma(x)$, and $c$ (see Section~\ref{sec:problem} for an overview and Section~\ref{sec:preli} for a comprehensive explanation). Using these simpler functions, we break down the convoluted Eq.~\eqref{eq:inverting_leverage_score} into simpler pieces and analyze each of these pieces respectively.

\paragraph{Gradient and Hessian Computation.}

The derivation of the first-order gradients and second-order Hessians for the nonconvex optimization problem of inverting leverage score distributions requires the extensive application of matrix differential calculus properties. Specifically, obtaining the gradient of the loss function with respect to the model parameters, namely $\nabla_{x \in \R^d} L_c(x)$, requires understanding how small variations in $x$ propagate through the interconnected leverage score mapping $\sigma(x)$. Leveraging the differentiation formulas for matrix inverses, products, and other constituent operations, we get an expression encompassing terms of the form 
\begin{align*}
    \sigma(x) \diag(A(x)_{*,j} ) \sigma(x)
\end{align*}
(see Section~\ref{sec:gradient} for details).

The computation of the Hessian matrix, represented by $\nabla_{x \in \R^d}^2 L_c(x)$, is a more intricate task. This requires analyzing the second-order partial derivatives. Nonetheless, with careful application of derivative rules and algebraic techniques, we can decompose the dense Hessian into a sum of 6 matrices (see Lemma~\ref{lem:hessian_decompose_informal} for an overview and Section~\ref{sec:hessian} for details) denoted as $D_{i, k} \in \R^{d \times d}$, for $k \in [6]$. Each $D_{i, k}$ corresponds to a distinct second-order interaction between the components of $x \in \R^{d}$, mediated by $A(x) \in \R^{n \times d}$ and $\sigma(x) \in \R^{n \times n}$. For example, $D_1$ isolates the coupling between $\sigma(x)^{\circ 2} \in \R^{n \times n}$ and parameter changes, while $D_2$ focuses on correlations between $\sigma(x)$ and its transpose, $\sigma(x)^\top$.

\paragraph{Showing Hessian is Positive Definite.}

Establishing the positive definite property of the Hessian matrix is extremely important. It is not only for presenting the geometry of the nonconvex loss function from the leverage score inversion problem (Definition~\ref{def:inverting_leverage_score}) but also for ensuring the convergence of second-order optimization algorithms. Specifically, we assume for positive real numbers $R > 0$ and $\beta \in (0, 0.1)$, $\|A\| \leq R$, $\|x\|_2 \leq R$, and $\sigma_{\min} (A(x)) \geq \beta$. Then, we decompose the Hessian $\frac{\d^2 L}{\d x^2} $ into
\begin{align}\label{eq:Ltot_second_derivative}
    H(x) = \frac{\d^2 L}{\d x^2} 
    = & ~ \underbrace{A^\top G(x) A}_{\frac{\d^2 L_{\exp}}{\d x^2}} + \underbrace{A^\top W^2 A}_{\frac{\d^2 L_{\reg}}{\d x^2}} \notag\\
    = & ~ A^\top (G(x) + W^2) A,
\end{align}
where 
\begin{align*}
    G(x) = S(x)^{-1} \sum_{q=1}^6 D_q(x) S(x)^{-1}
\end{align*}
and $D_q(x)$ are defined as in Definition~\ref{def:further_decomposition}. Note that $\frac{\d^2 L_{\exp}}{\d x^2}$ denote the Hessian matrix $\nabla_{x \in \R^d}^2 L_c(x)$\footnote{In this paper, we use $L_c$ and $L_{\exp}$ to denote the same thing.}. We use the regularization term $L_{\reg}$ as a tool to make our loss function to be convex, under the assumption for all $i \in [n]$, $w_{i}^2 \geq -44 \beta + l/\sigma_{\min}(A)^2$ to construct the matrix $W$ as shown in Eq.~\eqref{eq:Ltot_second_derivative}. 

Specifically, theoretical analyses leveraging the boundedness of spectral norms for each $q \in [6]$ of $D_q(x)$, we can get 
\begin{align}\label{eq:bound_g}
    -44 \beta I_d \preceq G(x) \preceq 44 \beta I_d.
\end{align}
Combining Eq.~\eqref{eq:bound_g} with the assumption on $w_{i}^2 \geq -44 \beta + l/\sigma_{\min}(A)^2$, we can conclude the positive definiteness of their aggregate $H(x)$ based on Eq.~\eqref{eq:Ltot_second_derivative}. Formally, existing results demonstrate that $\frac{l}{\sigma_{\min}(A)^2} I_n$ lower bound the matrix $G(x) + W^2$, thereby supporting the overall positive definite curvature of $\frac{\d^2 L}{\d x^2}$. This collective positive definiteness of the Hessian in turn geometrically pins down the presence of a single global loss minimum that second-order methods can effectively seek out.

\paragraph{Showing Hessian is Lipschitz.}

Demonstrating the Lipschitz continuity of the Hessian matrix is fundamental both for certifying the well-defined nature of the optimization problem induced by the leverage score inversion task and for furnishing convergence rate guarantees when deploying second-order iterative methods. This task requires a sophisticated approach that involves breaking down the Hessian matrix and applying precise matrix analysis techniques. The goal is to obtain separate bounds for various components, and then we can combine these bounds using the triangle inequality.

More precisely, we use the matrix calculus techniques to analyze the dense Hessian matrix $H(x)$ computed from our loss function $L_c(x)$. We express it as the sum of simpler matrices, namely 
\begin{align*}
    H(x) = \sum_{q = 1}^6 D_q(x)
\end{align*}
(see Section~\ref{sub:property:lip} for a summary of our key result and Section~\ref{sec:Lipschitz_app} for details).

Then, we consider the bound $\|H(x) - H(y)\|$ for the vectors $x, y \in \R^d$ that are close to each other. To find this bound, we need to establish bounds for each individual $D_q(x)$ matrix. We can take advantage of the stability of leverage score operators to analyze the variations between $D_q(x)$ and $D_q(y)$, expressed as 
\begin{align*}
    \|D_q(x) - D_q(y)\| \leq \xi_q \|x - y\|,
\end{align*}
where $\xi_q \in \R_+$ represents positive Lipschitz constants that quantify how much each individual $D_q$ factor varies (see Lemma~\ref{lem:d_1_lips}, \ref{lem:d_2_lips}, \ref{lem:d_3_lips}, \ref{lem:d_4_lips}, \ref{lem:d_6_lips}, \ref{lem:d_7_lips} for the details of how we compute the exact values of $\xi_q$, $\forall q \in [6]$).

By applying the triangle inequality to find the upper bound of
\begin{align*}
   \|\sum_{q = 1}^6 (D_q(x) - D_q(y))\| 
   \leq & ~ \sum_{q = 1}^6 \|(D_q(x) - D_q(y))\|\\
   \leq & ~ \sum_{q = 1}^6 \xi_q \|x - y\|,
\end{align*}
we finally obtain the Lipschitz constant of $\sum_{q = 1}^6 \xi_q$ for the entire Hessian matrix (see Lemma~\ref{lem:lipschitz_informal} for summary and Lemma~\ref{lem:lipschitz_formal} for details).

\paragraph{The Use of Gradient Descent and Newton's Method Under the Support of the Positive Definite and Lipschitz Properties of the Hessian Matrix.}

Our final step is to use the properties we have analyzed, namely Hessian is positive definite and Lipschitz, to support our construction of the first-order and second-order algorithms (Algorithm~\ref{alg:first_order} and Algorithm~\ref{alg:second_order}) for solving the inverting leverage score problem (see Definition~\ref{def:inverting_leverage_score}).

Specifically, employing first-order method--gradient descent iterations requires to show that the Hessian matrix's positive definiteness, as established in Lemma~\ref{lem:convex_informal}, directly implies that the loss function we're dealing with is strongly convex (see Lemma~\ref{lem:pd_is_convex}). This strong convexity property further allows us to get a convergence rate guarantee for projected gradient descent (as shown in Lemma~\ref{lem:gradient_convergence}). By defining the parameters related to the optimization error's bounded decrease in terms of the condition number and norm bounds, we derive the number of iterations needed to reduce the gradient descent errors to any accuracy.

Regarding the second-order method--Newton method, we use the Lipschitz continuity of the Hessian matrix and combine it with the positive definite property. These two important properties ensure that the conditions of $(l,M)$-good loss function are satisfied, which allows us to employ the quadratic convergence framework. It couples with the quadratic convergence of the optimization error to zero by Lemma~\ref{lem:one_step_shrinking}. Finally, by using induction of ever-shrinking neighborhoods surrounding the optimal parameters formalized in Lemma~\ref{lem:newton_induction}, we derive a complete convergence rate analysis.


\section{Problem Formulation}
\label{sec:problem}

As outlined in our technique overview (see Section~\ref{sec:technique}), we first break the inverting leverage score problem into simpler pieces.

In this section, we present the definitions of the important functions we use. We use these definitions to express our inverting leverage score problem (see Definition~\ref{def:inverting_leverage_score}) into another form for the convenience of further analysis. First, we present the formal definition of leverage score below:
\begin{definition}[Leverage Score]\label{def:leverage_score}
Given a matrix $A \in \R^{n \times d}$, we define its leverage score $\sigma$ to be $A ( A^\top A )^{-1} A^\top \in \R^{n \times n}$.
\end{definition}

Now, we define more matrices and vectors for the purpose of decomposing the complicated inverting leverage score problem.

\begin{definition}\label{def:sigma_A_s}
    Let $A \in \R^{n \times d}$ and $b \in \R^n$. We consider the leverage score of $A(x)$, denoted by 
    \begin{align*}
        \sigma(x) := A(x) ( A(x)^\top A(x) )^{-1} A(x)^\top \in \R^{n \times n},
    \end{align*}
    where $A(x) := S(x)^{-1} A \in \R^{n \times d}$ and $S(x) := \diag( A x - b ) \in \R^{n \times n}$.
\end{definition}

Given a vector $c \in \R^n$, we consider reversibly solving $x$ so that the resulting leverage score $\sigma(x)$ matches the given $c$.

To be specific, we consider the following inversion problem.
\begin{definition}
We consider solving the inversion problem in the following optimization form:
\begin{align*}
    \argmin_x~L_c(x) := \frac{1}{2} \| ( \sigma(x) \circ I_n ) - \diag(c) \|_F^2.
\end{align*}
\end{definition}

For convenience, we express our loss function in another form.
\begin{claim}\label{cla:reformulation}
The loss function $L_c(x)$ can be re-written as 
\begin{align*}
    L_c(x) = L_{\exp} = 0.5 \sum_{i=1}^n (\sigma_{i,i}(x) - c_i )^2
\end{align*}
\end{claim}

Also, we define the regularization term $L_{\reg}$ to resolve the non-convex problem of our inverting leverage score problem (Definition~\ref{def:inverting_leverage_score}).

\begin{definition}\label{def:L_reg}
Given matrix $A \in \R^{n \times d}$, we let $\beta \in (0, 0.1)$, $l > 0$, and $\sigma_{\min}(A)$ be the smallest singular value of $A$.

For a given vector $w \in \R^n$ satisfying for all $i \in [n]$, 
\begin{align*}
    w_{i}^2 \geq -44 \beta + l/\sigma_{\min}(A)^2,
\end{align*}
we let $W = \diag(w)$. 
We define $L_{\reg} : \R^d \rightarrow \R$ as follows
\begin{align*}
L_{\reg}(x):= 0.5 \| W A x\|_2^2
\end{align*}
\end{definition}

Finally, we formulate the final version of the regularized inverting leverage score problem as follows:
\begin{definition}[Regularized inverting leverage score problem]\label{def:regularized_version}
    Let $x \in \R^d$. Let $L_{\reg}(x) \in \R$ be defined as in Definition~\ref{def:L_reg} and $L_{\exp}(x) = L_c(x) \in \R$ be formulated as in Claim~\ref{cla:reformulation}.
    
    Then, we define the regularized inverting leverage score problem $L(x) \in \R$ as
    \begin{align*}
        L(x) 
        := & ~ L_{\exp}(x) + L_{\reg}(x) \\
        = & ~ 0.5 \sum_{i=1}^n (\sigma_{i,i}(x) - c_i )^2 + 0.5 \| W A x\|_2^2.
    \end{align*}
\end{definition}

Note that the gradient and Hessian of the regularization term $L_{\reg}$ is analyzed in \cite{dls23} (we listed this in Lemma~\ref{lem:L_reg_gradient_hessian}). Therefore, in this paper, it suffices to compute the gradient and Hessian of $L_{\exp}(x)$.

\section{Gradient and Hessian}
\label{sec:gradient_hessian}

In this section, we give an overview of our results about the gradient and Hessian. Specifically, in Section~\ref{sub:gradient_hessian:gradient}, we summarize our important result related to the gradient, and in Section~\ref{sub:gradient_hessian:hessian}, we present our important result related to Hessian.

\subsection{Gradient Computation}
\label{sub:gradient_hessian:gradient}

The gradient of the leverage score is presented below:

\begin{lemma}[Informal version of Lemma~\ref{lem:gradient_leverage}]\label{lem:gradient_leverage_informal}
Let $A(x) \in \R^{n \times d}$ and $\sigma(x) \in \R^{n \times n}$ be defined as in Definition~\ref{def:sigma_A_s}.

Then, we have: for each $j \in [d]$,
\begin{itemize}
    \item the derivative of $\sigma(x)_{i,l} \in \R$ with respect to $x_j \in \R$ is:
    \begin{align*}
     \frac{ \d \sigma(x)_{i,l}  }{ \d x_j } 
    = & ~ 2  \langle \sigma(x)_{*,i} \circ \sigma(x)_{*,l} , A(x)_{*,j} \rangle \\
    - & ~ \sigma(x)_{i,l} \cdot (A(x)_{i,j} + A(x)_{l,j} )
   \end{align*} 
   \item the derivative of $\sigma(x) \in \R^{n \times n}$ with respect to $x_j \in \R$ is:
   \begin{align*} 
        \frac{ \d \sigma(x) }{ \d x_j} 
        = & ~ 2 \sigma(x)  \diag(A(x)_{*,j}) \sigma(x) \\
        - & ~ \diag(A(x)_{*,j}) \sigma(x)  -   \sigma(x) \diag( A(x)_{*,j} ) 
   \end{align*}
   \item the derivative of $\sigma(x)_{*,i} \in \R^{n}$ with respect to $x_j \in \R$ is:
   \begin{align*}
        \frac{\d \sigma(x)_{*,i}}{\d x_j} 
        = & ~ 2 \sigma(x) \diag( A(x)_{*,j} ) \sigma_{*,i}(x) \\
        - & ~ \diag( A(x)_{*,j} ) \sigma_{*,i}(x) - \sigma(x)_{*,i} A(x)_{i,j}
   \end{align*}
\end{itemize}
\end{lemma}

\subsection{Hessian Computation}
\label{sub:gradient_hessian:hessian}

In this section, we present our computation of Hessian.

\begin{lemma}[Informal version of Lemma~\ref{lem:hessian_decompose}]\label{lem:hessian_decompose_informal}
Let $A(x) \in \R^{n \times d}$ and $\sigma(x) \in \R^{n \times n}$ be defined as in Definition~\ref{def:sigma_A_s}. Let $H = \frac{\d^2 L_{\exp}(x) }{ \d x^2}$, where $L_{\exp} = L_c(x)$ is formulated as in Claim~\ref{cla:reformulation} and $x_k$ and $x_j$ are two arbitrary entries of $x \in \R^d$. Let $c \in \R^n$.

Then, we can get the Hessian matrix $H \in \R^{d \times d}$ such that
\begin{align*}
   H  = D_{i, 1} + D_{i, 2} + D_{i, 3} + D_{i, 4} + D_{i, 5} + D_{i, 6}
\end{align*}
where
\begin{align*}
    D_{i, 1} &= 4\underbrace{ A(x)^\top }_{d \times n} \underbrace{ \sigma(x)_{*,i}^{\circ 2} }_{n \times 1} \cdot \underbrace{ (\sigma(x)_{*,i}^{\circ 2})^\top }_{1 \times n} \underbrace{ A(x) }_{n \times d} \\
    D_{i, 2} &= -8 \underbrace{ \sigma(x)_{i,i} }_{ \mathrm{scalar} } \underbrace{ ( A(x)_{i,*} )^\top }_{ d \times 1 } \underbrace{ (\sigma(x)_{*,i}^{\circ 2})^\top }_{1 \times n} \underbrace{A(x)}_{n \times d}\\
    D_{i, 3} &= -8 \underbrace{ ( A(x)_{i,*} )^\top }_{d \times 1} \underbrace{ (\sigma(x)_{*,i} \circ \sigma(x)_{*,i})^{\top} }_{1 \times n} \sigma(x)_{i,i} \underbrace{ A(x) }_{n \times d} \\
    D_{i, 4} &= 10  \underbrace{ ( A(x)_{i,*} )^\top }_{d \times 1} \sigma(x)_{i,i}^2 \underbrace{A(x)_{i,*}}_{1 \times d} \\
    D_{i, 5} &= \\
    & 8 \sigma(x)_{i,i} \underbrace{A(x)^\top}_{d \times n} \underbrace{\diag( \sigma_{*,i}(x) )}_{n \times n} \underbrace{\sigma(x)}_{n \times n} \underbrace{\diag( \sigma_{*,i}(x) )}_{n \times n} \underbrace{A(x)}_{n \times d}\\
    D_{i, 6} &= -6 \underbrace{A(x)^{\top}}_{d \times n} \sigma(x)_{i,i}  \underbrace{\diag( \sigma(x)_{*,i}^{\circ 2})}_{n \times n} \underbrace{A(x)}_{n \times d}.
\end{align*}
\end{lemma}

\begin{lemma}[Lemma 4.9 in \cite{dls23}]\label{lem:L_reg_gradient_hessian}
Let $A \in \R^{n \times d}$. For a given vector $w \in \R^n$, we let $W = \diag(w) \in \R^{n \times n}$ and $L_{\reg} : \R^d \rightarrow \R$ be defined as Definition~\ref{def:L_reg}.

Then, we have
\begin{itemize}
\item The gradient is
\begin{align*}
\frac{\d L_{\reg}}{ \d x} = A^\top W^2 Ax
\end{align*}
\item The Hessian is
\begin{align*}
\frac{\d^2 L_{\reg}}{ \d x^2} = A^\top W^2 A
\end{align*}
\end{itemize}
\end{lemma}

Finally, combining Lemma~\ref{lem:hessian_decompose_informal} and Lemma~\ref{lem:L_reg_gradient_hessian}, we can get the Hessian matrix $H(x) = \frac{\d^2 L_{\exp}}{ \d x^2} + \frac{\d^2 L_{\reg}}{ \d x^2} \in \R^{d \times d}$ for our regularized inverting leverage score problem (Definition~\ref{def:regularized_version}).

\section{Properties of Hessian}

\label{sec:property}

In Section~\ref{sub:property:pd}, we show that the Hessian matrix is positive definite. In Section~\ref{sub:property:lip}, we show that the Hessian matrix is Lipschitz.

\subsection{Hessian Matrix is Positive Definite}
\label{sub:property:pd}

In this section, we show that the Hessian matrix $H(x) = \frac{\d^2 L}{\d x^2}$ is positive definite.

\begin{lemma}[Informal version of Lemma~\ref{lem:convex}]\label{lem:convex_informal}

    Let $L(x) = L_{\exp}(x) + L_{\reg}(x)$ be defined as in Definition~\ref{def:regularized_version}. Let $l > 0$.

Then, we have
\begin{align*}
    \frac{\d^2 L}{\d x^2} \succeq l \cdot I_d
\end{align*}
\end{lemma}

\subsection{Hessian Matrix is Lipschitz}
\label{sub:property:lip}

In this section, we show that the Hessian matrix is Lipschitz. We start with formally defining what is $M$-Lipschitz.

\begin{definition}[Hessian is $M$-Lipschitz]\label{def:hessian_lipschitz}

Consider a function $L : \mathbb{R}^d \rightarrow \mathbb{R}$. Let $M > 0$. We say that the Hessian matrix of $L$ is $M$-Lipschitz if for all $x$ and $y$ in $\mathbb{R}^d$:
\begin{align*}
    \| \nabla^2 L(y) - \nabla^2 L(x) \| \leq M \cdot \| y - x \|_2.
\end{align*}
\end{definition}

Then, we present how the Hessian matrix satisfies the definition of $M$-Lipschitz. 

\begin{lemma}[Informal version of Lemma~\ref{lem:lipschitz_formal}]\label{lem:lipschitz_informal}
    Let $H(x)$ be the Hessian matrix, where $H(x) = \frac{\d^2 L}{\d x^2} = \frac{\d^2 L_{\exp}}{\d x^2} + \frac{\d^2 L_{\reg}}{\d x^2}$. Let $R > 0$ and $\beta \in (0, 0.1)$.
    Then we have
    \begin{align*}
        \|H(x) - H(\wh{x})\| \leq 812 \beta^{-9}R^5 \|x - \wh{x}\|_2
    \end{align*}
\end{lemma}

We can see that by picking $M = 812 \beta^{-9}R^5$, the Hessian of the loss function $L$ is $M$-Lipschitz. 

\section{The Analysis of the First-Order Method--Gradient Descent}
\label{sec:order_1}

In this section, we present how we can use the properties of Hessian, namely Hessian is positive definite and Lipschitz to derive the convergent of the gradient descent. First, we give the formal definition of gradient descent.

\begin{definition}[Gradient descent]
    Let $x_t \in \R^d$. Let $L: \R^d \to \R$. For $\gamma > 0$, the following iteration
    \begin{align*}
        x_{t + 1} = x_t - \gamma \nabla L(x_t)
    \end{align*}
    is called the gradient descent.
\end{definition}

Then, we give the formal definition of $\alpha$-strongly convex.

\begin{definition}[$\alpha$-strongly convex \cite{b15}]\label{def:strongly_convex}
    We say that $L : \R^d \rightarrow \mathbb{R}$ is $\alpha$-strongly convex if it satisfies the following improved subgradient inequality:
    \begin{align*}
        L(x) - L(y) \leq \nabla L(x)^\top (x - y) - \frac{\alpha}{2} \|x - y\|_2^2
    \end{align*}
\end{definition}

Now, we present the lemma of showing the positive definiteness of the Hessian matrix may imply the $\alpha$-strongly convex.

\begin{lemma}[Page 459 of \cite{bv04}]\label{lem:pd_is_convex}
    Let $L : \R^d \rightarrow \mathbb{R}$. Let $l > 0$. If the Hessian matrix of $L$ is positive definite, namely $\frac{\d^2 L}{dx^2} \succeq l \cdot I_d$, then $L$ is $l$-strongly convex.
\end{lemma}

Finally, we show that the function possessing the $\alpha$-strongly convex and Lipschitz properties may result in the convergence of the gradient descent.

\begin{lemma}[Theorem 3.9 of \cite{b15}]\label{lem:gradient_convergence}
Let $L : \R^d \rightarrow \mathbb{R}$ be $\alpha$-strongly convex and $l$-Lipschitz. 

Then the projected subgradient descent after $T$ steps with step size
    $\gamma_k = \frac{2}{\alpha(k + 1)}$
satisfies
\begin{align*}
    L( \frac{1}{T(T + 1)} \sum_{k=1}^{T} \frac{2k}{T(T + 1)} x_k ) - L(x^*) \leq \frac{2l^2}{\alpha(T + 1)}.
\end{align*}
\end{lemma}

\section{The Analysis of the Second-Order Method--Newton Method}
\label{sec:newton}

In this section, we introduce the Newton method. We start with introducing the definition of $l$-local minimum.

\begin{definition}[$l$-local minimum]\label{def:local_min}

Consider a function $L : \mathbb{R}^d \rightarrow \mathbb{R}$.
Let $l > 0$ be a positive real number.
If there exists a vector $x^* \in \mathbb{R}^d$ satisfy
\begin{align*}
    \nabla L(x^*) &= \mathbf{0}_d \\
    \nabla^2 L(x^*) &\succeq l \cdot I_d,
\end{align*}
then we say $x^*$ is $l$-local minimum.

\end{definition}

Now we present the definition of the good initialization point.

\begin{definition}[Good Initialization Point]\label{def:good_point}

For a function $L$ that maps from $\mathbb{R}^d$ to $\mathbb{R}$. Suppose $x_0$ and $x^*$ are in $\mathbb{R}^d$. Let $r_0$ be $r_0 = \|x_0 - x^*\|_2$. Also, let $M$ be a positive real number. If $r_0$ satisfies the condition $r_0 M \leq 0.1 l$, where $l$ is some constant, then we define $x_0$ as a good initialization point.

\end{definition}

Then, we expect our loss function $L$ to exhibit certain desirable properties that align with the use of the Newton method. These properties are displayed as follows:

\begin{definition}[$(l,M)$-good Loss function]\label{def:f_ass}

Consider a function $L : \mathbb{R}^d \rightarrow \mathbb{R}$.
We say $L$ $(l,M)$-good function if it meets the following criteria:
\begin{itemize}
    \item It is $l$-local minimum (as defined in Definition~\ref{def:local_min}).
    \item Its Hessian matrix is $M$-Lipschitz continuous (as defined in Definition~\ref{def:hessian_lipschitz}).
    \item It had a good initialization point (as defined in Definition~\ref{def:good_point}).
\end{itemize}

\end{definition}

Then, we present the exact update of Newton's method.
\begin{definition}[Exact update of the Newton method]\label{def:exact_update_variant}

Let $H$ denote the Hessian matrix and $g$ denote the gradient. Then, we define the exact update of Newton's method as the following iteration:
\begin{align*}
    x_{t+1} = x_t - H(x_t)^{-1} \cdot g(x_t)
\end{align*}
\end{definition}

Now, we present a lemma from \cite{lsz23}.
\begin{lemma}[Lemma 6.9 of \cite{lsz23}]\label{lem:one_step_shrinking}

Consider a function $L : \mathbb{R}^d \rightarrow \mathbb{R}$ that is defined as a $(l,M)$-good function, as described in Definition~\ref{def:f_ass}. Let $\epsilon_0$ be a positive real number within the interval $(0,0.1)$. Suppose $x_t$ and $x^*$ are both elements of $\mathbb{R}^d$, and let $r_t := \| x_t - x^* \|_2$. Furthermore, let $M$ be a positive real number and define $\overline{r}_t := M \cdot r_t$.

Under these conditions, the following inequality holds:
\begin{align*}
    r_{t+1} \leq 2 \cdot ( \epsilon_0 + \frac{\overline{r}_t}{l - \overline{r}_t} ) \cdot r_t.
\end{align*}

\end{lemma}

We use $T$ to denote the total number of iterations conducted by the algorithm. For using Lemma~\ref{lem:one_step_shrinking}, we need to establish the following inductive hypothesis, which is developed in \cite{lsz23}.

\begin{lemma}[Mathematical induction, Lemma 6.10 on page 34 of \cite{lsz23}]\label{lem:newton_induction}

Consider an index $i\in [t]$. Suppose $x_i$ and $x^*$ are both elements of $\mathbb{R}^d$, and let $r_i := \| x_i - x^* \|_2$. Let $\epsilon_0$ be a positive real number within the range $(0,0.1)$. Assume that $0.4 \cdot r_{i-1} \geq r_{i}$. Additionally, ensure that $0.1 \cdot l \geq M \cdot r_i$, where $M > 0$ is a positive real number.

Then, we have
\begin{itemize}
    \item $0.4 r_t \geq r_{t+1}$
    \item $0.1 l \geq M \cdot r_{t+1}$
\end{itemize}

\end{lemma}

\section{Conclusion}\label{sec:conclusion}

In this paper, we propose and analyze the novel problem of inverting leverage score distributions to recover model parameters from known leverage scores. We introduce a regularized term to address this non-convex problem. Additionally, we compute the gradient and Hessian of our loss function. Furthermore, we demonstrate that this problem possesses useful properties, such as positive definiteness and Lipschitz continuity of the Hessian matrix. These properties enable the development and convergence rate analysis of both first-order (gradient descent) and second-order (Newton method) algorithms for determining the parameters that produced the given leverage scores.

We believe that our theoretical analysis of inverting leverage scores may open up numerous new applications, including model interpretation, data recovery, and security. The ability to reverse engineer parameters from leverage score distributions facilitates a better understanding of model behavior and vulnerabilities. It also creates new possibilities for reconstructing private training data. Overall, our work has laid important foundations in inversion methodology, optimization guarantees, and algorithm development for the novel task of inverting leverage scores. Further exploration of more complex data distributions, kernel methods, and additional model families can build upon these results.

\section*{Impact Statement}

This paper presents work whose goal is to advance the field of Machine Learning. There are many potential societal consequences of our work, none of which we feel must be specifically highlighted here.

\ifdefined\isarxiv
\else
\bibliography{ref}
\bibliographystyle{icml2024}

\fi

\newpage
\onecolumn
\appendix

\paragraph{Roadmap}
We organize the appendix as follows. In Section~\ref{sec:preli} we provide the notations to be used in this paper and some useful tools for differential computation, exact computation and approximate computation. In Section~\ref{sec:gradient}, we compute the gradient for $\sigma(x)$ step by step. In Section~\ref{sec:hessian}, we compute the hessian for $L_{\exp}(x)$ step by step, and further decomposed it for further analysis. In Section~\ref{sec:psd}, we proved that $L$ is convex. In Section~\ref{sec:Lipschitz_app}, we are able to show that $L_{\exp}$ is lipschitz continuous and computed its lipschitz constant. In Section~\ref{sec:computation}, we analyzed the running time required for computing $\nabla L$ and $\nabla^2 L$ step by step. In Section~\ref{sec:main_result}, we state our result in an formal way, which suggests that we are able to solve the inverting leverage score problem with high accuracy and acceptable running time.

\section{Preliminary}
\label{sec:preli}

In this section, we provide the preliminaries to be used in our paper. First, we define the notations of our paper. Then, in Section~\ref{sec:preli:facts}, we provide tools for differential computation, exact computation and approximate computation.

\paragraph{Notations.}

First, we define the sets. Let $\Z_+ := \{1, 2, 3, \dots\}$. Let $n, d \in \Z_+$.

Now, consider the vectors $x, y \in \R^d$. We define $x_i \in \mathbb{R}$ to be the $i$-th element of vector $x$, where $i$ ranges over $[d]$. $\circ$ denotes the entrywise product between the same dimensional matrices or vectors.
 Also, we have $x^{\circ 2} = x \circ x$. For all $p \in \Z_+$, we define the $\ell_p$ norm of the vector $x$, denoted as $\|x\|_p$ to be $\sqrt[p]{\sum_{i = 1}^d |x_i|^p}$. Additionally, we define $\|x\|_\infty := \max_{i \in [d]} |x_i|$. We define ${\bf 1}_n$ as the all $1$ vector. $e_k$ is a vector whose $k$-th entry equals 1 and other entries are 0. When dealing with iterations, we use $x_t$ to denote the $t$-th iteration. In this paper, we only use the letters $t$ and $T$ for expressing iterations. $\langle \cdot, \cdot \rangle$ is the inner product. We define the function $\mathrm{diag} : \mathbb{R}^d \rightarrow \mathbb{R}^{d \times d}$ such that $\mathrm{diag}(x)_{i, i} := x_i$ and $\mathrm{diag}(x)_{i, j} := 0$ if $i$ is not equal to $j$.

Consider $A \in \mathbb{R}^{n \times d}$. For any $i \in [n]$ and $j \in [d]$, we denote $A_{i, j} \in \mathbb{R}$ as the entry of $A$ positioned at the $i$-th row and $j$-th column. We define $(A_{i, *})^\top \in \mathbb{R}^d$ as the $i$-th row vector of $A$ and $A_{*, j} \in \mathbb{R}^n$ as the $j$-th column vector of $A$. The matrix $A^\top \in \mathbb{R}^{d \times n}$ represents the transpose of $A$, and $I_n$ denotes the $n$-dimensional identity matrix.

For $A$, we define the spectral norm and the Frobenius norm as $\|A\| := \max_{x \in \mathbb{R}^d} \|Ax\|_2 / \|x\|_2$ where $\|x\|_2 \neq 0$ and $\|A\|_F := \sqrt{\sum_{i = 1}^{n}\sum_{j = 1}^{d}| A_{i, j} |^2}$, respectively.

\begin{definition}\label{def:Tmat}
For any positive integers $a,b,c$, we use $\Tmat(a,b,c)$ to denote the time of multiplying an $a \times b$ matrix with another $b \times c$ matrix.
\end{definition}

\begin{fact}\label{fac:basic_Tmat}
\begin{align*}
\Tmat(a,b,c,d) = O( \min \{ \Tmat(a,b,c) + \Tmat(a,c,d) , \Tmat(b,c,d) + \Tmat(a,b,d)  \} )
\end{align*}
\end{fact}

\begin{fact}\label{fac:Tmat}
We have
\begin{align*}
    \Tmat(a,b,c) = O(\Tmat(a,c,b)) = O( \Tmat(b,a,c) ) = O(\Tmat(b,c,a) ) = O(\Tmat(c,a,b)) = O( \Tmat(c,b,a) )
\end{align*}
\end{fact}

\begin{fact}\label{fac:max_vs_Tmat}
We have
\begin{align*}
 \max\{ ab, bc, ac\} = O (\Tmat(a,b,c))
\end{align*} 
\end{fact}

\subsection{Basic Definitions}

\begin{definition}\label{def:s}
Given $A \in \R^{n \times d}$, and $b \in \R^{n}, x \in \R^d$, we define $s(x) \in \R^n $
\begin{align*}
    s(x) := Ax - b
\end{align*}
We define diagonal matrix $S(x) = \diag(s(x)) \in \R^{n \times n}$
\end{definition}

\begin{definition}\label{def:Ax}
We define $A(x) \in \R^{n \times d}$
\begin{align*}
    A(x) := S(x)^{-1} A
\end{align*}
For convenient, for each $i \in [n]$ we use $a(x)_i^\top \in \R^{1 \times d}$ to denote the $i$-th row of matrix $A(x) \in \R^{n \times d}$
\end{definition}

\begin{definition}\label{def:sigma}
We define matrix $\sigma(x) \in \R^{n \times n}$
\begin{align*}
    \sigma(x) := \underbrace{ A(x) }_{n \times d} \underbrace{ ( A(x)^\top A(x) )^{-1} }_{d \times d} \underbrace{ A(x)^\top }_{d \times n}
\end{align*}
Then it is easy to see that
\begin{itemize}
    \item $\sigma(x)_{i,i} = a(x)_i^\top ( A(x)^\top A(x) )^{-1} a(x)_i$ for each $ i \in [n]$
    \item $\sigma(x)_{i,l} = a(x)_i^\top ( A(x)^\top A(x) )^{-1} a(x)_l$ for each $i \in [n]$ and for each $l \in [n]$
\end{itemize}
\end{definition}

\begin{definition}\label{def:Q}
We define matrix $Q(x) \in \R^{n \times n}$ such that
\begin{align*}
  Q(x) : = \sigma(x) \circ \sigma(x)
\end{align*}
It is easy to see that
\begin{itemize}
    \item $Q(x)_{i,l} = \sigma(x)_{i,l}^2$ for each $i \in [n]$, for each $l \in [n]$
\end{itemize}
\end{definition}

\begin{definition}\label{def:Sigma}
We deine matrix $\Sigma(x) \in \R^{n \times n}$ such that 
\begin{align*}
    \Sigma(x) := \sigma(x) \circ I_n.
\end{align*}
\end{definition}

\subsection{Basic Mathematical Facts}\label{sec:preli:facts}

\begin{fact}\label{fac:diag_rule}
    For a matrix $D \in \R^{n \times n}$ and a vector $x \in \R^n$, 
    \begin{itemize}
        \item {\bf Part 1.} $D \diag(x) = \diag(Dx)$
        \item {\bf Part 2.} For $i \in [n] $, $\frac{\d \diag(x)}{\d x_i} = \diag(\frac{\d x}{\d x_i})$ 
        \item {\bf Part 3.} Let $D$ be an invertible square matrix, then we have $\frac{\d D^{-1}}{\d t} = - D^{-1} \frac{\d D}{\d t} D^{-1}$
    \end{itemize}
\end{fact}

\begin{fact} \label{fac:exponential_der_rule}
Consider functions $g$ and $f$ mapping from $\mathbb{R}^d$ to $\mathbb{R}^n$, and a function $q$ mapping from $\mathbb{R}^d$ to $\mathbb{R}$.

Take any vector $x$ from $\mathbb{R}^d$ and any real number $a$ from $\mathbb{R}$.

Then, we have
    \begin{itemize}
        \item {\bf Part 1.} $\frac{\d q(x)^a}{\d x} =  a\cdot q(x)^{a-1} \cdot \frac{\d q(x)}{\d x}$
        \item {\bf Part 2.} $\frac {\d \|f(x) \|^2_2}{\d t} = 2 \langle f(x) , \frac{\d f(x)}{\d t} \rangle $
        \item {\bf Part 3.} $\frac{\d \langle f(x), g(x) \rangle}{\d t} = \langle \frac{\d f(x)}{\d t} , g(x) \rangle + \langle f(x), \frac{\d g(x)}{\d t} \rangle$
        \item {\bf Part 4.} $\frac{\d (g(x) \circ f(x))}{\d t} = \frac{\d g(x)}{\d t} \circ f(x) + g(x) \circ \frac{\d f(x)}{\d t}$
        \item {\bf Part 5.} Suppose $f(x) \in \R^n$, $\frac{\d f(x)^q }{\d t} = q \cdot ( \frac{\d f(x)}{\d t} ) \circ f(x)^{q-1} $
        \item {\bf Part 6.} $\frac{\d^2 q(x)^2 }{\d x_i x_j} = 2  \frac{\d q(x)}{\d x_i} \cdot \frac{\d q(x) }{\d x_j} + 2q(x) \cdot \frac{\d^2 q(x)}{\d x_i \d x_j}$
    \end{itemize}
\end{fact}

\begin{proof}

{\bf Proof of Part 6.}
\begin{align*}
    \frac{\d^2 q(x)^2 }{\d x_i x_j}
    = & ~ \frac{\d }{\d x_i} ( \frac{ \d q(x)^2 }{\d x_j } ) \\
    = & ~ \frac{\d }{\d x_i} ( 2 q(x) \cdot \frac{\d q(x) }{\d x_j} ) \\
    = & ~ 2 \frac{\d q(x)}{\d x_i} \cdot \frac{\d q(x) }{\d x_j} + 2q(x) \cdot \frac{\d^2 q(x)}{\d x_i \d x_j}
\end{align*}
\end{proof}

\begin{fact}[Rank-$1$ decomposition]\label{fac:rank_1_decomposition}
Let $A \in \R^{n \times d}$, let $a_l^\top \in \R^{1 \times d}$ be the $l$-th row of matrix $A \in \R^{n \times d}$. Let $D \in \R^{n \times n}$ denote a diagonal matrix. Then we have 
\begin{align*}
    A^\top D A = \sum_{l=1}^n \underbrace{ a_l }_{d \times 1} D_{l,l} \underbrace{a_l^\top }_{1 \times d}
\end{align*}
\end{fact}

\begin{fact}\label{fac:circ_rule}

    For any real numbers $a$ and $b$, for every vectors $u$, $v$, and $w$ belonging to $\mathbb{R}^{n}$, we have:
    \begin{itemize}
        \item $\langle u,v \rangle = \langle u \circ v, {\bf 1}_n \rangle =  u^\top \mathrm{diag}(v)  {\bf 1}_n $
        \item $\langle u \circ v, w\rangle = \langle u \circ w, v\rangle$
        \item $\langle u \circ v, w \rangle =  \langle u \circ v \circ w, {\bf 1}_n  \rangle = u^\top \diag(v) w$
        \item $\langle u \circ v \circ w \circ z , {\bf 1}_n \rangle = u^\top \diag(v \circ w) z$
        \item $u \circ  v = v \circ u = \diag (u) \cdot v = \diag (v) \cdot u$ 
        \item $u^{\top}(v \circ w) = v^{\top}(u \circ w) = w^{\top}(u \circ v)= u^{\top}\diag(v) w = v^{\top}\diag(u) w = w^{\top}\diag(u) v$
        \item $ \diag (u)^{\top} = \diag (u)$
        \item $\diag (u) \cdot \diag (v) \cdot {\bf 1}_n = \diag(u) v$
        \item $\diag (u \circ v) = \diag (u) \diag (v)$
        \item $\diag (u) + \diag (v) = \diag (u +v)$
        \item $\langle u,v \rangle = \langle v,u \rangle$
        \item $\langle u,v \rangle = u^\top v = v^\top u$
        \item $a\langle w, v \rangle + b\langle u, v \rangle = \langle aw + bu, v \rangle = \langle v, aw + bu \rangle = a\langle v, w \rangle + b\langle v, u \rangle$.
        \item Let $A \in \R^{k \times k}$ and $x \in \R^{k}$, then we have $\langle A, x x^\top \rangle = x^\top A x$
    \end{itemize}
    
\end{fact}

\begin{fact}\label{fac:vector_norm}
For vectors $u, v \in \R^n$, we have
\begin{itemize}
    \item $\langle u, v \rangle \leq \| u \|_2 \cdot \| v \|_2$ (Cauchy-Schwarz inequality)
    \item $\|\diag(u)\| \leq \|u\|_{\infty}$
    \item $\| u \circ v \|_2 \leq \| u \|_{\infty} \cdot \| v \|_2$
    \item $\| u \|_{\infty} \leq \| u \|_2 \leq \sqrt{n} \cdot \|u\|_\infty$
    \item $\|u\|_2 \leq \|u\|_1 \leq \sqrt{n}\cdot \| u \|_2$
    \item $\| \exp(u) \|_{\infty} \leq  \exp( \| u \|_{\infty}) \leq \exp(\| u \|_2)$ 
    \item Let $\alpha$ be a scalar, then $\| \alpha \cdot u \|_2 = |\alpha| \cdot \| u \|_2$
    \item $\|u + v\|_2 \leq \|u\|_2 + \|v\|_2$.
    \item For any $\| u - v \|_{\infty} \leq 0.01$, we have $\| \exp(u) - \exp(v) \|_2 \leq \| \exp(u) \|_2 \cdot 2 \| u - v \|_{\infty}$
    \item For any $u, v \in \R^d$ such that $\|u\|_2, \|v\|_2 \leq R$, we have $\|\exp(u) - \exp(v)\| \leq \exp(R)\|u - v\|_2$
\end{itemize}
\end{fact}

\begin{fact}\label{fac:matrix_norm}
For matrices $U,V$, we have 
\begin{itemize}
    \item $\| U^\top \| = \| U \|$
    \item $\| U \| \geq \| V \| - \| U - V \|$
    \item $\| U + V \| \leq \| U \| + \| V \|$
    \item $\| U \cdot V \| \leq \| U \| \cdot \| V \|$ 
    \item If $U \preceq \alpha \cdot V$, then $\| U \| \leq \alpha \cdot \| V \|$
    \item For scalar $\alpha \in \R$, we have $\| \alpha \cdot U \| \leq |\alpha| \cdot \| U \|$
    \item For any vector $v$, we have $\| U v \|_2 \leq \| U \| \cdot \| v \|_2$.
    \item Let $u, v \in \R^n$ denote two vectors, then we have $\| u v^\top \| \leq \| u \|_2 \| v \|_2$
\end{itemize}
\end{fact}

\section{Gradient}
\label{sec:gradient}

In this section, we compute the gradient for leverage scores step by step. First, in Section~\ref{sec:gradient:basic}, we computed the gradient of some basic terms. Then, in Section~\ref{sec:gradient:final}, we computed the gradient for leverage scores by using the gradient computed in the previous section.

\subsection{Basic Gradients}\label{sec:gradient:basic}
\begin{lemma}\label{lem:basic_gradient}
Suppose we have
\begin{itemize}
\item  Define $s(x)$ as described in Definition~\ref{def:s}, where $s(x)$ is a vector in $\mathbb{R}^n$.
\item  Define $S(x)$ as stated in Definition~\ref{def:s}, where $S(x)$ is a matrix in $\mathbb{R}^{n \times n}$.
\item  Define $A(x)$ according to Definition~\ref{def:Ax}, where $A(x)$ is a matrix in $\mathbb{R}^{n \times d}$.
    \item $a_i^\top$ is the $i$-th row of $A$, for each $i \in [n]$
    \item Let $a_i(x)^\top$ denote the $i$-th row of $A(x)$, for each $i \in [n]$
    \item Let $A \in \R^{n \times d}$
    \item Let $A_{*,j} \in \R^n$ denote the $j$-th column of matrix $A$
\end{itemize}
Then, we have
\begin{itemize}
    \item {\bf Part 1.} For each $j \in [d]$
    \begin{align*}
        \underbrace{\frac{\d s(x) }{\d x_j}}_{n \times 1} = \underbrace{A_{*,j}}_{ n \times 1}
    \end{align*}
    \item {\bf Part 2.} For each $j \in [d]$
    \begin{align*}
        \underbrace{\frac{\d S(x) }{\d x_j}}_{ n \times n} = \underbrace{\diag( A_{*,j} )}_{n \times n}
    \end{align*}
    \item {\bf Part 3.}
    \begin{align*}
        \underbrace{\frac{\d S(x)^{-1} }{\d x_j}}_{n \times n} = - \underbrace{\diag( \underbrace{ A(x)_{*,j}}_{n \times 1} ) }_{n \times n }\underbrace{S(x)^{-1}}_{n \times n}
    \end{align*}
    \item {\bf Part 4.}
    \begin{align*}
        \underbrace{\frac{ \d A(x) }{ \d x_j }}_{n \times d} =  - \underbrace{\diag(A(x)_{*,j})}_{n \times n} \underbrace{A (x) }_{n \times d}
    \end{align*}
    \item {\bf Part 5.}
    \begin{align*}
        \underbrace{ \frac{\d S(x)^{-2}}{\d x_j} }_{n \times n} =  2 \underbrace{\diag( - \underbrace{S(x)^{-3}}_{n \times n} \underbrace{A_{*,j}}_{n \times 1} )}_{n \times n}
    \end{align*}
    \item {\bf Part 6.} 
    \begin{align*}
        \underbrace{ \frac{ \d A^\top S(x)^{-2} A }{ \d x_j } }_{d \times d} = 2 \underbrace{ A^\top }_{d \times n} \underbrace{ \diag( - S(x)^{-3} A_{*,j} ) }_{n \times n} \underbrace{ A }_{n \times d}
    \end{align*}
    \item {\bf Part 7.}
    \begin{align*}
        \underbrace{\frac{\d A(x)^\top A(x) }{\d x_j}}_{d \times d} = - 2 \underbrace{A(x)^{\top}}_{d \times n}  \underbrace{\diag(A(x)_{*,j})}_{n \times n} \underbrace{A(x)}_{n \times d} 
    \end{align*}
    \item  {\bf Part 8.}
    \begin{align*}
        \underbrace{ \frac{\d A(x)_{*,j}}{ \d x_k} }_{n \times 1} = & ~ - \underbrace{ A(x)_{*,k} }_{n \times 1} \circ \underbrace{ A(x)_{*,j} }_{n \times 1}
    \end{align*}
    \item  {\bf Part 9.}
    \begin{align*}
            \frac{\d S(x)^{-1}_{i,i}}{\d x_k}
    = & ~ - S(x)_{i,i}^{-1} A(x)_{i,k}
    \end{align*}
    \item {\bf Part 10.}
    \begin{align*}
        \frac{\d A(x)_{i,j}}{\d x_k} = - A(x)_{i,k}A(x)_{i,j}
    \end{align*}
\end{itemize}
\end{lemma}

\begin{proof}

{\bf Proof of Part 1.}
\begin{align*}
    \frac{\d s(x) }{\d x_j} 
    = & ~ \frac{\d Ax -b }{\d x_j} \\
    = & ~ A_{*,j},
\end{align*}
where the initial step stems from the definition of $s(x)$, while the subsequent step results from straightforward algebraic manipulation.

{\bf Proof of Part 2.}
\begin{align*}
     \frac{\d S(x) }{\d x_j} 
     = & ~ \frac{\d \diag(s(x))}{\d x_j} \\
     = & ~ \diag( A_{*,j}),
\end{align*}
where the first step follows from the definition of $S(x)$, and the second step follows from Fact~\ref{fac:diag_rule} and {\bf Part 1}.

  {\bf Proof of Part 3.}
    \begin{align*}
         \frac{\d S(x)^{-1} }{\d x_j} = & ~ -S(x)^{-2} \frac{\d S(x) }{\d x_j} \\
         = & ~ - S(x)^{-2} \diag( A_{*,j} ) \\
         = & ~ - \diag(A(x)_{*,j}) S(x)^{-1},
    \end{align*}
    the initial step arises from utilizing Fact~\ref{fac:exponential_der_rule} on every diagonal element of $S(x)^{-1}$, the second step ensues from {\bf Part 2}, and the final step derives from the definition of $A(x)$ and Fact~\ref{fac:diag_rule}.

    {\bf Proof of Part 4.}
    \begin{align*}
        \frac{ \d A(x) }{ \d x_j } 
        = & ~   \frac{ \d S(x)^{-1}A }{ \d x_j } \\ 
        = & ~ \frac{\d S(x)^{-1}}{\d x_j} A \\
        = & ~ - \diag(A(x)_{*,j})S(x)^{-1}   A \\
        = & ~ - \diag( A(x)_{*,j} ) A(x),
    \end{align*}
    where the first step follows from the definition of $A(x)$, the second step follows from $\frac{ \d A }{ \d x_j } = {\bf 0}_{n \times d}$, the third step follows from {\bf Part 3}, and the last step follows from the definition of $A(x)$ (see Definition~\ref{def:Ax}). 
    
    {\bf Proof of Part 5.}
    \begin{align*}
        \frac{\d S(x)^{-2} }{\d x_j}  = & ~ -2 S(x)^{-3} \frac{\d S(x) }{\d x_j} \\
        = & ~ - 2 S(x)^{-3} \diag( A_{*,j} ) \\
        = & ~  2 \diag( - S(x)^{-3} A_{*,j} )
    \end{align*}
    where the 1st step is a result of applying Fact~\ref{fac:exponential_der_rule} to every diagonal component of $S(x)^{-2}$, the 2nd step comes from {\bf Part 2}, and the final step arises from Fact~\ref{fac:diag_rule}.

    {\bf Proof of Part 6.}
    \begin{align*}
        \frac{ \d A^\top S(x)^{-2} A }{ \d x_j } = & ~ A^\top\frac{\d  S(x)^{-2} }{\d x_j }  A \\
        = & ~ 2 A^\top\diag( - S(x)^{-3} A_{*,j} )A
    \end{align*}    
    where the first step is through using the chain rule and the 2nd step is because of {\bf Part 5 }.

    {\bf Proof of Part 7.}
    \begin{align*}
        \frac{\d A(x)^\top A(x) }{\d x_j}
        = & ~  \frac{\d A(x)^\top }{\d x_j}A(x) + A(x)^\top \frac{\d A(x) }{\d x_j} \\
        = & ~ - (  \diag(A(x)_{*,j}) A(x) )^{\top}A(x)  - A(x)^\top   \diag(A(x)_{*,j}) A(x) \\
        = & ~ - 2A(x)^{\top}  \diag(A(x)_{*,j})A(x)
    \end{align*}
    where the first step follows from applying Fact~\ref{fac:exponential_der_rule}, the second step follows from {\bf Part 4}, the third step follows from simple algebra, and the last step follows from simple algebra.

{\bf Proof of Part 8.}
\begin{align*}
        \frac{\d A(x)_{*,j}}{ \d x_k} 
        = & ~ \frac{\d S(x)^{-1}  }{ \d x_k } \cdot A_{*,j} \\
        = & ~ - \diag( A(x)_{*,k} )  S(x)^{-1} A_{*,j} \\
        = & ~ - \diag( A(x)_{*,k} ) A(x)_{*,j} \\
        = & ~ - A(x)_{*,k} \circ A(x)_{*,j} 
    \end{align*}
where the initial step arises from the definition of $A(x)_{*,j}$, the 2nd step comes from {\bf Part 3}, the third step stems from the definition of $A(x)_{*,j}$, and the final step results from Fact~\ref{fac:circ_rule}.

{\bf Proof of Part 9.}
\begin{align*}
    \frac{\d S(x)^{-1}_{i,i}}{\d x_k} 
    = & ~ - ( \diag(A(x)_{*,k}) S(x)^{-1} )_{i,i} \\
    = & ~ - S(x)_{i,i}^{-1} A(x)_{i,k}
\end{align*}
where the first step follows from {\bf Part 3},  and the second step follows from $S(x)$ is a diagonal matrix .

{\bf Proof of Part 10.}
\begin{align*}
    \frac{\d A(x)_{i,j}}{\d x_k}
    = & ~ \frac{ \d ( S(x)^{-1} A)_{i,j} }{ \d x_k} \\
    = & ~ \frac{\d S(x)^{-1}_{i,i}}{\d x_k} \cdot A_{i,j}\\
    = & ~ -S(x)_{i,i}^{-1} A(x)_{i,k} A_{i,j}\\
    = & -A(x)_{i,k} A(x)_{i,j}
\end{align*}
where the initial step arises from $A(x)=S(x)^{-1} A$, the second step is based on the fact that $S(x)$ is a diagonal matrix, the third step follows from {\bf Part 9}, and the final step results from $A(x)=S(x)^{-1} A$.
\end{proof}

\subsection{Gradient for Leverage Scores}\label{sec:gradient:final}

\begin{lemma}\label{lem:gradient_leverage}
Under the following conditions:
\begin{itemize}
\item Define $A(x) \in \mathbb{R}^{n \times d}$ as described in Definition~\ref{def:Ax}.
\item Define $\sigma(x) \in \mathbb{R}^{n \times n}$ as specified in Definition~\ref{def:sigma}.
\item Define $\Sigma(x) \in \mathbb{R}^{n \times n}$ according to Definition~\ref{def:Sigma}.
\item Denote $a(x)_i^\top \in \mathbb{R}^{1 \times d}$ as the $i$-th row of $A(x) \in \mathbb{R}^{n \times d}$ for each $i \in [n]$ (refer to Definition~\ref{def:Ax}).
\end{itemize}
Then, we have: for each $j \in [d]$
\begin{itemize}
    \item {\bf Part 1.} For each $i \in [n]$
    \begin{align*}
        \frac{\d a(x)_i }{\d x_j} = - A(x)_{i,j} a(x)_i
    \end{align*}
    \item {\bf Part 2.}
    \begin{align*}
        \frac{ \d (A(x)^\top  A(x))^{-1} }{ \d x_j } = 2 (A(x)^\top  A(x))^{-1} A(x)^\top \diag( A(x)_{*,j} ) A(x) (A(x)^\top  A(x))^{-1}
    \end{align*}
    \item {\bf Part 3.} For each $i \in [n]$
    \begin{align*}
      \frac{\d \sigma(x)_{i,i}}{\d x_j} =  2 \langle \sigma(x)_{*,i} \circ \sigma(x)_{*,i}, A(x)_{*,j} \rangle - 2 \sigma(x)_{i,i}  A(x)_{i,j} 
    \end{align*}
    \item {\bf Part 4.} 
    \begin{align*}
        \frac{\d  \Sigma(x) }{\d x_j} = 2 \diag( (Q(x) - \Sigma(x) ) A(x)_{*,j} )
    \end{align*}
    \item {\bf Part 5.}
    \begin{align*}
     \frac{ \d \sigma(x)_{i,l}  }{ \d x_j } 
    = & ~ 2  \langle \sigma(x)_{*,i} \circ \sigma(x)_{*,l} , A(x)_{*,j} \rangle -  \sigma(x)_{i,l} \cdot (A(x)_{i,j} + A(x)_{l,j} )
   \end{align*} 
   \item {\bf Part 6.}
   \begin{align*} 
        \frac{ \d \sigma(x) }{ \d x_j} = 2 \underbrace{ \sigma(x) }_{n \times n} \diag(A(x)_{*,j}) \sigma(x) -  \diag(A(x)_{*,j}) \sigma(x)  -   \sigma(x) \diag( A(x)_{*,j} ) 
   \end{align*}
   \item {\bf Part 7.}
   \begin{align*}
        \frac{\d \sigma_{*,i}(x)}{\d x_j} = 2 \sigma(x) \diag( A(x)_{*,j} ) \sigma_{*,i}(x) - \diag( A(x)_{*,j} ) \sigma_{*,i}(x) - \sigma(x)_{*,i} A(x)_{i,j}
   \end{align*}
\end{itemize}
\end{lemma}
\begin{proof}

{\bf Proof of Part 1.}
This is a direction application of $\frac{\d A(x)}{\d x_j}$ by selecting $i$-th row.

{\bf Proof of Part 2.}
\begin{align*}
     \frac{ \d (A(x)^\top  A(x))^{-1} }{ \d x_j } 
     = & ~ - (A(x)^\top  A(x))^{-1} \frac{\d (A(x)^\top  A(x)) }{\d x_j} (A(x)^\top  A(x))^{-1} \\
     = & ~  (A(x)^\top  A(x))^{-1} 2 A(x)^\top \diag( A(x)_{*,j} ) A(x) (A(x)^\top  A(x))^{-1} \\
     = & ~ 2 (A(x)^\top  A(x))^{-1} A(x)^\top \diag( A(x)_{*,j} ) A(x) (A(x)^\top  A(x))^{-1}
\end{align*}
where the initial step is derived from {\bf Part 3} of Fact~\ref{fac:diag_rule}, the second step stems from {\bf Part 7} of Lemma~\ref{lem:basic_gradient}, and the final step results from straightforward algebraic manipulation.

{\bf Proof of Part 3.}

We can show
\begin{align*}
      \frac{\d \sigma(x)_{i,i}}{\d x_j} = & ~ \frac{ \d a(x)_i^\top (A(x)^\top  A(x))^{-1} a(x)_i }{\d x_j } \\
      = & ~ + 2   \frac{ \d a(x)_i^\top }{\d x_j }  (A(x)^\top  A(x))^{-1} a(x)_i \\
      & ~ + a(x)_i^\top \frac{\d (A(x)^\top  A(x))^{-1} }{\d x_j} a(x)_i \\
      = & ~  -2 A(x)_{i,j} a(x)_i^\top (A(x)^\top  A(x))^{-1} a(x)_i \\
      & ~ + 2 a(x)_i^\top  (A(x)^\top  A(x))^{-1} A(x)^\top \diag( A(x)_{*,j} ) A(x) (A(x)^\top  A(x))^{-1} a(x)_i 
\end{align*}
where the initial step is based on the definition of $\sigma(x)$, the second step results from a fundamental differential rule, and the third step arises from {\bf Part 1} and {\bf Part 2}.

For the first term in the above, we have
\begin{align}\label{eq:first_term_sigma}
-2 A(x)_{i,j} a(x)_i^\top (A(x)^\top  A(x))^{-1} a(x)_i = - 2 A(x)_{i,j} \sigma(x)_{i,i}
\end{align}
which a consequence of the definition of $\sigma(x)$.

For the second term in the above
\begin{align}\label{eq:second_term_sigam}
     & ~ 2 a(x)_i^\top (A(x)^\top  A(x))^{-1} A(x)^\top \diag( A(x)_{*,j} ) A(x) (A(x)^\top  A(x))^{-1} a(x)_i \notag \\
    = & ~ 2 a(x)_i^\top ( A(x)^\top A(x) )^{-1} \cdot ( \sum_{l=1}^n a(x)_l A(x)_{l,j} a(x)_l^{\top} ) \cdot (A(x)^\top  A(x))^{-1} a(x)_i \notag \\
    = & ~ 2 \sum_{l=1}^n a(x)_i^\top ( A(x)^\top A(x) )^{-1} a(x)_l \cdot A(x)_{l,j} \cdot a(x)_l^{\top} (A(x)^\top  A(x))^{-1} a(x)_i \notag \\
    = & ~ 2 \sum_{l=1}^n \sigma(x)_{i,l} A(x)_{l,j} \sigma(x)_{i,l} \notag\\
    = & ~ 2 \langle \sigma(x)_{i,*} \circ \sigma(x)_{i,*}, A(x)_{*,j} \rangle
\end{align}
where the initial step is based on Fact~\ref{fac:rank_1_decomposition}, the second step is a result of straightforward algebraic manipulation, the 3rd step stems from $\sigma(x)$'s definition, and the final step arises from basic algebraic operations.

Putting Eq. \eqref{eq:first_term_sigma} and Eq. \eqref{eq:second_term_sigam} together, we have
\begin{align*}
    \frac{\d \sigma(x)_{i,i}}{\d x_j} = 2 \langle \sigma(x)_{i,*} \circ \sigma(x)_{i,*}, \circ A(x)_{*,j} \rangle - 2 A(x)_{i,j} \sigma(x)_{i,i}
\end{align*}

{\bf Proof of Part 4.}

This follows from re-grouping the entries in Part 3 in a diagonal matrix.

Thus, we have
\begin{align*}
    \frac{\d  \Sigma(x) }{\d x_j} 
    = & ~ 2 \diag(  Q(x) A(x)_{*,j} - \Sigma(x) A(x)_{*,j} ) \\
    = & ~ 2 \diag( (Q(x) - \Sigma(x) ) A(x)_{*,j} ),
\end{align*}
where the second step follows from simple algebra.

{\bf Proof of Part 5.}

We can show
\begin{align*}
      \frac{\d \sigma(x)_{i,l}}{\d x_j} = & ~ \frac{ \d a(x)_i^\top (A(x)^\top  A(x))^{-1} a(x)_l }{\d x_j } \\
      = & ~    \frac{ \d a(x)_i^\top }{\d x_j }  (A(x)^\top  A(x))^{-1} a(x)_l \\
      & + ~    a(x)_i^\top  x_j   (A(x)^\top  A(x))^{-1} \frac{\d a(x)_l}{\d x_j} \\
      & ~ + a(x)_i^\top \frac{\d (A(x)^\top  A(x))^{-1} }{\d x_j} a(x)_l \\
      = & ~  - A(x)_{i,j} a(x)_i^\top (A(x)^\top  A(x))^{-1} a(x)_l \\
      & - ~   A(x)_{l,j} a(x)_i^\top (A(x)^\top  A(x))^{-1} a(x)_l \\
      & ~ + a(x)_i^\top 2 (A(x)^\top  A(x))^{-1} A(x)^\top \diag( A(x)_{*,j} ) A(x) (A(x)^\top  A(x))^{-1} a(x)_l 
\end{align*}
where in the initial step, we utilize the definition of $\sigma(x)$, as described in Definition~\ref{def:sigma}, the subsequent step follows a basic rule of differentiation, and the third step relies on {\bf Part 1} and {\bf Part 2}.

Regarding the first term, we obtain:
\begin{align}\label{eq:first_term_sigma_il}
& ~ - A(x)_{i,j} a(x)_i^\top (A(x)^\top A(x))^{-1} a(x)_l \notag\\
= & ~ A(x)_{i,j}\sigma(x)_{i,l}
\end{align}
Here, this equation is derived from the definition of $\sigma(x)$, as outlined in Definition~\ref{def:sigma}.

Concerning the second term:
\begin{align}\label{eq:second_term_sigma_il}
& ~ - A(x)_{l,j} a(x)_i^\top (A(x)^\top A(x))^{-1} a(x)_l \notag\\
= & ~ - A(x)_{l,j}\sigma(x)_{i,l}
\end{align}
Here, the equation is based on the definition of $\sigma(x)$, as per Definition~\ref{def:sigma}.

For the third term, we have 
\begin{align}\label{eq:third_term_sigma_il}
  & ~ a(x)_i^\top 2 (A(x)^\top  A(x))^{-1} A(x)^\top \diag( A(x)_{*,j} ) A(x) (A(x)^\top  A(x))^{-1} a(x)_l  \notag\\
  = & ~ 2 a(x)_i^\top ( A(x)^\top A(x) )^{-1} \cdot ( \sum_{l=1}^n a(x)_l A(x)_{l,j} a(x)_l^{\top} ) \cdot (A(x)^\top  A(x))^{-1} a(x)_l \notag \\
    = & ~ 2 \sum_{l=1}^n a(x)_i^\top ( A(x)^\top A(x) )^{-1} a(x)_l \cdot A(x)_{l,j} \cdot a(x)_l^{\top} (A(x)^\top  A(x))^{-1} a(x)_l \notag \\
    = & ~ 2 \sum_{l=1}^n \sigma(x)_{i,l} A(x)_{l,j} \sigma(x)_{l,l} \notag \\
    = & ~ 2 \langle \sigma(x)_{i,*} \circ \sigma(x)_{l,*}, A(x)_{*,j} \rangle
\end{align}
where the initial step is derived from Fact~\ref{fac:rank_1_decomposition}, the second step results from basic algebraic manipulations, the third step stems from the definition of $\sigma(x)$ as provided in Definition~\ref{def:sigma}, and the final step arises from straightforward algebraic operations.

Putting Eq. \eqref{eq:first_term_sigma_il}, Eq. \eqref{eq:second_term_sigma_il} and Eq. \eqref{eq:third_term_sigma_il} together, we have
\begin{align*}
      \frac{\d \sigma(x)_{i,l}}{\d x_j} =  & ~ 2  \langle \sigma(x)_{*,i} \circ \sigma(x)_{*,l} , A(x)_{*,j} \rangle -  \sigma(x)_{i,l} \cdot (A(x)_{i,j} + A(x)_{l,j} )
\end{align*}

{\bf Proof of Part 6.}
\begin{align*}
    \frac{\d \sigma(x)}{\d x_j} 
    = & ~ 2 \sigma(x) \diag(A(x)_{*,j}) \sigma(x) - \sigma(x)  \diag(A(x)_{*,j})  - \diag(A(x)_{*,j})\sigma(x)
\end{align*}
To verify the correctness, you can just select the $(i,l)$-th entry of the above $n \times n$ matrix and see if it is equal to the result in Part 5.

{\bf Proof of Part 7.}
\begin{align*}
    \frac{\d \sigma_{*,i}(x)}{\d x_j} 
    = & ~ \frac{\d A(x)(A(x)^{\top} A(x))^{-1} a(x)_i}{\d x_j} \\
    = & ~ \frac{\d A(x)}{\d x_j} (A(x)^{\top} A(x))^{-1} a(x)_i\\
    & ~ + A(x)(A(x)^{\top} A(x))^{-1} \frac{\d a(x)_i}{\d x_j} \\
    & ~ + A(x)\frac{\d(A(x)^{\top} A(x))^{-1} }{ \d x_j} a(x)_i \\
    = & ~ - \diag(A(x)_{*,j}) A(x)(A(x)^{\top} A(x))^{-1} a(x)_i \\
    & ~ - A(x)(A(x)^{\top} A(x))^{-1}A(x)_{i,j} a(x)_i \\
    & ~ + 2A(x) (A(x)^\top  A(x))^{-1} A(x)^\top \diag( A(x)_{*,j} ) A(x) (A(x)^\top  A(x))^{-1}a(x)_i\\
    = & ~ 2\sigma(x)\diag(A(x)_{*,j})\sigma_{*,i}(x) - \diag(A(x)_{*,j})\sigma_{*,i}(x) - \sigma_{*,i}(x)A(x)_{i,j} 
\end{align*}
where the 1st step is based on the definition of $\sigma_{*,i}(x)$, the 2nd step follows from the basic derivative rule, the 3rd step stems from {\bf Part 1} of Lemma~\ref{lem:gradient_leverage} and {\bf Part 4} of Lemma~\ref{lem:basic_gradient}, and the final step arises from the definition of $\sigma_{*,i}(x)$.
\end{proof}

\section{Hessian}
\label{sec:hessian}

In this section, we computed the hessian for $L_{\exp}$ step by step. In Section~\ref{sec:hessian:sigma_entry}, we computed the hessian for a single entry of $\sigma(x)$. In Section~\ref{sec:hessian:sigma_row}, we computed the hessian for a row of $\sigma(x)$. In Section~\ref{sec:hessian:final}, we computed the hessian of $L_{\exp}$ and further decomposed it for the analysis of its convexity and lipschitz continuity. 

\subsection{Hessian for \texorpdfstring{$\sigma_{i,l}$}{}}\label{sec:hessian:sigma_entry}

\begin{lemma}\label{lem:sigma_i_l}

If we have:
\begin{itemize}
\item Define $A(x) \in \mathbb{R}^{n \times d}$ as specified in Definition~\ref{def:Ax}.
\item Define $\sigma(x) \in \mathbb{R}^{n \times n}$ according to Definition~\ref{def:sigma}.
\item Define $\Sigma(x) \in \mathbb{R}^{n \times n}$ as described in Definition~\ref{def:Sigma}.
\item Denote by $a(x)_i^\top$ the $i$-th row of $A(x)$.
\end{itemize}
Then, we have
\begin{itemize}
    \item {\bf Part 1.}
    \begin{align*}
        \frac{\d^2 \sigma(x)_{i,i}}{ \d x_k \d x_j } = C_1 + C_2 + C_3 + C_4 + C_5
    \end{align*}
    where
    \begin{itemize}
        \item $C_1 = + 8 \cdot \langle   \sigma(x) , ( A(x)_{*,j} \circ \sigma(x)_{*,i} ) \cdot ( A(x)_{*,k} \circ \sigma_{*,i}(x) )^\top   \rangle$
        \item $C_2 = - 6 \cdot \langle  \sigma(x)_{*,i} \circ \sigma(x)_{*,i} ,  A(x)_{*,k} \circ A(x)_{*,j} \rangle$
        \item $C_3 = - 4 \cdot \langle \sigma(x)_{*,i} \circ \sigma(x)_{*,i}, A(x)_{*,j} \rangle A(x)_{i,k}$
        \item $C_4 = - 4 \cdot \langle \sigma(x)_{*,i} \circ \sigma(x)_{*,i}, A(x)_{*,k} \rangle A(x)_{i,j}$
        \item $C_5 = + 6 \cdot \sigma(x)_{i,i} A(x)_{i,k} A(x)_{i,j}$
    \end{itemize}
    \item {\bf Part 2.}
    \begin{align*}
        \frac{\d^2 \sigma(x)_{i,l}}{ \d x_k \d x_j } = C_1 + C_2+ C_3+ C_4+ C_5 
    \end{align*}
    \begin{itemize}
        \item $C_1 = + 4 \cdot \langle   \sigma(x) , ( A(x)_{*,j} \circ \sigma(x)_{*,i} ) \cdot ( A(x)_{*,k} \circ \sigma_{*,l}(x) )^\top   \rangle + 4\langle   \sigma(x) , ( A(x)_{*,k} \circ \sigma(x)_{*,i} ) \cdot ( A(x)_{*,j} \circ \sigma_{*,l}(x) )^\top   \rangle$
        \item $C_2= -6 \cdot \langle  \sigma(x)_{*,i} \circ \sigma(x)_{*,l} ,  A(x)_{*,k} \circ A(x)_{*,j} \rangle$
        \item $C_3 = - 4 \cdot \langle \sigma(x)_{*,i} \circ \sigma(x)_{*,l}, A(x)_{*,j} \rangle A(x)_{i,k} - 4  \langle \sigma(x)_{*,i} \circ \sigma(x)_{*,l} , A(x)_{*,k} \rangle A(x)_{i,j}$
        \item $C_4= + 2 \cdot \sigma(x)_{i,l}  A(x)_{i,k}  A(x)_{i,j} + 2\sigma(x)_{i,l} A(x)_{l,k} A(x)_{i,j}$
        \item $C_5 = + \sigma(x)_{i,l} A(x)_{i,k} A(x)_{i,j} + \sigma(x)_{i,l} A(x)_{l,k} A(x)_{l,j}$
    \end{itemize} 
\end{itemize}
\end{lemma}
\begin{proof}
{\bf Proof of Part 1.}

First, we define:
\begin{align}
    B_1 = & ~ + 4\langle \frac{\d \sigma(x)_{*,i}  }{\d x_k} \circ \sigma(x)_{*,i}, A(x)_{*,j}  \rangle \label{eq:ij_part1_B1}\\
    B_2 = & ~ + 2 \langle  \sigma(x)_{*,i} \circ \sigma(x)_{*,i} , \frac{\d A(x)_{*,j}}{\d x_k} \rangle \label{eq:ij_part1_B2}\\
    B_3 = & ~ - 2 \frac{\d \sigma(x)_{i,i}}{\d x_k}A(x)_{i,j} \label{eq:ij_part1_B3}\\
    B_4 =  & ~ - 2 \sigma(x)_{i,i} \frac{\d A(x)_{i,j}}{\d x_k} \label{eq:ij_part1_B4}
\end{align} 

Now, we have
\begin{align*}
    \frac{\d^2 \sigma(x)_{i,i}}{ \d x_k \d x_j } =  & ~ \frac{\d}{\d x_k} \frac{\d \sigma(x)_{i,i}}{\d x_j} \\
    = & ~ \frac{\d}{\d x_k}  (2 \langle \sigma(x)_{*,i} \circ \sigma(x)_{*,i}, A(x)_{*,j} \rangle - 2 \sigma(x)_{i,i}  A(x)_{i,j} )\\
    = & ~ \frac{\d}{\d x_k}(2 \langle \sigma(x)_{*,i} \circ \sigma(x)_{*,i}, A(x)_{*,j} \rangle) - \frac{\d}{\d x_k}(2 \sigma(x)_{i,i}  A(x)_{i,j} ) \\
    = & ~ B_1 + B_2 + B_3 + B_4,
\end{align*}
The initial step arises from basic algebraic manipulation, the second step is due to {\bf Part 3} of Lemma~\ref{lem:gradient_leverage}, the third step stems from the sum rule of calculus, and the final step follows from the definition of $B_1$, $B_2$, $B_3$, and $B_4$ (see Eq.~\eqref{eq:ij_part1_B1}, Eq.~\eqref{eq:ij_part1_B2}, Eq.~\eqref{eq:ij_part1_B3}, and Eq.~\eqref{eq:ij_part1_B4}).

Now, we compute $B_1$, $B_2$, $B_3$, and $B_4$ separately. 

First, we compute $B_1$:
\begin{align*}
    B_1 = & ~ 4\langle \frac{\d \sigma(x)_{*,i}  }{\d x_k} \circ \sigma(x)_{*,i}, A(x)_{*,j}  \rangle \\
    = & ~ 4 \langle  (2 \sigma(x) \diag( A(x)_{*,k} ) \sigma_{*,i}(x) - \diag( A(x)_{*,k} ) \sigma_{*,i}(x) - \sigma(x)_{*,i} A(x)_{i,k}) \circ \sigma(x)_{*,i}, A(x)_{*,j} \rangle \\
    = & ~ + 8 \langle   \sigma(x) , ( A(x)_{*,j} \circ \sigma(x)_{*,i} ) \cdot ( A(x)_{*,k} \circ \sigma_{*,i}(x) )^\top   \rangle \\
    & ~ - 4 \langle  \sigma(x)_{*,i} \circ \sigma(x)_{*,i} ,  A(x)_{*,k} \circ A(x)_{*,j} \rangle \\
    & ~ - 4 \langle \sigma(x)_{*,i} \circ \sigma(x)_{*,i}, A(x)_{*,j} \rangle A(x)_{i,k}
\end{align*}
where the initial step is based on the definition of $B_1$, the 2nd step is by {\bf Part 7} of Lemma~\ref{lem:gradient_leverage}, and the third step derives from Fact~\ref{fac:circ_rule}.

Next, we compute $B_2$:
\begin{align*}
    B_2 = & ~ 2 \langle  \sigma(x)_{*,i} \circ \sigma(x)_{*,i} , \frac{\d A(x)_{*,j}}{\d x_k} \rangle \\
    = & ~ -2 \langle  \sigma(x)_{*,i} \circ \sigma(x)_{*,i} ,  A(x)_{*,k} \circ A(x)_{*,j}  \rangle
\end{align*}
where the initial step is derived from the definition of $B_2$, and the 2nd step is due to {\bf Part 8} of Lemma~\ref{lem:basic_gradient}.

Then, we compute $B_3$:
\begin{align*}
    B_3 = & ~ - 2 \frac{\d \sigma(x)_{i,i}}{\d x_k}A(x)_{i,j} \\
    = & ~ - 4 ( \langle \sigma(x)_{i,*} \circ \sigma(x)_{i,*}, A(x)_{*,k} \rangle -  A(x)_{i,k} \sigma(x)_{i,i}) A(x)_{i,j} \\
    = & ~ - 4  \langle \sigma(x)_{i,*} \circ \sigma(x)_{i,*}, A(x)_{*,k} \rangle A(x)_{i,j} +4  A(x)_{i,k} \sigma(x)_{i,i}A(x)_{i,j}
\end{align*}
where the 1st step comes from the definition of $B_3$, the 2nd step is through {\bf Part 3} of Lemma~\ref{lem:gradient_leverage}, and the last step arises from basic algebraic manipulation.

Finally, we compute $B_4$:
\begin{align*}
    B_4 =  & ~ - 2 \sigma(x)_{i,i} \frac{\d A(x)_{i,j}}{\d x_k} \\
    =  & ~ 2 \sigma(x)_{i,i} A(x)_{i,k} A(x)_{i,j}
\end{align*}
The initial step stems from the definition of $B_4$, and the 2nd step comes from {\bf Part 10} of Lemma~\ref{lem:basic_gradient}.

According to definition of $C$'s (from the Lemma Statement) and $B$'s (see Eq.~\eqref{eq:ij_part1_B1}, Eq.~\eqref{eq:ij_part1_B2}, Eq.~\eqref{eq:ij_part1_B3}, and Eq.~\eqref{eq:ij_part1_B4}), we know
\begin{align*}
C_1 =  &~ B_{1,1}  \\
C_2 = & ~B_{1,2} + B_2\\
C_3 = & ~ B_{1,3}  \\
C_4 = & ~ B_{3,1} \\
C_5 = & ~ B_{3,2}+B_4
\end{align*}
where $B_{1,1}$ denotes the first term in $B_1$, $B_{1,2}$ denotes the second term in $B_1$, $B_{3,1}$ denotes the first term in $B_3$, and $B_{3,2}$ denotes the second term in $B_3$.

{\bf Proof of Part 2.}

First, we define:
\begin{align}
    B_1 = & ~ 2  \langle  \frac{\d}{\d x_k} \sigma(x)_{*,i} \circ \sigma(x)_{*,l} , A(x)_{*,j} \rangle \label{eq:ij_part2_B1}\\
    B_2 = & ~ 2  \langle   \sigma(x)_{*,i} \circ \frac{\d}{\d x_k} \sigma(x)_{*,l} , A(x)_{*,j} \rangle \label{eq:ij_part2_B2}\\
    B_3 = & ~ 2  \langle   \sigma(x)_{*,i} \circ  \sigma(x)_{*,l} , \frac{\d}{\d x_k} A(x)_{*,j} \rangle \label{eq:ij_part2_B3}\\
    B_4 = & ~ - \frac{\d}{\d x_k} \sigma(x)_{i,l}A(x)_{i,j} \label{eq:ij_part2_B4}\\
    B_5 = & ~ - \sigma(x)_{i,l} \frac{\d}{\d x_k}A(x)_{i,j} \label{eq:ij_part2_B5}\\
    B_6 = & ~ - \frac{\d}{\d x_k} \sigma(x)_{i,l}A(x)_{l,j} \label{eq:ij_part2_B6}\\
    B_7 = & ~ - \sigma(x)_{i,l} \frac{\d}{\d x_k} A(x)_{l,j} \label{eq:ij_part2_B7}
\end{align}

Now, we have
\begin{align*}
     \frac{\d^2 \sigma(x)_{i,l}}{ \d x_k \d x_j } 
     =  & ~ \frac{\d}{\d x_k} \frac{\d \sigma(x)_{i,l}}{\d x_j} \\
    = & ~ \frac{\d}{\d x_k}   (2  \langle \sigma(x)_{*,i} \circ \sigma(x)_{*,l} , A(x)_{*,j} \rangle -  \sigma(x)_{i,l} \cdot (A(x)_{i,j} + A(x)_{l,j} ))\\
    =& ~ 2  \frac{\d}{\d x_k}  \langle \sigma(x)_{*,i} \circ \sigma(x)_{*,l} , A(x)_{*,j} \rangle - \frac{\d}{\d x_k} \sigma(x)_{i,l}A(x)_{i,j} - \frac{\d}{\d x_k} \sigma(x)_{i,l}A(x)_{l,j} \\
    = & ~ B_1 + B_2 + B_3 + B_4 + B_5 + B_6 + B_7,
\end{align*}
where the initial step arises from basic algebraic manipulation, the 2nd step arises from {\bf Part 5} of Lemma~\ref{lem:gradient_leverage}, the third step stems from the sum rule of calculus, and the final step follows from the definitions of $B_1$, $B_2$, $B_3$, $B_4$, $B_5$, $B_6$, and $B_7$ (see Eq.~\eqref{eq:ij_part2_B1}, Eq.~\eqref{eq:ij_part2_B2}, Eq.~\eqref{eq:ij_part2_B3}, Eq.~\eqref{eq:ij_part2_B4}, Eq.~\eqref{eq:ij_part2_B5}, Eq.~\eqref{eq:ij_part2_B6}, and Eq.~\eqref{eq:ij_part2_B7}).

Now, we compute $B_i, \forall i \in [6]$ separately, first we compute $B_1$:
\begin{align*}
    B_1 = & ~ 2  \langle  \frac{\d}{\d x_k} \sigma(x)_{*,i} \circ \sigma(x)_{*,l} , A(x)_{*,j} \rangle \\
    = & ~ 2  \langle  (2 \sigma(x) \diag( A(x)_{*,k} ) \sigma_{*,i}(x) - \diag( A(x)_{*,k} ) \sigma_{*,i}(x) - \sigma(x)_{*,i} A(x)_{i,k}) \circ \sigma(x)_{*,l} , A(x)_{*,j} \rangle \\
    = & ~ 4\langle   \sigma(x) , ( A(x)_{*,j} \circ \sigma(x)_{*,i} ) \cdot ( A(x)_{*,k} \circ \sigma_{*,l}(x) )^\top   \rangle   \\
    & ~ - 2 \langle  \sigma(x)_{*,i} \circ \sigma(x)_{*,l} ,  A(x)_{*,k} \circ A(x)_{*,j} \rangle \\
    & ~ - 2 \langle \sigma(x)_{*,i} \circ \sigma(x)_{*,l}, A(x)_{*,j} \rangle A(x)_{i,k}
\end{align*}
where the initial step arises from the definition of $B_1$, the subsequent one from {\bf Part 7} of Lemma~\ref{lem:gradient_leverage}, and the final one from Fact~\ref{fac:circ_rule}.

Next, we compute $B_2$:
\begin{align*}
    B_2 = & ~ 2  \langle \sigma(x)_{*,i} \circ \frac{\d}{\d x_k} \sigma(x)_{*,l} , A(x)_{*,j} \rangle \\
    = & ~ 2  \langle \sigma(x)_{*,i} \circ  (2 \sigma(x)\diag( A(x)_{*,k} )\sigma(x)_{*,l} - \diag(A(x)_{*,k})\sigma(x)_{*,l} - \sigma_{*,l}(x)A(x)_{i,k} , A(x)_{*,j} \rangle \\
    = & ~ 4\langle   \sigma(x) , ( A(x)_{*,k} \circ \sigma(x)_{*,i} ) \cdot ( A(x)_{*,j} \circ \sigma_{*,l}(x) )^\top   \rangle   \\
    & ~ - 2 \langle  \sigma(x)_{*,i} \circ \sigma(x)_{*,l} ,  A(x)_{*,k} \circ A(x)_{*,j} \rangle \\
    & ~ - 2 \langle \sigma(x)_{*,i} \circ \sigma(x)_{*,l}, A(x)_{*,j} \rangle A(x)_{i,k}
\end{align*}
where the initial step is derived from the definition of $B_2$, the subsequent one from {\bf Part 7} of Lemma~\ref{lem:gradient_leverage}, and the final one from Fact~\ref{fac:circ_rule}.

Next, we compute $B_3$:
\begin{align*}
    B_3 = & ~ 2  \langle   \sigma(x)_{*,i} \circ  \sigma(x)_{*,l} , \frac{\d}{\d x_k} A(x)_{*,j} \rangle \\
    = & ~ -2  \langle   \sigma(x)_{*,i} \circ  \sigma(x)_{*,l} , A(x)_{*,k} \circ A(x)_{*,j}  \rangle
\end{align*}
where the initial step is based on the definition of $B_3$, and the second step follows from {\bf Part 8} of Lemma~\ref{lem:basic_gradient}.

Next, we compute $B_4$:
\begin{align*}
    B_4 = & ~ - \frac{\d}{\d x_k} \sigma(x)_{i,l}A(x)_{i,j} \\
    = & ~ - (2  \langle \sigma(x)_{*,i} \circ \sigma(x)_{*,l} , A(x)_{*,k} \rangle -  \sigma(x)_{i,l} \cdot (A(x)_{i,k} + A(x)_{l,k} )) A(x)_{i,j} \\
    = & ~  - 2  \langle \sigma(x)_{*,i} \circ \sigma(x)_{*,l} , A(x)_{*,k} \rangle A(x)_{i,j}\\
    & ~ +  \sigma(x)_{i,l}  A(x)_{i,k}  A(x)_{i,j} \\
    & ~ + \sigma(x)_{i,l} A(x)_{l,k} A(x)_{i,j}
\end{align*}
where the 1st step is because of the definition of $B_4$,
the 2nd step comes from {\bf Part 5} of Lemma~\ref{lem:gradient_leverage}, and the 3rd step is due to simple algebra.

Next, we compute $B_5$:
\begin{align*}
    B_5 = & ~ - \sigma(x)_{i,l} \frac{\d}{\d x_k}A(x)_{i,j} \\
    = & ~ \sigma(x)_{i,l} A(x)_{i,k} A(x)_{i,j}
\end{align*}
where the first step follows from the definition of $B_5$,
the second step follows from {\bf Part 10} of Lemma~\ref{lem:basic_gradient}.

Next, we compute $B_6$:
\begin{align*}
    B_6 = & ~ - \frac{\d}{\d x_k} \sigma(x)_{i,l}A(x)_{l,j} \\
    = & ~ -(2  \langle \sigma(x)_{*,i} \circ \sigma(x)_{*,l} , A(x)_{*,k} \rangle -  \sigma(x)_{i,l} \cdot (A(x)_{i,k} + A(x)_{l,k} ))A(x)_{l,j} \\
    = & ~ - 2 \langle \sigma(x)_{*,i} \circ \sigma(x)_{*,l} , A(x)_{*,k} \rangle A(x)_{l,j} \\
    & ~ + \sigma(x)_{i,l} \cdot A(x)_{i,k}A(x)_{l,j}  \\
    & ~ + \sigma(x)_{i,l}A(x)_{l,k} A(x)_{l,j},
\end{align*}
where the initial step arises from the definition of $B_6$, the subsequent one from {\bf Part 5} of Lemma~\ref{lem:gradient_leverage}, and the final one from straightforward algebraic manipulations.

\begin{align*}
    B_7 = & ~ - \sigma(x)_{i,l} \frac{\d}{\d x_k} A(x)_{l,j} \\
    = & ~ \sigma(x)_{i,l} A(x)_{l,k} A(x)_{l,j}
\end{align*}
where the initial step stems from the definition of $B_7$, while the subsequent step is derived from {\bf Part 10} of Lemma~\ref{lem:basic_gradient}.

According to the definition of $B$'s (see Eq.~\eqref{eq:ij_part2_B1}, Eq.~\eqref{eq:ij_part2_B2}, Eq.~\eqref{eq:ij_part2_B3}, Eq.~\eqref{eq:ij_part2_B4}, Eq.~\eqref{eq:ij_part2_B5}, Eq.~\eqref{eq:ij_part2_B6}, and Eq.~\eqref{eq:ij_part2_B7}) and $C$'s (see from the Lemma statement), we know
\begin{align*}
        C_1 =& ~ B_{1,1} + B_{2,1}\\
        C_2 =& ~ B_{1,2} + B_{2,2}+B_3 \\
        C_3 =& ~ B_{1,3} + B_{2,3} + B_{4,1} + B_{6_1} \\
        C_4=& ~ B_{4,2} +B_{4,3} +B_{6,2} +B_{6,3} \\
        C_5 =& ~ B_5 + B_7
\end{align*}
where $B_{1,1}$ denotes the first term in $B_1$, $B_{1,2}$ denotes the second term in $B_1$,$B_{1,3}$ denotes the third term in $B_2$,$B_{2,1}$ denotes the first term in $B_2$, $B_{2,2}$ denotes the second term in $B_2$,$B_{2,3}$ denotes the third term in $B_2$, $B_{4,1}$ denotes the first term in $B_4$, $B_{4,2}$ denotes the second term in $B_4$,  $B_{6,1}$ denotes the first term in $B_6$, and $B_{6,2}$ denotes the second term in $B_6$,
\end{proof}

\subsection{Hessian of \texorpdfstring{$\sigma(x)_{*,i}$}{}}\label{sec:hessian:sigma_row}

\begin{lemma}\label{lem:sigma_*_i}

If we have
\begin{itemize}
\item Define $A(x) \in \mathbb{R}^{n \times d}$ as specified in Definition~\ref{def:Ax}.
\item Define $\sigma(x) \in \mathbb{R}^{n \times n}$ according to Definition~\ref{def:sigma}.
\item Define $\Sigma(x) \in \mathbb{R}^{n \times n}$ as described in Definition~\ref{def:Sigma}.
\item Denote by $a(x)_i^\top \in \mathbb{R}^{1 \times d}$ the $i$-th row of $A(x)$.
\end{itemize}
    Then, we have
    \begin{itemize}
        \item {\bf Part 1.}  Hessian 
        \begin{align*}
            \frac{\d^2 \sigma(x)_{*,i} }{ \d x_k \d x_j }
            = & ~ C_1 + C_2 + C_3 + C_4 + C_5 + C_6 + C_7 
        \end{align*}
        where
        \begin{itemize}
            \item $C_1 =+4 \sigma(x) \diag(A(x)_{*,k}) \sigma(x)\diag( A(x)_{*,j} ) \sigma_{*,i}(x) +  4 \sigma(x) \diag( A(x)_{*,j}) \sigma(x) \diag( A(x)_{*,k} ) \sigma_{*,i}(x) $
            \item $C_2 = - 6 \sigma(x)  \diag(A(x)_{*,k} \circ A(x)_{*,j}  )\sigma_{*,i}(x) $
            \item $C_3 = - 2 \diag(A(x)_{*,k})\sigma(x) \diag( A(x)_{*,j} ) \sigma_{*,i}(x)  - 2 \diag( A(x)_{*,j})\sigma(x) \diag( A(x)_{*,k} ) \sigma_{*,i}(x) $
            \item $C_4 = -2\sigma(x) \diag( A(x)_{*,j})\sigma(x)_{*,i} A(x)_{i,k} -2\sigma(x) \diag( A(x)_{*,k} ) \sigma_{*,i}(x)A(x)_{i,j}$
            \item  $C_5 = +2\diag( A(x)_{*,k} \circ A(x)_{*,j} ) \sigma_{*,i}(x)$
            \item  $C_6 = +\diag( A(x)_{*,j})\sigma(x)_{*,i} A(x)_{i,k} + \diag( A(x)_{*,k} ) \sigma_{*,i}(x)A(x)_{i,j}$
            \item  $C_7 = +2\sigma(x)_{*,i} A(x)_{i,k}A(x)_{i,j}$
        \end{itemize}
        \item {\bf Part 2.}  Hessian 
        \begin{align*}
            \frac{\d^2 \sigma(x) }{ \d x_k \d x_j }
            = & ~ C_1 + C_2 + C_3 + C_4 + C_5 + C_6 + C_7
        \end{align*}
        where 
        \begin{itemize}
            \item $C_1 =+4 \sigma(x) \diag(A(x)_{*,k}) \sigma(x)\diag(A(x)_{*,j}) \sigma(x) + 4\sigma(x)\diag(A(x)_{*,j})\sigma(x) \diag(A(x)_{*,k}) \sigma(x)  $
            \item $C_2 =-6 \sigma(x)  \diag(A(x)_{*,k} \circ A(x)_{*,j} ) \sigma(x)  $
            \item $C_3 =  -2 \diag(A(x)_{*,k})\sigma(x)\diag(A(x)_{*,j}) \sigma(x) - 2 \diag(A(x)_{*,j})\sigma(x) \diag(A(x)_{*,k}) \sigma(x)$
            \item $C_4 = - 2 \sigma(x)\diag(A(x)_{*,j})\sigma(x)  \diag(A(x)_{*,k})-2 \sigma(x) \diag(A(x)_{*,k}) \sigma(x)\diag(A(x)_{*,j})$
            \item $C_5 = +2\diag(A(x)_{*,k} \circ A(x)_{*,j}) \sigma(x)   $
            \item  $C_6 =+ 2\sigma(x)  \diag(A(x)_{*,k} \circ A(x)_{*,j})  $
            \item $C_7 = +\diag(A(x)_{*,j}) \sigma(x)  \diag(A(x)_{*,k}) +\diag(A(x)_{*,k})\sigma(x) \diag(A(x)_{*,j})$
        \end{itemize}
        
    \end{itemize}
\end{lemma}

\begin{proof}
{\bf Proof of Part 1.}

First, we define:
\begin{align}
    B_1 = & ~ 2\frac{\d \sigma(x)}{\d x_k
    }  \diag( A(x)_{*,j} ) \sigma_{*,i}(x) \label{eq:*j_part1_B1}\\
    B_2 = & ~2 \sigma(x)\frac{\d \diag( A(x)_{*,j} ) }{\d x_k}\sigma_{*,i}(x) \label{eq:*j_part1_B2}\\
    B_3 = & ~ 2 \sigma(x) \diag( A(x)_{*,j}) \frac{\d \sigma_{*,i}(x)}{\d x_k} \label{eq:*j_part1_B3}\\
    B_4 = & ~ - \frac{\d \diag( A(x)_{*,j}) }{\d x_k}\sigma_{*,i}(x) \label{eq:*j_part1_B4}\\
    B_5 = & ~ - \diag( A(x)_{*,j}) \frac{\d \sigma_{*,i}(x)}{\d x_k} \label{eq:*j_part1_B5}\\
    B_6 = & ~ - \frac{\d \sigma(x)_{*,i}}{\d x_k} A(x)_{i,j} \label{eq:*j_part1_B6}\\
    B_7 = & ~ - \sigma(x)_{*,i}\frac{\d A(x)_{i,j}}{\d x_k} \label{eq:*j_part1_B7}
\end{align}

Now, we have
\begin{align*}
    \frac{\d^2 \sigma(x)_{*,i} }{ \d x_k \d x_j }
    = & ~ \frac{\d}{\d x_k} \frac{\d \sigma(x)_{*,i}}{\d x_j } \\
    = & ~ \frac{\d}{\d x_k} (2 \sigma(x) \diag( A(x)_{*,j} ) \sigma_{*,i}(x) - \diag( A(x)_{*,j} ) \sigma_{*,i}(x) - \sigma(x)_{*,i} A(x)_{i,j}) \\
    = & ~ \frac{\d}{\d x_k}(2 \sigma(x) \diag( A(x)_{*,j} ) \sigma_{*,i}(x)) - \frac{\d}{\d x_k} (\diag( A(x)_{*,j}) \sigma_{*,i}(x) ) - \frac{\d}{\d x_k}(\sigma(x)_{*,i} A(x)_{i,j})\\
    = & ~ B_1 + B_2 + B_3 + B_4 + B_5 + B_6 + B_7  
\end{align*}
where the initial step arises from elementary algebra, the subsequent one from {\bf Part 7} of Lemma~\ref{lem:gradient_leverage}, the third step from the sum rule in calculus, and the final step from $B_1$, $B_2$, $B_3$, $B_4$, $B_5$, $B_6$, and $B_7$ (see Eq.~\eqref{eq:*j_part1_B1}, Eq.~\eqref{eq:*j_part1_B2}, Eq.~\eqref{eq:*j_part1_B3}, Eq.~\eqref{eq:*j_part1_B4}, Eq.~\eqref{eq:*j_part1_B5}, Eq.~\eqref{eq:*j_part1_B6}, and Eq.~\eqref{eq:*j_part1_B7}).

Now, we compute them separately.

Firstly, we compute $B_1$
\begin{align*}
     B_1 = & ~ 2\frac{\d \sigma(x)}{\d x_k
    }  \diag( A(x)_{*,j} ) \sigma_{*,i}(x) \\
    = & ~ 2 (2\sigma(x) \diag(A(x)_{*,k}) \sigma(x) - \sigma(x)  \diag(A(x)_{*,k})  - \diag(A(x)_{*,k})\sigma(x)) \diag( A(x)_{*,j} ) \sigma_{*,i}(x)\\
    = & ~ 4 \sigma(x) \diag(A(x)_{*,k}) \sigma(x)\diag( A(x)_{*,j} ) \sigma_{*,i}(x) \\
    & ~ - 2 \sigma(x)  \diag(A(x)_{*,k} \circ A(x)_{*,j}  )\sigma_{*,i}(x) \\
    & ~ - 2 \diag(A(x)_{*,k})\sigma(x) \diag( A(x)_{*,j} ) \sigma_{*,i}(x),
\end{align*}
where the initial step is based on the definition of $B_1$, the subsequent one on {\bf Part 6} of Lemma~\ref{lem:gradient_leverage}, and the final step on basic algebraic manipulation.

Then, we compute $B_2$
\begin{align*}
    B_2 
    = & ~ 2 \sigma(x)\frac{\d \diag( A(x)_{*,j} ) }{\d x_k}\sigma_{*,i}(x) \\
    = & ~ 2 \sigma(x) \diag(- A(x)_{*,k} \circ A(x)_{*,j} ) \sigma_{*,i}(x)\\
    = & ~ -2  \sigma(x) \diag( A(x)_{*,k} \circ A(x)_{*,j} ) \sigma_{*,i}(x)
\end{align*}
where the initial step arises from the definition of $B_2$, the subsequent one from {\bf Part 8} of Lemma~\ref{lem:basic_gradient}, and the final step from straightforward algebraic operations.

Next, we compute $B_3$
\begin{align*}
    B_3 = & ~ 2 \sigma(x) \diag( A(x)_{*,j}) \frac{\d \sigma_{*,i}(x)}{\d x_k} \\
    = & ~ 2 \sigma(x) \diag( A(x)_{*,j})\cdot (2 \sigma(x) \diag( A(x)_{*,k} ) \sigma_{*,i}(x) - \diag( A(x)_{*,k} ) \sigma_{*,i}(x) - \sigma(x)_{*,i} A(x)_{i,k}) \\
    = & ~ 4 \sigma(x) \diag( A(x)_{*,j}) \sigma(x) \diag( A(x)_{*,k} ) \sigma_{*,i}(x)\\
     & ~ -2 \sigma(x) \diag( A(x)_{*,j} \circ A(x)_{*,k} )\sigma_{*,i}(x) \\
     & ~ -2\sigma(x) \diag( A(x)_{*,j})\sigma(x)_{*,i} A(x)_{i,k}
\end{align*}
where the initial step is based on the definition of $B_3$, the subsequent one on {\bf Part 7} of Lemma~\ref{lem:gradient_leverage}, and the final step on basic algebraic manipulation.

Next, we compute $B_4$
\begin{align*}
    B_4 = & ~ - \frac{\d \diag( A(x)_{*,j}) }{\d x_k}\sigma_{*,i}(x)\\
    = & ~ \diag( A(x)_{*,k} \circ A(x)_{*,j} ) \sigma_{*,i}(x)
\end{align*}
where the initial step arises from the definition of $B_4$, while the subsequent step follows from {\bf Part 8} of Lemma~\ref{lem:basic_gradient}.

Next, we compute $B_5$
\begin{align*}
    B_5 = & ~ - \diag( A(x)_{*,j}) \frac{\d \sigma_{*,i}(x)}{\d x_k} \\ 
    = & ~ - \diag( A(x)_{*,j}) \cdot (2 \sigma(x) \diag( A(x)_{*,k} ) \sigma_{*,i}(x) - \diag( A(x)_{*,k} ) \sigma_{*,i}(x) - \sigma(x)_{*,i} A(x)_{i,k})\\
    = & ~ - 2 \diag( A(x)_{*,j})\sigma(x) \diag( A(x)_{*,k} ) \sigma_{*,i}(x) \\
    & ~ +  \diag( A(x)_{*,j} \circ A(x)_{*,k} ) \sigma_{*,i}(x)\\
    & ~ + \diag( A(x)_{*,j})\sigma(x)_{*,i} A(x)_{i,k}
\end{align*}
where the initial step is derived from the definition of $B_5$, the subsequent one from {\bf Part 7} of Lemma~\ref{lem:gradient_leverage}, and the final step from straightforward algebraic manipulation.

Next, we compute $B_6$
\begin{align*}
     B_6 = & ~ - \frac{\d \sigma(x)_{*,i}}{\d x_k} A(x)_{i,j}\\
     = & ~ - (2 \sigma(x) \diag( A(x)_{*,k} ) \sigma_{*,i}(x) - \diag( A(x)_{*,k} ) \sigma_{*,i}(x) - \sigma(x)_{*,i} A(x)_{i,k}) \cdot A(x)_{i,j} \\
     = & ~ -2\sigma(x) \diag( A(x)_{*,k} ) \sigma_{*,i}(x)A(x)_{i,j} \\
     & ~ +\diag( A(x)_{*,k} ) \sigma_{*,i}(x)A(x)_{i,j}\\
     & ~ + \sigma(x)_{*,i} A(x)_{i,k}A(x)_{i,j}
\end{align*}
where the initial step stems from the definition of $B_6$, the subsequent one from {\bf Part 7} of Lemma~\ref{lem:gradient_leverage}, and the final step from basic algebraic manipulation.

Next, we compute $B_7$
\begin{align*}
    B_7 = & ~ - \sigma(x)_{*,i}\frac{\d A(x)_{i,j}}{\d x_k}\\
    = & ~ \sigma(x)_{*,i}A(x)_{i,k} A(x)_{i,j}
\end{align*}
where the initial step arises from the definition of $B_3$, while the subsequent step is based on {\bf Part 10} of Lemma~\ref{lem:basic_gradient}.

From the definition of $C$'s (see from the Lemma statement) and $B$'s (see Eq.~\eqref{eq:*j_part1_B1}, Eq.~\eqref{eq:*j_part1_B2}, Eq.~\eqref{eq:*j_part1_B3}, Eq.~\eqref{eq:*j_part1_B4}, Eq.~\eqref{eq:*j_part1_B5}, Eq.~\eqref{eq:*j_part1_B6}, and Eq.~\eqref{eq:*j_part1_B7}), we know
\begin{align*}
            C_1 = & ~ B_{1,1} + B_{3,1}\\
            C_2 = & ~ B_{1,2} + B_2 + B_{3,2} \\
            C_3 = & ~ B_{1,3}+ B_{5,1}\\
            C_4 = & ~ B_{3,3} + B_{6,1}\\
            C_5 = & ~ B_{4} + B_{5,2}\\
            C_6 = & ~ B_{5,3} + B_{6,2}\\
            C_7 = & ~ B_{6,3} + B_{7}
        \end{align*}

{\bf Proof of Part 2.}

First, we define:
\begin{align}
    B_1 = & ~ 2\frac{\d \sigma(x)}{\d x_k} \diag(A(x)_{*,j}) \sigma(x) \label{eq:*j_part2_B1}\\
    B_2 = &  ~ 2\sigma(x)\frac{\d \diag(A(x)_{*,j})}{ \d x_k}\sigma(x) \label{eq:*j_part2_B2}\\
    B_3 = & ~ 2\sigma(x)\diag(A(x)_{*,j})  \frac{\d \sigma(x)}{\d x_k} \label{eq:*j_part2_B3}\\
    B_4 = & ~ -\frac{\d \sigma(x)}{\d x_k}  \diag(A(x)_{*,j}) \label{eq:*j_part2_B4}\\
    B_5 = & ~ -\sigma(x)\frac{\d \diag(A(x)_{*,j})}{\d x_k} \label{eq:*j_part2_B5}\\
    B_6 = & ~ -\frac{\d\diag(A(x)_{*,j})}{\d x_k} \sigma(x) \label{eq:*j_part2_B6}\\
    B_7 = & ~ -\diag(A(x)_{*,j})\frac{\d \sigma(x)}{\d x_k} \label{eq:*j_part2_B7}
\end{align}

Now, we have
\begin{align*}
    \frac{\d^2 \sigma(x) }{ \d x_k \d x_j }
    = & ~ \frac{\d}{\d x_k}\frac{\d \sigma(x) }{\d x_j }\\
    = & ~\frac{\d}{\d x_k} (2 \sigma(x) \diag(A(x)_{*,j}) \sigma(x) - \sigma(x)  \diag(A(x)_{*,j})  - \diag(A(x)_{*,j})\sigma(x))\\
    = & ~ \frac{\d}{\d x_k}(2\sigma(x) \diag(A(x)_{*,j}) \sigma(x)) - \frac{\d}{\d x_k}(\sigma(x)  \diag(A(x)_{*,j})) -\frac{\d}{\d x_k} (\diag(A(x)_{*,j})\sigma(x)) \\
    = & ~ B_1 + B_2 + B_3+ B_4+ B_5+ B_6+ B_7
\end{align*}
where the initial step arises from basic algebraic operations, the subsequent one from {\bf Part 6} of Lemma~\ref{lem:gradient_leverage}, the third step from the summation rule in calculus, and the final step from the definitions of $B_1$, $B_2$, $B_3$, $B_4$, $B_5$, $B_6$, and $B_7$ (see Eq.~\eqref{eq:*j_part2_B1}, Eq.~\eqref{eq:*j_part2_B2}, Eq.~\eqref{eq:*j_part2_B3}, Eq.~\eqref{eq:*j_part2_B4}, Eq.~\eqref{eq:*j_part2_B5}, Eq.~\eqref{eq:*j_part2_B6}, and Eq.~\eqref{eq:*j_part2_B7}).

Now, we compute $B_1$ 
\begin{align*}
    B_1 = & ~ 2\frac{\d \sigma(x)}{\d x_k} \diag(A(x)_{*,j}) \sigma(x) \\
    = & ~ 2(2 \sigma(x) \diag(A(x)_{*,k}) \sigma(x) - \sigma(x)  \diag(A(x)_{*,k})  - \diag(A(x)_{*,k})\sigma(x))\diag(A(x)_{*,j}) \sigma(x) \\
    = & ~ 4 \sigma(x) \diag(A(x)_{*,k}) \sigma(x)\diag(A(x)_{*,j}) \sigma(x) \\
    & ~ -2 \sigma(x)  \diag(A(x)_{*,k} \circ A(x)_{*,j} ) \sigma(x) \\
    & ~ -2 \diag(A(x)_{*,k})\sigma(x)\diag(A(x)_{*,j}) \sigma(x)
\end{align*}
where the initial step is derived from the definition of $B_1$, the subsequent one from {\bf Part 6} of Lemma~\ref{lem:gradient_leverage}, and the final step from basic algebraic manipulation.

Then, we compute $B_2$
\begin{align*}
    B_2 = &  ~ 2\sigma(x)\frac{\d \diag(A(x)_{*,j})}{ \d x_k}\sigma(x)\\
    = & ~ -2  \sigma(x) \diag( A(x)_{*,k} \circ A(x)_{*,j} ) \sigma(x)
\end{align*}
where the initial step arises from the definition of $B_2$, while the subsequent step is based on {\bf Part 8} of Lemma~\ref{lem:basic_gradient}.

Next, we compute $B_3$
\begin{align*}
    B_3 = & ~ 2\sigma(x)\diag(A(x)_{*,j})  \frac{\d \sigma(x)}{\d x_k} \\
    = & ~ 2\sigma(x)\diag(A(x)_{*,j})(2 \sigma(x) \diag(A(x)_{*,k}) \sigma(x) - \sigma(x)  \diag(A(x)_{*,k})  - \diag(A(x)_{*,k})\sigma(x)) \\
    = & ~ 4\sigma(x)\diag(A(x)_{*,j})\sigma(x) \diag(A(x)_{*,k}) \sigma(x) \\
    & ~ - 2 \sigma(x)\diag(A(x)_{*,j})\sigma(x)  \diag(A(x)_{*,k})\\
    & ~ - 2\sigma(x)\diag(A(x)_{*,j} \circ A(x)_{*,k})\sigma(x)
\end{align*}
where the initial step stems from the definition of $B_3$, the subsequent one from {\bf Part 6} of Lemma~\ref{lem:gradient_leverage}, and the final step from basic algebraic manipulation.

Next, we compute $B_4$
\begin{align*}
    B_4 = & ~ \frac{\d \sigma(x)}{\d x_k}  \diag(A(x)_{*,j})\\
    = & ~ -(2 \sigma(x) \diag(A(x)_{*,k}) \sigma(x) - \sigma(x)  \diag(A(x)_{*,k})  - \diag(A(x)_{*,k})\sigma(x) )\diag(A(x)_{*,j})\\
    = & ~ -2 \sigma(x) \diag(A(x)_{*,k}) \sigma(x)\diag(A(x)_{*,j}) \\
    & ~ +\sigma(x)  \diag(A(x)_{*,k} \circ A(x)_{*,j}) \\
    & ~ + \diag(A(x)_{*,k})\sigma(x) \diag(A(x)_{*,j})
\end{align*}
where the initial step is derived from the definition of $B_4$, the subsequent one from {\bf Part 6} of Lemma~\ref{lem:gradient_leverage}, and the final step from basic algebraic manipulation.

Next, we compute $B_5$
\begin{align*}
    B_5 = & ~ -\sigma(x)\frac{\d \diag(A(x)_{*,j})}{\d x_k} \\
    = & ~ \sigma(x) \diag(A(x)_{*,k} \circ A(x)_{*,j}) 
\end{align*}
where the initial step arises from the definition of $B_5$, while the subsequent step is based on {\bf Part 8} of Lemma~\ref{lem:basic_gradient}.

Next, we compute $B_6$
\begin{align*}
     B_6 = & ~ -\frac{\d\diag(A(x)_{*,j})}{\d x_k} \sigma(x) \\
     = & ~ \diag(A(x)_{*,k} \circ A(x)_{*,j}) \sigma(x) 
\end{align*}
where the initial step is derived from the definition of $B_6$, while the subsequent step is based on {\bf Part 8} of Lemma~\ref{lem:basic_gradient}.

Next, we compute $B_7$
\begin{align*}
    B_7 = & ~ -\diag(A(x)_{*,j})\frac{\d \sigma(x)}{\d x_k} \\
    = & ~  -\diag(A(x)_{*,j}) \cdot (2 \sigma(x) \diag(A(x)_{*,k}) \sigma(x) - \sigma(x)  \diag(A(x)_{*,k})  - \diag(A(x)_{*,k})\sigma(x)) \\
    = & ~ - 2 \diag(A(x)_{*,j})\sigma(x) \diag(A(x)_{*,k}) \sigma(x) \\
    & ~ + \diag(A(x)_{*,j}) \sigma(x)  \diag(A(x)_{*,k})\\
     & ~ + \diag(A(x)_{*,j} \circ A(x)_{*,k})\sigma(x)
\end{align*}
where the initial step arises from the definition of $B_7$, the subsequent one from {\bf Part 6} of Lemma~\ref{lem:gradient_leverage}, and the final step from basic algebraic manipulation.

From the definition of $B$'s (see Eq.~\eqref{eq:*j_part2_B1}, Eq.~\eqref{eq:*j_part2_B2}, Eq.~\eqref{eq:*j_part2_B3}, Eq.~\eqref{eq:*j_part2_B4}, Eq.~\eqref{eq:*j_part2_B5}, Eq.~\eqref{eq:*j_part2_B6}, and Eq.~\eqref{eq:*j_part2_B7}) and $C$'s (see from the Lemma statement), we know
\begin{align*}
            C_1 = & ~ B_{1,1} + B_{3,1}\\
            C_2 = & ~ B_{1,2} + B_2 + B_{3,3} \\
            C_3 = & ~ B_{1,3}+ B_{7,1}\\
            C_4 = & ~ B_{3,2} + B_{4,1}\\
            C_5 = & ~ B_{6} + B_{7,3}\\
            C_6 = & ~ B_{5} + B_{4,2}\\
            C_7 = & ~ B_{7,2} + B_{4,3}
        \end{align*}
\end{proof}


\subsection{Some short writeup}\label{sec:hessian:final}

\begin{lemma}[Formal version of Lemma~\ref{lem:hessian_decompose_informal}]\label{lem:hessian_decompose}

If we can get:
\begin{itemize}
\item Define $A(x) \in \mathbb{R}^{n \times d}$ as specified in Definition~\ref{def:Ax}.
\item Define $\sigma(x) \in \mathbb{R}^{n \times n}$ according to Definition~\ref{def:sigma}.
\item Define $\Sigma(x) \in \mathbb{R}^{n \times n}$ as described in Definition~\ref{def:Sigma}.
\item Denote by $a(x)_i^\top \in \mathbb{R}^{1 \times d}$ the $i$-th row of $A(x)$.
\item Let $H = \frac{\d^2 L_{\exp}(x) }{ \d x^2}$, where $L_{\exp} = L_c(x)$ is formulated as in Claim~\ref{cla:reformulation} and $x_k$ and $x_j$ are two arbitrary entries of $x \in \R^d$.
\end{itemize}

Then, we have 
\begin{align*}
   H_i  = D_{i, 1} + D_{i, 2} + D_{i, 3} + D_{i, 4} + D_{i, 5} + D_{i, 6}
\end{align*}

{\bf Part 1.} 

We show
\begin{itemize}
    \item $[D_{i, 1}]_{k,j} = 4\langle \sigma(x)_{*,i} \circ \sigma(x)_{*,i}, A(x)_{*,k} \rangle \langle \sigma(x)_{*,i} \circ \sigma(x)_{*,i}, A(x)_{*,j} \rangle$
    \item $[D_{i, 2}]_{k,j} = - 8 \sigma(x)_{i,i}\cdot \langle \sigma(x)_{*,i} \circ \sigma(x)_{*,i}, A(x)_{*,j} \rangle A(x)_{i,k}$
    \item $[D_{i, 3}]_{k,j} = - 8 \sigma(x)_{i,i}\cdot \langle \sigma(x)_{*,i} \circ \sigma(x)_{*,i}, A(x)_{*,k} \rangle A(x)_{i,j}$
    \item $[D_{i, 4}]_{k,j} = + 10 \sigma(x)_{i,i}\cdot \sigma(x)_{i,i} A(x)_{i,k} A(x)_{i,j}$
    \item $[D_{i, 5}]_{k,j}  = + 8 \sigma(x)_{i,i}\cdot \langle   \sigma(x) , ( A(x)_{*,j} \circ \sigma(x)_{*,i} ) \cdot ( A(x)_{*,k} \circ \sigma_{*,i}(x) )^\top   \rangle$
    \item $[D_{i, 6}]_{k,j}  = - 6 \sigma(x)_{i,i}\cdot \langle  \sigma(x)_{*,i} \circ \sigma(x)_{*,i} ,  A(x)_{*,k} \circ A(x)_{*,j} \rangle$
\end{itemize}

{\bf Part 2.} Further, we know
\begin{itemize}
    \item $D_{i, 1} = 4\underbrace{ A(x)^\top }_{d \times n} \underbrace{ \sigma(x)_{*,i}^{\circ 2} }_{n \times 1} \cdot \underbrace{ (\sigma(x)_{*,i}^{\circ 2})^\top }_{1 \times n} \underbrace{ A(x) }_{n \times d} $ 
    \item $D_{i, 2} = -8 \underbrace{ \sigma(x)_{i,i} }_{ \mathrm{scalar} } \underbrace{ ( A(x)_{i,*} )^\top }_{ d \times 1 } \underbrace{ (\sigma(x)_{*,i}^{\circ 2})^\top }_{1 \times n} \underbrace{A(x)}_{n \times d}$
    \item $D_{i, 3} = -8 \underbrace{ ( A(x)_{i,*} )^\top }_{d \times 1} \underbrace{ (\sigma(x)_{*,i} \circ \sigma(x)_{*,i})^{\top} }_{1 \times n} \sigma(x)_{i,i} \underbrace{ A(x) }_{n \times d}$ 
    \item $D_{i, 4} = 10  \underbrace{ ( A(x)_{i,*} )^\top }_{d \times 1} \sigma(x)_{i,i}^2 \underbrace{A(x)_{i,*}}_{1 \times d}$ 
    \item $D_{i, 5} = 8 \sigma(x)_{i,i} \underbrace{A(x)^\top}_{d \times n} \underbrace{\diag( \sigma_{*,i}(x) )}_{n \times n} \underbrace{\sigma(x)}_{n \times n} \underbrace{\diag( \sigma_{*,i}(x) )}_{n \times n} \underbrace{A(x)}_{n \times d}$
    \item $D_{i, 6} = -6 \underbrace{A(x)^{\top}}_{d \times n} \sigma(x)_{i,i}  \underbrace{\diag( \sigma(x)_{*,i}^{\circ 2})}_{n \times n} \underbrace{A(x)}_{n \times d}$  
\end{itemize}

\end{lemma}

\begin{proof}

{\bf Proof of Part 1.}

We have
\begin{align*}
H_{k,j} 
= & ~ 0.5 \frac{\d^2  (\sigma(x)_{i,i} - c_i)^2 }{ \d x_k \d x_j} \\
= & ~ \frac{\d \sigma(x)_{i,i}}{\d x_k} \frac{\d \sigma(x)_{i,i}}{\d x_j} + \sigma(x)_{i,i} \frac{\d^2 \sigma(x)_{i,i}}{\d x_k x_j}\\
= & ~ e_k^\top Q e_j,
\end{align*}
where the initial step is based on the definition of $H_{k,j}$ while the subsequent step follows from Fact~\ref{fac:exponential_der_rule}.

We define:
\begin{align*}
    B_{1} := & ~ \frac{\d \sigma(x)_{i,i}}{\d x_k} \frac{\d \sigma(x)_{i,i}}{\d x_j} \\
    B_2 := & ~  \sigma(x)_{i,i} \frac{\d^2 \sigma(x)_{i,i}}{\d x_k x_j}
\end{align*}
Let's compute $B_1$ first.
\begin{align*}
    B_1 
    = & ~ \frac{\d \sigma(x)_{i,i}}{\d x_k} \frac{\d \sigma(x)_{i,i}}{\d x_j} \\
    = & ~ (2 \langle \sigma(x)_{*,i} \circ \sigma(x)_{*,i}, A(x)_{*,k} \rangle - 2 \sigma(x)_{i,i}  A(x)_{i,k} ) (2 \langle \sigma(x)_{*,i} \circ \sigma(x)_{*,i}, A(x)_{*,j} \rangle - 2 \sigma(x)_{i,i}  A(x)_{i,j} )\\
    = & ~ 4\langle \sigma(x)_{*,i} \circ \sigma(x)_{*,i}, A(x)_{*,k} \rangle \langle \sigma(x)_{*,i} \circ \sigma(x)_{*,i}, A(x)_{*,j} \rangle \\
    & ~ - 4  \sigma(x)_{i,i}  A(x)_{i,k} \cdot \langle \sigma(x)_{*,i} \circ \sigma(x)_{*,i}, A(x)_{*,j} \rangle  \\
    & ~ -4  \langle \sigma(x)_{*,i} \circ \sigma(x)_{*,i}, A(x)_{*,k} \rangle \cdot \sigma(x)_{i,i}  A(x)_{i,j} ) \\
    & ~ + 4 \sigma(x)_{i,i}  A(x)_{i,k}\sigma(x)_{i,i}  A(x)_{i,j},
\end{align*}
where the initial step arises from the definition of $B_1$, the subsequent one from {\bf Part 3} of Lemma~\ref{lem:gradient_leverage}, and the final step from basic algebraic manipulation.

Then, we compute $B_2$
\begin{align*}
    B_2 
    = & ~  \sigma(x)_{i,i} \frac{\d^2 \sigma(x)_{i,i}}{\d x_k x_j} \\
    = & ~ C_1 + C_2 + C_3+ C_4 + C_5,
\end{align*}
where
\begin{itemize}
\item $C_1 = + 8 \sigma(x)_{i,i}\cdot \langle   \sigma(x) , ( A(x)_{*,j} \circ \sigma(x)_{*,i} ) \cdot ( A(x)_{*,k} \circ \sigma_{*,i}(x) )^\top   \rangle$
        \item $C_2 = - 6 \sigma(x)_{i,i}\cdot \langle  \sigma(x)_{*,i} \circ \sigma(x)_{*,i} ,  A(x)_{*,k} \circ A(x)_{*,j} \rangle$
        \item $C_3 = - 4 \sigma(x)_{i,i}\cdot \langle \sigma(x)_{*,i} \circ \sigma(x)_{*,i}, A(x)_{*,j} \rangle A(x)_{i,k}$
        \item $C_4 = - 4 \sigma(x)_{i,i}\cdot \langle \sigma(x)_{*,i} \circ \sigma(x)_{*,i}, A(x)_{*,k} \rangle A(x)_{i,j}$
        \item $C_5 = + 6 \sigma(x)_{i,i}\cdot \sigma(x)_{i,i} A(x)_{i,k} A(x)_{i,j}$
\end{itemize}

From the definition of $B,C$ and $D$, we have
\begin{align*}
    [D_1]_{k,j} = & ~ B_{1,1} = 4\langle \sigma(x)_{*,i} \circ \sigma(x)_{*,i}, A(x)_{*,k} \rangle \langle \sigma(x)_{*,i} \circ \sigma(x)_{*,i}, A(x)_{*,j} \rangle \\
    [D_2]_{k,j} = & ~ B_{1,2} + C_3 = - 8 \sigma(x)_{i,i}\cdot \langle \sigma(x)_{*,i} \circ \sigma(x)_{*,i}, A(x)_{*,j} \rangle A(x)_{i,k} \\
    [D_3]_{k,j} = & ~ B_{1,3} + C_4 = - 8 \sigma(x)_{i,i}\cdot \langle \sigma(x)_{*,i} \circ \sigma(x)_{*,i}, A(x)_{*,k} \rangle A(x)_{i,j}\\
    [D_4]_{k,j} = & ~ B_{1,4} + C_5 = + 10 \sigma(x)_{i,i}\cdot \sigma(x)_{i,i} A(x)_{i,k} A(x)_{i,j} \\
    [D_5]_{k,j} = & ~ C_1 = + 8 \sigma(x)_{i,i}\cdot \langle   \sigma(x) , ( A(x)_{*,j} \circ \sigma(x)_{*,i} ) \cdot ( A(x)_{*,k} \circ \sigma_{*,i}(x) )^\top   \rangle\\
    [D_6]_{k,j} = & ~ C_2 = - 6 \sigma(x)_{i,i}\cdot \langle  \sigma(x)_{*,i} \circ \sigma(x)_{*,i} ,  A(x)_{*,k} \circ A(x)_{*,j} \rangle
\end{align*}

{\bf Proof of Part 2.}

Let us consider about $D_1$, 
\begin{align*}
[D_1]_{k,j} = & ~ 4\langle \sigma(x)_{*,i} \circ \sigma(x)_{*,i}, A(x)_{*,k} \rangle \langle \sigma(x)_{*,i} \circ \sigma(x)_{*,i}, A(x)_{*,j} \rangle  \\
= & ~ 4 A(x)_{*,k}^\top \sigma(x)_{*,i}^{\circ 2} \cdot (\sigma(x)_{*,i}^{\circ 2})^\top A(x)_{*,j} \\
= & ~ 4 ( A(x) e_k)^\top  \sigma(x)_{*,i}^{\circ 2} \cdot (\sigma(x)_{*,i}^{\circ 2})^\top A(x) e_j  \\
= & ~ 4 e_k^\top A(x)^\top  \sigma(x)_{*,i}^{\circ 2} \cdot (\sigma(x)_{*,i}^{\circ 2})^\top A(x) e_j,
\end{align*}
where the first step follows from the definition of $[D_1]_{k,j}$, the 2nd step is by Fact~\ref{fac:circ_rule}, the 3rd step comes from vector decomposition, and the 4th step is due to simple algebra.

Thus, we have
\begin{align*}
    D_1 = 4\underbrace{ A(x)^\top }_{d \times n} \underbrace{ \sigma(x)_{*,i}^{\circ 2} }_{n \times 1} \cdot \underbrace{ (\sigma(x)_{*,i}^{\circ 2})^\top }_{1 \times n} \underbrace{ A(x) }_{n \times d}
\end{align*}

Next, we compute $D_2$:
\begin{align*}
    [D_2]_{k,j} = & ~ - 8  \sigma(x)_{i,i}  A(x)_{i,k} \cdot \langle \sigma(x)_{*,i} \circ \sigma(x)_{*,i}, A(x)_{*,j} \rangle \\
    = & ~ -8 A(x)_{i,k} \sigma(x)_{i,i} (\sigma(x)_{*,i} \circ \sigma(x)_{*,i})^\top A(x)_{*,j} \\
    = & ~ -8 \underbrace{ e_k^\top }_{1 \times d}  \underbrace{ \sigma(x)_{i,i} }_{ \mathrm{scalar} } \underbrace{ A(x)_{i,*}^{\top} }_{d \times 1}  \underbrace{ (\sigma(x)_{*,i}^{\circ 2} )^\top }_{1 \times n} \underbrace{ A(x) }_{n \times d} \underbrace{ e_j }_{d \times 1}
\end{align*}
where the initial step arises from the definition of $[D_2]_{k,j}$, the subsequent one from Fact~\ref{fac:circ_rule}, and the final step from vector decomposition.

Thus, we have
\begin{align*}
    D_2 = -8 \underbrace{ \sigma(x)_{i,i} }_{ \mathrm{scalar} } \underbrace{ A(x)_{i,*}^{\top} }_{ d \times 1 } \underbrace{ (\sigma(x)_{*,i}^{\circ 2})^\top }_{1 \times n} \underbrace{A(x)}_{n \times d}
\end{align*}

Similar to $D_2$, we have
\begin{align*}
    D_3 = -8 \underbrace{ (A(x)_{i,*})^{\top} }_{d \times 1} \underbrace{ (\sigma(x)_{*,i} \circ \sigma(x)_{*,i})^{\top} }_{1 \times n} \sigma(x)_{i,i} \underbrace{ A(x) }_{n \times d}
\end{align*}

Next, we compute $D_4$:
\begin{align*}
    [D_4]_{k,j} = & ~ 10 \sigma(x)_{i,i}  A(x)_{i,k}\sigma(x)_{i,i}  A(x)_{i,j} \\
    = & ~ 10  A(x)_{i,k} \sigma(x)_{i,i}^2 A(x)_{i,j} \\
    = & ~ 10 \underbrace{e_k^\top}_{1 \times d}  \underbrace{A(x)_{i,*}^{\top}}_{d \times 1} \sigma(x)_{i,i}^2 \underbrace{A(x)_{i,*}}_{1 \times d} \underbrace{e_j}_{d \times 1}
\end{align*}
where the first step follows from the definition of $[D_4]_{k,j}$, the second step follows from the simple algebra, and the last step follows from the vector decomposition.

Thus, we have
\begin{align*}
    D_4 = 10  \underbrace{A(x)_{i,*}^{\top}}_{d \times 1} \sigma(x)_{i,i}^2 \underbrace{A(x)_{i,*}}_{1 \times d}
\end{align*}

Next, we compute $D_5$:
\begin{align*}
    [D_5]_{k,j} = & ~ 8 \sigma(x)_{i,i}\cdot \langle   \sigma(x) , ( A(x)_{*,j} \circ \sigma(x)_{*,i} ) \cdot ( A(x)_{*,k} \circ \sigma_{*,i}(x) )^\top   \rangle \\
    = & ~ 8 \sigma(x)_{i,i} (A(x)_{*,k} \circ \sigma_{*,i}(x))^\top \sigma(x) ( A(x)_{*,j} \circ \sigma(x)_{*,i} )  \\
    = & ~ 8 \sigma(x)_{i,i} (A(x)_{*,k})^\top \diag( \sigma_{*,i}(x) ) \sigma(x) \diag( \sigma(x)_{*,i} ) A(x)_{*,j}    \\
    = & ~ 8 \sigma(x)_{i,i} \underbrace{e_k^\top}_{1 \times d} \underbrace{A(x)^\top}_{d \times n} \underbrace{\diag( \sigma_{*,i}(x) )}_{n \times n} \underbrace{\sigma(x)}_{n \times n} \underbrace{\diag( \sigma_{*,i}(x) )}_{n \times n} \underbrace{A(x)}_{n \times d} \underbrace{e_j}_{d \times 1}
\end{align*}
where the initial step arises from the definition of $[D_5]_{k,j}$, the subsequent ones from Fact~\ref{fac:circ_rule}, and the final step from vector decomposition.

Thus, we have
\begin{align*}
    D_5 = 8 \sigma(x)_{i,i} \underbrace{A(x)^\top}_{d \times n} \underbrace{\diag( \sigma_{*,i}(x) )}_{n \times n} \underbrace{\sigma(x)}_{n \times n} \underbrace{\diag( \sigma_{*,i}(x) )}_{n \times n} \underbrace{A(x)}_{n \times d}
\end{align*}

Finally, we compute $D_6$:
\begin{align*}
    [D_6]_{k,j} = & ~ - 6 \sigma(x)_{i,i}\cdot \langle  \sigma(x)_{*,i} \circ \sigma(x)_{*,i} ,  A(x)_{*,k} \circ A(x)_{*,j} \rangle \\
    = & ~ - 6 \sigma(x)_{i,i}\cdot \langle  (\sigma(x)_{*,i} \circ \sigma(x)_{*,i}) \circ A(x)_{*,k} , A(x)_{*,j} \rangle \\
    = & ~ - 6 \sigma(x)_{i,i}\cdot \langle  \diag(\sigma(x)_{*,i} \circ \sigma(x)_{*,i})A(x)_{*,k} , A(x)_{*,j} \rangle \\
    = & ~ - 6 A(x)_{*,k}^\top \sigma(x)_{i,i} \diag(\sigma(x)_{*,i} \circ \sigma(x)_{*,i})A(x)_{*,j} \\
    = & ~ -6 \underbrace{e_k^\top}_{1 \times d} \underbrace{A(x)^{\top}}_{d \times n} \sigma(x)_{i,i} \underbrace{\diag( \sigma(x)_{*,i}^{\circ 2})}_{n \times n} \underbrace{A(x)}_{n \times d} \underbrace{e_j}_{d \times 1}
\end{align*}
where the initial step arises from the definition of $[D_6]_{k,j}$, the subsequent ones from Fact~\ref{fac:circ_rule}, and the final step from vector decomposition.

Thus, we have
\begin{align*}
    D_6 = -6 \underbrace{A(x)^{\top}}_{d \times n} \sigma(x)_{i,i}  \underbrace{\diag( \sigma(x)_{*,i}^{\circ 2})}_{n \times n} \underbrace{A(x)}_{n \times d}
\end{align*}
\end{proof}

\begin{fact}\label{fac:hessian_decomposition_final}
    If we have:
    \begin{itemize}
        \item Let $D_i, i \in [6]$ be defined in Definition~\ref{def:further_decomposition}
        \item Let $H(x) = \frac{\d^2 0.5 \sum_{i=1}^n (\sigma_{i,i}(x) - c_i )^2 }{ \d x^2}$
    \end{itemize}
    Then we have
    \begin{align*}
        H(x) = A(x)^\top (\sum_{i=1}^6 D_i) A(x)
    \end{align*}
\end{fact}

\section{Hessian is Positive Definite}\label{sec:psd}

In this section, we are able to show that $L$ is convex. In Section~\ref{sec:psd:norm_bounds}, we provide norm bounds for some basic terms. In Section~\ref{sec:psd:entry_psd}, we leverage the tools from the previous section and prove that a large part of $\nabla^2 L_{\exp}$ is positive definite. In Section~\ref{sec:psd:final}, we proved that $\nabla^2 L$ is positive definite, and thus the loss function for the inverting leverage score problem is convex.

\subsection{Norm Bounds for Basic Terms}\label{sec:psd:norm_bounds}

\begin{lemma}\label{lem:norm_bounds}

If we have:
\begin{itemize}
\item The spectral norm of $A$ is bounded by $R$.
\item The $\ell_2$ norm of $x$ is bounded by $R$.
\item $S(x)$ is defined as per Definition~\ref{def:s}.
\item $\sigma(x)$ is defined according to Definition~\ref{def:sigma}.
\item The $\sigma_{\min}(A(x))$ is greater than or equal to $\beta$.
\end{itemize}
    Then we have
    \begin{itemize}
        \item Part 1. $\|\sigma(x)\| \leq 1$
        \item Part 2. $|\sigma_{i,i}(x)| \leq 1$
        \item Part 3. $\|\sigma_{*,i}(x)\|_2 \leq 1$
        \item Part 4. $\|a(x)_i^\top\|_2 \leq \beta R$
        \item Part 5. $\|A(x)^{-1}\| \leq \beta^{-1}$
        \item Part 6. $\|(A(x)^\top A(x))^{-1}\| \leq \beta^{-2}$

    \end{itemize}
\end{lemma}
\begin{proof}
    {\bf Proof of Part 1.}

    If a matrix $P$ satisfy that $P^2 = P$, then $P$ is called a projection matrix.

    By property of projection matrix, we know that $\| P \| \leq 1$. 

    Since $\sigma(x) \sigma(x) = \sigma(x)$, thus $\sigma(x)$ is a projection matrix, thus, we have $\| \sigma(x) \| \leq 1$.

    {\bf Proof of Part 2.}
    
    It follows from Part 2 directly.

    {\bf Proof of Part 3.}
    
    It follows from Part 1 directly.

    {\bf Proof of Part 4.}
    It's apparent that $\|A(x)\| \leq \beta R$, and thus it follows from the definition of $a(x)_i^\top$ that $\|a(x)_i^\top\|_2 \leq \beta R$.

    {\bf Proof of Part 5.}

    We can show
    \begin{align*}
        \| A(x)^{-1} \| = \sigma_{\min}(A(x))^{-1} \leq \beta^{-1}
    \end{align*}

    {\bf Proof of Part 6}

    \begin{align*}
        \| (A(x)^\top A(x))^{-1} \| = \sigma_{\min}(A(x))^{-2} \leq \beta^{-2}
    \end{align*}
\end{proof}

\subsection{Rewrite the Hessian}\label{sec:psd:entry_psd}

\begin{lemma}\label{lem:psd_for_di}
    If we have:
    \begin{itemize}
        \item Let $D_i, i \in [6]$ be defined in Definition~\ref{def:further_decomposition}
        \item Let $\|A\| \leq R$
        \item Let $S(x)$ be defined in Definition~\ref{def:s}
        \item Let $\|S(x)\| \geq \beta$ where $\beta \in (0, 0.1)$ 
    \end{itemize}
    Then we have
    \begin{itemize}
        \item $-4I_d \preceq D_1 \preceq 4I_d$
        \item $-8 I_d \preceq D_2 \preceq 8 I_d$
        \item $-8 I_d \preceq D_3 \preceq 8 I_d$
        \item $-10 I_d \preceq D_4 \preceq 10 I_d$
        \item $-8 I_d \preceq D_5 \preceq 8 I_d$
        \item $-6 I_d \preceq D_6 \preceq 5 I_d$
    \end{itemize}
\end{lemma}
\begin{proof}
    These trivially follows from applying Lemma~\ref{lem:norm_bounds}, Fact~\ref{fac:vector_norm} and Fact~\ref{fac:matrix_norm}.
\end{proof}

\begin{fact}\label{fac:hessian_reform}
    If we have:
    \begin{itemize}
        \item  Let $\frac{d^2 L}{\d x^2}$ be computed in Lemma~\ref{lem:hessian_decompose}
        \item Let $D_i(x), i \in [6]$ be defined in Definition~\ref{def:further_decomposition}
    \end{itemize}
    
    Then, $\frac{d^2 L}{\d x^2}$ can be reformulated in the form of
    \begin{align*}
        \frac{d^2 L}{\d x^2} = A^\top G(x) A
    \end{align*}
    where 
    \begin{align*}
        G(x) = S(x)^{-1} \sum_{i=1}^6 D_i(x) S(x)^{-1}
    \end{align*}
\end{fact}

\begin{lemma}\label{lem:double_bound_on_hessian}
    If we have:
    \begin{itemize}
        \item Let $D_i, i \in [6]$ be defined in Definition~\ref{def:further_decomposition}
        \item Let $H = \frac{\d^2 0.5 \sum_{i=1}^n (\sigma_{i,i}(x) - c_i )^2 }{ \d x^2}$
        \item Let $\|A\| \leq R$

\item Define $A(x)$ as specified in Definition~\ref{def:Ax}.
\item Define $S(x)$ according to Definition~\ref{def:Ax}.
\item Ensure that $\|S(x)\|$ is greater than or equal to $\beta$, where $\beta$ in $(0, 0.1)$.
        \item Let $G(x)$ be given in Fact~\ref{fac:hessian_reform}
    \end{itemize}
    Then we have
    \begin{align*}
        -44 \beta I_d \preceq G(x) \preceq 44 \beta I_d
    \end{align*}
    This trivially follows from Lemma~\ref{lem:psd_for_di}, the definition of $A(x)$, and the assumption that $S(x) \geq \beta$.
\end{lemma}

\subsection{Hessian is Positive Definite}\label{sec:psd:final}

\begin{lemma}[Formal version of Lemma~\ref{lem:convex_informal}]\label{lem:convex}
If we have:
\begin{itemize}
    \item Given matrix $A \in \R^{n \times d}$.
    \item Let $L_{\exp}(x) = 0.5 \sum_{i=1}^n (\sigma(x)_{i,i} - c_i)^2$.
    \item Let $L_{\reg}(x)$ be defined as Definition~\ref{def:L_reg}.
    \item Let $L(x) = L_{\exp}(x) + L_{\reg}(x)$. 
    \item Let $W = \diag(w) \in \R^{n \times n}$. Let $W^2 \in \R^{n \times n}$ denote the matrix that $i$-th diagonal entry is $w_{i,i}^2$.
    \item Let $\sigma_{\min}(A)$ denote the minimum singular value of $A$.
    \item Let $l > 0$ denote a scalar.
    \item Let $w_{i}^2 \geq -44 \beta + l/\sigma_{\min}(A)^2$
\end{itemize}
Then, we have
    \begin{align*}
    \frac{\d^2 L}{\d x^2} \succeq l \cdot I_d
    \end{align*}
\end{lemma}

\begin{proof}
    By applying Lemma~\ref{lem:double_bound_on_hessian} we have
    \begin{align*}
        \frac{\d^2 L_{\exp}}{\d x^2} \succeq -44 \beta I_n
    \end{align*}
    
    Also, it's trivial that
    \begin{align*}
        \frac{\d^2 L}{\d x^2} 
        = & ~ \frac{\d^2 L_{\exp}}{\d x^2} + \frac{\d^2 L_{\reg}
        }{\d x^2} \\
        = & ~ A^\top G(x) A + A^\top W^2 A \\
        = & ~ A^\top (G(x) + W^2) A
    \end{align*}

    Then we can write $\frac{\d^2 L}{\d x^2}$ as
    \begin{align*}
        \frac{\d^2 L}{\d x^2} = A^\top D A
    \end{align*}
    where
    \begin{align*}
        D = G(x) + W^2
    \end{align*}

    We can then bound $D$ as follows
    \begin{align*}
        D
        \succeq & ~ - 44 \beta I_n + w_{\min}^2 I_n \\
        = & ~ ( - 44 \beta + w_{\min}^2) I_n \\ 
        \succeq & ~ \frac{l}{\sigma_{\min}(A)^2} I_n
    \end{align*}
    where the initial step stems from Lemma~\ref{lem:double_bound_on_hessian}, the subsequent step arises from basic algebra, and the third step results from the assumption that $w_{i}^2 \geq -44 \beta + l/\sigma_{\min}(A)^2$.

    Since $D$ is positive definite, then we have 
    \begin{align*}
        A^\top D A \succeq \sigma_{\min}(D) \cdot \sigma_{\min}(A)^2 I_d \succeq l \cdot I_d
    \end{align*}
    Thus, Hessian is positive definite forever and thus $L$ is convex.
\end{proof}

\section{Lipschitz for Hessian}
\label{sec:Lipschitz_app}

In this section, we proved that the hessian of the loss function for Inverting Leverage Score problem is lipschitz continuous. In Section~\ref{sec:lipschitz_app:basic}, we proved that some basic terms is lipschitz continuous. In Section~\ref{sec:lipschitz_app:steps}, we leverage the decomposition of $\nabla^2 L_{\exp}$ to prove that $\nabla^2 L_{\exp}$ step by step.

\subsection{Lipschitz Continuity for Basic Functions}\label{sec:lipschitz_app:basic}

\begin{lemma}\label{lem:basic_lips}
    If we have:
    \begin{itemize}
        \item Let $\|A\| \leq R$
        \item Let $\|A(x - \wh{x})\| \leq 0.01$
        \item Let $S(x)$ be defined in Definition~\ref{def:s}
        \item Let $\|S(x)\| \geq \beta$ where $\beta \in (0, 0.1)$ 
        \item Define $A(x)$ as described in Definition~\ref{def:Ax}.
        \item Define $\sigma(x)$ according to Definition~\ref{def:sigma}.
    \end{itemize}
    Then we have
    \begin{itemize}
        \item {\bf Part 1.} $\|S(x) - S(\wh{x})\| \leq R \|x - \wh{x}\|_2$
        \item {\bf Part 2.} $\|S(x)^{-1} - S(\wh{x})^{-1}\| \leq \beta^{-2} R \|x - \wh{x}\|_2$
        \item {\bf Part 3.} $\|A(x) - A(\wh{x})\| \leq \beta^{-2} R^2 \|x - \wh{x}\|_2$
        \item {\bf Part 4.} $\|A(x)^{-1} - A(\wh{x})^{-1}\| \leq \beta^{-4} R^2 \|x - \wh{x}\|_2$
        \item {\bf Part 5.} $\| (A(x)^{\top } A(x) )^{-1} - (A(\wh{x})^{\top } A(\wh{x}) )^{-1}\| \leq 2\beta^{-5} R^2 \|x - \wh{x}\|_2$
        \item {\bf Part 6.} $\|\sigma(x) - \sigma(\wh{x})\| \leq 3\beta^{-7} R^3 \|x - \wh{x}\|_2$
        \item {\bf Part 7.} $|\sigma_{i,i}(x) - \sigma_{i,i}(\wh{x})| \leq  \| \sigma(x) - \sigma( \wh{x} ) \| \leq 3\beta^{-7} R^3 \|x - \wh{x}\|_2$
        \item {\bf Part 8.} $\|\sigma(x)_{*,i} - \sigma(\wh{x})_{*,i}\|_2 \leq \| \sigma(x) - \sigma( \wh{x} ) \| \leq 3\beta^{-7} R^3 \|x - \wh{x}\|_2$
    \end{itemize}
\end{lemma}
\begin{proof}
    {\bf Proof of Part 1}
    \begin{align*}
        \|S(x) - S(\wh{x})\|
        = & ~ \|A(x - \wh{x})\| \\
        \leq & ~ \|A\| \|x - \wh{x}\|_2 \\
        \leq & ~ R \|x - \wh{x}\|_2
    \end{align*}
    where the initial step arises from the definition of $S(x)$,
the subsequent step stems from Fact~\ref{fac:vector_norm},
and the final step results from the assumption that $\|A\| \leq R$.

    {\bf Proof of Part 2}
    \begin{align*}
        \|S(x)^{-1} - S(\wh{x})^{-1}\|
        = & ~ \|S(x)^{-1}\|\|\|S(\wh{x})^{-1}\| \|S(x) - S(\wh{x})\| \\
        \leq & ~ \beta^{-2} R \|x - \wh{x}\|_2
    \end{align*}
    where the 1st step is due to basic algebra,
    the 2nd step comes from the assumption that $S(x) \geq \beta$ and {\bf Part 1}.
    
    {\bf Proof of Part 3}
    \begin{align*}
        \|A(x) - A(\wh{x})\|
        = & ~ \|A(S(x)^{-1} - S(\wh{x}))\| \\
        \leq & ~ \|A\| \|S(x)^{-1} - S(\wh{x})^{-1}\| \\
        \leq & ~ \beta^{-2} R^2 \|x - \wh{x}\|_2
    \end{align*}
    where the initial step stems from the definition of $A(x)$, the 2nd step is from Fact~\ref{fac:vector_norm}, and the final step is based on the assumption that $|A| \leq R$ and {\bf Part 2}.

    {\bf Proof of Part 4}
    \begin{align*}
        \|A(x)^{-1} - A(\wh{x})^{-1}\|
        = & ~ \|(A(x)^{-1}A(\wh{x})^{-1}) A(x) - A(\wh{x}) \|  \\
        \leq & ~ \| A(x)^{-1}\|^2 \|A(x) -A(\wh{x}) \| \\
        \leq & ~ \beta^{-4} R^2 \|x - \wh{x}\|_2
    \end{align*}
    where the first step follows from simple algebra,
    the second step follows from Fact~\ref{fac:matrix_norm},
    the last step follows from {\bf Part 3} and {\bf Part 5} of Lemma~\ref{lem:norm_bounds}.

    {\bf Proof of Part 5}
    \begin{align*}
        \| (A(x)^{\top } A(x) )^{-1} - (A(\wh{x})^{\top } A(\wh{x}) )^{-1}\| 
        = & ~ \|A(x)^{-1} A(x)^{-\top} - A(\wh{x})^{-1} A(\wh{x})^{-\top}\|\\
        \leq & ~ \|A(x)^{-1} - A(\wh{x})^{-1}\| \|A(x)^{-\top} \| \\
        & ~ + \| A(\wh{x})^{-1}\| \| A(x)^{-\top} - A(\wh{x})^{-\top}\| \\
        \leq & ~ 2\beta^{-5} R^2 \|x - \wh{x}\|_2
    \end{align*}
    where the initial steps arise from straightforward algebraic manipulation, the subsequent one follows from basic algebra, the 3rd step stems from Fact~\ref{fac:matrix_norm}, and the final step is based on {\bf Part 4} and {\bf Part 5} of Lemma~\ref{lem:norm_bounds}.

    {\bf Proof of Part 6}
    \begin{align*}
        \|\sigma(x) - \sigma(\wh{x})\|
        = & ~ \|A(x)( A(x)^\top A(x) )^{-1} A(x)^\top - A(\wh{x})( A(\wh{x})^\top A(\wh{x}) )^{-1} A(\wh{x})^\top\|
    \end{align*}

    For simplicity, we define $T(x) := ( A(x)^\top A(x) )^{-1}$, then we have
    \begin{align*}
        \|\sigma(x) - \sigma(\wh{x})\| = \|A(x)T(x)A(x)^\top - A(\wh{x})T(\wh{x})A A(\wh{x})^\top\|
    \end{align*}

    Define
    \begin{align*}
        G_1 : & = A(x)T(x)A(x)^\top - A(x)T(x)A(\wh{x})^\top \\
        G_2 : & = A(x)T(x)A(\wh{x})^\top - A(x)T(\wh{x})A(\wh{x})^\top \\
        G_3 : & = A(x)T(\wh{x})A(\wh{x})^\top - A(\wh{x})T(\wh{x})A(\wh{x})^\top
    \end{align*}

    First, upper bound $\|G_1\|$:
    \begin{align*}
        \|G_1\|
        = & ~ \|A(x)T(x)A(x)^\top - A(x)T(x)A(\wh{x})^\top\| \\
        \leq & ~ \|A(x)\|\|T(x)\| \|A(x) - A(\wh{x})\| \\
        \leq & ~ \beta^{-5} R^3 \|x - \wh{x}\|_2 
    \end{align*}
    where the first step follows from the definition of $G_1$,
    the second step follows from Fact~\ref{fac:matrix_norm},
    the last step follows from {\bf Part 6} of Lemma~\ref{lem:norm_bounds}, {\bf Part 3} and the bound of $\|A(x)\|$.

    By symmetry we have
    \begin{align*}
        \|G_3\| \leq & ~ \beta^{-5} R^3 \|x - \wh{x}\|_2 
    \end{align*}

    Next, we upper bound $\|G_2\|$:
    \begin{align*}
        \|G_2\|
        = & ~  \|A(x)T(x)A(\wh{x})^\top - A(x)T(\wh{x})A(\wh{x})^\top\| \\
        \leq & ~ \|A(x)\|^2 \|T(x) - T(\wh{x})\| \\
        \leq & ~ 2\beta^{-7} R^2 \|x - \wh{x}\|_2
    \end{align*}
    where the first step follows from the definition of $G_2$,
    the second step follows from Fact~\ref{fac:matrix_norm},
    the last step follows from {\bf Part 5} and the bound of $\|A(x)\|$.

    By combining the bounds of $\|G_i\|$, we have
    \begin{align*}
        \|\sigma(x) - \sigma(\wh{x})\| \leq & ~ 3\beta^{-7} R^3 \|x - \wh{x}\|_2
    \end{align*}

\end{proof}

\subsection{Six Steps for Proving the Lipschitz Continuous Property of Hessian}\label{sec:lipschitz_app:steps}

\begin{definition}\label{def:further_decomposition}
    Let $\sigma(x) \in \R^{n \times n}$ be defined as in Definition~\ref{def:sigma_A_s}.
    For the further decomposition of hessian, we define the following terms
    \begin{itemize}
        \item $D_1(x) = 4\sigma(x)_{*,i}^{\circ 2} \cdot (\sigma(x)_{*,i}^{\circ 2})^\top$
        \item $D_2(x) =  -8\sigma(x)_{i,i} (\sigma(x)_{*,i} \circ \sigma(x)_{*,i})^\top$
        \item $D_3(x) = -8(\sigma(x)_{*,i} \circ \sigma(x)_{*,i}) \sigma(x)_{i,i}$
        \item $D_4(x) = 10\sigma(x)_{i,i}^2$
        \item $D_5(x) = 8\sigma(x)_{i,i}\diag( \sigma_{*,i}(x) ) \sigma(x) \diag( \sigma_{*,i}(x) )$
        \item $D_6(x) = -6 \sigma(x)_{i,i} \diag(\sigma(x)_{*,i} \circ \sigma(x)_{*,i})^\top$
    \end{itemize}
\end{definition}

\begin{lemma}[Formal version of Lemma~\ref{lem:lipschitz_informal}]\label{lem:lipschitz_formal}
    Let $D_i$ be defined in Definition~\ref{def:further_decomposition}
    Then we have
    \begin{align*}
        \|H(x) - H(\wh{x})\| \leq 812 \beta^{-9}R^5 \|x - \wh{x}\|_2
    \end{align*}
\end{lemma}
\begin{proof}
    \begin{align*}
        \|H(x) - H(\wh{x})\|
        = & ~ \|\sum_{i=1}^6 (A(x)^\top D_i(x) A(x)) -  (A(\wh{x})^\top D_i(\wh{x}) A(\wh{x}))\| \\
        \leq & ~ \sum_{i=1}^6 (2 \|A(x)^\top D_i(x) A(x)) -  A(x)^\top D_i(x) A(\wh{x})\| + \|A(x)^\top D_i(x) A(\wh{x})) -  A(x)^\top D_i(\wh{x}) A(\wh{x})\|) \\
        \leq & ~ \sum_{i=1}^6 (2 \|A(x)\| \|D_i(x)\| \|A(x) - A(\wh{x})\| + \|A(x)\|^2 \|D_i(x) - D_i(\wh{x})\| ) \\
        \leq & ~ 7 (2 \|A(x)\| \max_{1 \leq i \leq 7}\|D_i(x)\| \|A(x) - A(\wh{x})\| + \|A(x)\|^2 \max_{1 \leq i \leq 7} \|D_i(x) - D_i(\wh{x})\| ) \\
        \leq & ~ 7 (20 \beta^{-3} R^3 \|x - \wh{x}\|_2 + 96 \beta^{-9}R^5 \|x -\wh{x} \|_2) \\
        \leq & ~ 812 \beta^{-9}R^5 \|x -\wh{x} \|_2
    \end{align*}
    where the initial step is based on Fact~\ref{fac:hessian_decomposition_final},
the subsequent step arises from straightforward algebra,
the 3rd step follows from the triangle inequality,
the 4th step stems from basic algebra,
the 5th step is derived from the outcomes of Lemma~\ref{lem:d_1_lips}--Lemma~\ref{lem:d_7_lips} and {\bf Part 3} of Lemma~\ref{lem:basic_lips},
and the final step results from basic algebraic manipulation.

\end{proof}

\begin{lemma}\label{lem:d_1_lips}
    Let $D_1(x) = \sigma(x)_{*,i}^{\circ 2} \cdot (\sigma(x)_{*,i}^{\circ 2})^\top$
    Then we have
    \begin{align*}
        \|D_1(x) - D_1(\wh{x})\| \leq 48\beta^{-7}R^3  \cdot \|x - \wh{x}\|_2
    \end{align*}
\end{lemma}
\begin{proof}
\begin{align*}
    \|D_1(x) - D_1(\wh{x})\| 
    = & ~ 4\|\sigma(x)_{*,i}^{\circ 2}  (\sigma(x)_{*,i}^{\circ 2})^{\top} -  \sigma(\wh{x})_{*,i}^{\circ 2}  (\sigma(\wh{x})_{*,i}^{\circ 2})^{\top} \| \\
    = & ~ 4\| \diag(\sigma(x)_{*,i}) \sigma(x)_{*,i}\sigma(x)_{*,i}^{\top} \diag(\sigma(x)_{*,i})- \diag(\sigma(\wh{x})_{*,i}) \sigma(\wh{x})_{*,i} \sigma(\wh{x})_{*,i}^{\top} \diag(\sigma(\wh{x})_{*,i})\| \\
    \leq & ~ 4(\|\diag(\sigma(x)_{*,i}) - \diag(\sigma(\wh{x})_{*,i}) \|\|\sigma(x)_{*,i}\|_2^2 \| \diag(\sigma(x)_{*,i})\| \\
    & ~ + \|\diag(\sigma(\wh{x})_{*,i}) \| \| \sigma(x)_{*,i} - \sigma(\wh{x})_{*,i} \| \diag(\sigma(x)_{*,i})\|\sigma(x)_{*,i} \|_2\\
    & ~ + \|\diag(\sigma(\wh{x})_{*,i})\|\sigma(x)_{*,i}\|_2 \|\sigma(x)_{*,i}^{\top}-\sigma(\wh{x})_{*,i}^{\top}  \|_2\sigma(\wh{x})_{*,i}\|_2
    \|\|\diag(\sigma(x)_{*,i}) \|\\
    & ~ + \|\diag(\sigma(\wh{x})_{*,i}) \| \|\sigma(\wh{x})_{*,i}\|_2^2   \|\diag(\sigma(x)_{*,i}) - \diag(\sigma(\wh{x})_{*,i}) \|)\\
  \leq & ~ 48\beta^{-7}R^3 \|x -\wh{x} \|_2
\end{align*}
where the initial step arises from the definition of $D_1(x)$, the subsequent step stems from Fact~\ref{fac:circ_rule}, the third step follows from the triangle inequality, and the final step is based on Fact~\ref{fac:circ_rule}, {\bf Part 8} of Lemma~\ref{lem:basic_lips}, and {\bf Part 3} of Lemma~\ref{lem:norm_bounds}.
\end{proof}

\begin{lemma}\label{lem:d_2_lips}
    Let $D_2(x) = -8\sigma(x)_{i,i}(\sigma(x)^{\circ 2}_{*,i})^\top$
    Then we have
    \begin{align*}
        \|D_2(x) - D_2(\wh{x})\| \leq 72\beta^{-7}R^3  \cdot \|x - \wh{x}\|_2
    \end{align*}
\end{lemma}
\begin{proof}
   \begin{align*}
        \|D_2(x) - D_2(\wh{x})\| 
        = & ~ \|-8\sigma(x)_{i,i}(\sigma(x)^{\circ 2}_{*,i})^\top -(-8\sigma(\wh{x})_{i,i} (\sigma(\wh{x})^{\circ 2}_{*,i})^\top) \| \\
        = & ~ \| 8\sigma(x)_{i,i}(\sigma(x)_{*,i} \circ \sigma(x)_{*,i})^\top -8\sigma(\wh{x})_{i,i}^\top (\sigma(\wh{x})_{*,i} \circ \sigma(\wh{x})_{*,i})^\top\| \\
        \leq & ~ 8 |\sigma(x)_{i,i} - \sigma(\wh{x})_{i,i}|\|\sigma(x)_{*,i}^\top \|_2 \|\diag(\sigma(x)_{*,i}) \| \\
        & ~ + 8 |\sigma(\wh{x})_{i,i}|\|\sigma(x)_{*,i}^\top  - \sigma(\wh{x})_{*,i}^\top \|_2 \|\diag(\sigma(x)_{*,i}) \|  \\
        & ~ + 8 |\sigma(\wh{x})_{i,i}|\|\sigma(\wh{x})_{*,i}^\top \|_2 \|\diag(\sigma(x)_{*,i}) - \diag(\sigma(x)_{*,i})\| \\
        \leq & ~ 72\beta^{-7}R^3 \|x -\wh{x} \|_2
   \end{align*} 
   where the initial step arises from the definition of $D_2(x)$, the subsequent step stems from Fact~\ref{fac:matrix_norm}, the 3rd step follows from the triangle inequality, and the final step is based on Fact~\ref{fac:circ_rule}, along with {\bf Part 7,8} of Lemma~\ref{lem:basic_lips} and {\bf Part 2,3} of Lemma~\ref{lem:norm_bounds}.
\end{proof}

\begin{lemma}\label{lem:d_3_lips}
    Let $D_3(x) = -8(\sigma(x)_{*,i} \circ \sigma(x)_{*,i}) \sigma(x)_{i,i}$
    Then we have
    \begin{align*}
        \|D_3(x) - D_3(\wh{x})\| \leq 72\beta^{-7}R^3 \cdot \|x - \wh{x}\|_2
    \end{align*}
\end{lemma}
\begin{proof}
    Similar to Lemma~\ref{lem:d_2_lips}, we have
    \begin{align*}
        \|D_3(x) - D_3(\wh{x})\| 
        \leq & ~ 72\beta^{-7}R^3 \|x -\wh{x} \|_2
    \end{align*}
\end{proof}

\begin{lemma}\label{lem:d_4_lips}
    Let $D_4(x) = 10\sigma(x)_{i,i}^2$
    Then we have
    \begin{align*}
        \|D_4(x) - D_4(\wh{x})\| \leq \beta^{-7}R^3  \cdot \|x - \wh{x}\|_2
    \end{align*}
\end{lemma}
\begin{proof}
    \begin{align*}
       |D_4(x) - D_4(\wh{x})| \leq & ~ |10\sigma(x)_{i,i}^2 - 10\sigma(\wh{x})_{i,i}^2 | \\
       \leq & ~ 30 \beta^{-7}R^3 \|x -\wh{x} \|_2
    \end{align*}
where the initial step arises from the definition of $D_2(x)$ and the final step is based on {\bf Part 7} of Lemma~\ref{lem:basic_lips}.
\end{proof}

\begin{lemma}\label{lem:d_6_lips}
    Let $D_5(x) = 8\sigma(x)_{i,i}\diag( \sigma_{*,i}(x) ) \sigma(x) \diag( \sigma_{*,i}(x) )$
    Then we have
    \begin{align*}
        \|D_5(x) - D_5(\wh{x})\| \leq 96 \beta^{-7}R^3 \cdot \|x - \wh{x}\|_2
    \end{align*}
\end{lemma}
\begin{proof}
    \begin{align*}
         \|D_5(x) - D_5(\wh{x})\| 
         = & ~ \| 8\sigma(x)_{i,i}\diag( \sigma_{*,i}(x) ) \sigma(x) \diag( \sigma_{*,i}(x) ) -8\sigma(\wh{x})_{i,i}\diag( \sigma_{*,i}(\wh{x}) ) \sigma(\wh{x}) \diag( \sigma_{*,i}(\wh{x}) )\| 
         \\
         \leq & ~ 8 | \sigma(x)_{i,i} - \sigma(\wh{x})_{i,i}| \|\diag( \sigma_{*,i}(x) ) \| \|\sigma(x)\|\| \diag( \sigma_{*,i}(x) ) \|\\
         & ~ +8 | \sigma(\wh{x})_{i,i} | \|\diag( \sigma_{*,i}(x) )  - \diag( \sigma_{*,i}(\wh{x}) ) \| \|\sigma(x)\|\| \diag( \sigma_{*,i}(x) ) \|\\
         & ~ +8 | \sigma(\wh{x})_{i,i} | \| \diag( \sigma_{*,i}(\wh{x}) ) \| \|\sigma(x)- \sigma(\wh{x})\|\| \diag( \sigma_{*,i}(x) ) \|\\
         & ~ +8 | \sigma(\wh{x})_{i,i} | \| \diag( \sigma_{*,i}(\wh{x}) ) \| \|\sigma(\wh{x})\|\| \diag( \sigma_{*,i}(x) ) -\diag( \sigma_{*,i}(\wh{x}) )\|\\
         \leq & ~ 96 \beta^{-7}R^3 \|x -\wh{x} \|_2
    \end{align*}
    where the initial step arises from the definition of $D_6(x)$, the subsequent step follows from the triangle inequality, and the final step is based on Fact~\ref{fac:circ_rule}, along with {\bf Part 6,7,8} of Lemma~\ref{lem:basic_lips} and {\bf Part 1,2,3} of Lemma~\ref{lem:norm_bounds}.
\end{proof}

\begin{lemma}\label{lem:d_7_lips}
    Let $D_6(x) = -6 \sigma(x)_{i,i} \diag(\sigma(x)_{*,i} \circ \sigma(x)_{*,i})^\top$
    Then we have
    \begin{align*}
        \|D_6(x) - D_6(\wh{x})\| \leq 54 \beta^{-7}R^3 \cdot \|x -\wh{x} \|_2
    \end{align*}
\end{lemma}
\begin{proof}
    \begin{align*}
        \|D_6(x) - D_6(\wh{x})\|
        \leq & ~ \| -6 \sigma(x)_{i,i} \diag(\sigma(x)_{*,i} \circ \sigma(x)_{*,i})^\top - ( -6 \sigma(\wh{x})_{i,i} \diag(\sigma(\wh{x})_{*,i} \circ \sigma(\wh{x})_{*,i})^\top) \|\\
         = & ~ \| 6 \sigma(x)_{i,i} \diag(\sigma(x)_{*,i} \circ \sigma(x)_{*,i})^\top - 6 \sigma(\wh{x})_{i,i} \diag(\sigma(\wh{x})_{*,i} \circ \sigma(\wh{x})_{*,i})^\top \|\\
         \leq & ~ 6 | \sigma(x)_{i,i}- \sigma(\wh{x})_{i,i}| \|\diag(\sigma(x)_{*,i} )\| \|\diag(\sigma(x)_{*,i})\|\\
         & ~ + 6 |  \sigma(\wh{x})_{i,i}| \|\diag(\sigma(x)_{*,i} ) - \diag(\sigma(\wh{x})_{*,i} )\| \|\diag(\sigma(x)_{*,i})\|\\
         & ~ + 6 |  \sigma(\wh{x})_{i,i}| \|\diag(\sigma(\wh{x})_{*,i} )\| \|\diag(\sigma(x)_{*,i}) -\diag(\sigma(\wh{x})_{*,i})\|\\
         \leq & ~ 54 \beta^{-7}R^3 \|x -\wh{x} \|_2
    \end{align*}
    where the initial step arises from the definition of $D_7(x)$, the subsequent step follows from the triangle inequality, and the final step is based on Fact~\ref{fac:circ_rule}, along with {\bf Part 7,8} of Lemma~\ref{lem:basic_lips} and {\bf Part 2,3} of Lemma~\ref{lem:norm_bounds}.
\end{proof}

\begin{algorithm}[!ht]\caption{First Order Method}\label{alg:first_order}
\begin{algorithmic}[1]
\Procedure{FirstOrderMethod}{$A,b,c$}
    \State Let $x_0 \in \R^d$ denote an initialization point
    \For{$t = 1 \to T$}
        \State Let $g(x_{t-1}) =\frac{\d L_c(x)}{\d x} \big|_{x = x_{t-1}} $
        \State $x_{t} \gets x_{t-1} - \eta \cdot g(x_{t-1})$
    \EndFor
    \State \Return $L(x_T)$ 
\EndProcedure
\end{algorithmic}
\end{algorithm}

\begin{algorithm}[!ht]\caption{Second Order Method}\label{alg:second_order}
\begin{algorithmic}[1]
\Procedure{SecondOrderMethod}{$A,b,c$}
    \State Let $x_0 \in \R^d$ denote an initialization point
    \For{$t = 1 \to T$}
        \State Let $H(x_{t-1}) = \frac{\d^2 L_c(x)}{\d x \d x^\top } \big|_{x = x_{t-1}}$ denote the hessian
        \State Let $g(x_{t-1}) =\frac{\d L_c(x)}{\d x} \big|_{x = x_{t-1}} $
        \State $x_{t} \gets x_{t-1} - H(x_{t-1})^{-1} \cdot g(x_{t-1})$
    \EndFor
    \State \Return $x_T$
\EndProcedure
\end{algorithmic}
\end{algorithm}

\section{Computation}\label{sec:computation}

In this section, we analyzed the running time required for computing the gradient and hessian step by step. In Section~\ref{sec:computation:A}, we analyzed the running time required for computing $A(x)$. In Section~\ref{sec:computation:diag_leverage_score_matrix}, we analyzed the running time required for computing $\Sigma(x)$. In Section~\ref{sec:computation:leverage_score_matrix}, we analyzed the running time required for computing $\sigma(x)$. In Section~\ref{sec:computation:hessian}, we analyzed the running time required for computing $\nabla \sigma(x)_{i,i}$.  In Section~\ref{sec:computation:hessian}, we analyzed the running time required for computing $\nabla^2 L_{\exp}$.

\subsection{Computation of \texorpdfstring{$A(x)$}{}}\label{sec:computation:A}
\begin{lemma}\label{lem:compute_A_x}
If we have:
\begin{itemize}
    \item Define $A(x) \in \mathbb{R}^{n \times d}$ as described in Definition~\ref{def:Ax}.
    \item Define $S(x) = \text{diag}(s(x)) \in \mathbb{R}^{n \times n}$ as specified in Definition~\ref{def:s}.
    \item Consider $A \in \mathbb{R}^{n \times d}$.
\end{itemize}

Then, we can compute $A(x)$ in $O(nd)$ time.
\end{lemma}
\begin{proof}
Note that by Definition~\ref{def:s}, we have $S(x) = \diag( \underbrace{A}_{n \times d} \underbrace{x}_{d \times 1} - b )$. Then we can compute $S(x) \in \R^{n \times n}$ in $O(nd)$.

Then, since computing the inverse of a $n \times n$ diagonal matrix is taking the reciprocal of numbers on the diagonal, which costs $O(n)$ time, we can compute $S(x)^{-1}$ in $O(n)$ time.

Note that by Definition~\ref{def:Ax}, we have
\begin{align*}
    A(x) = \underbrace{S(x)^{-1}}_{n \times n} \underbrace{A}_{n \times d}.
\end{align*}
Note that $S(x)^{-1}$ is a diagonal matrix, so we can regard it as a vector. Therefore, we can compute $ S(x)^{-1} A $ in  $O(nd)$ time.

\end{proof}

\subsection{Computation of Diagonal of Leverage Score Matrix}\label{sec:computation:diag_leverage_score_matrix}

\begin{lemma}[Computing $\Sigma(x) \in \R^{n \times n}$]\label{lem:compute_Sigma_x}
If we have:
\begin{itemize}
    \item Define $A(x) \in \mathbb{R}^{n \times d}$ as specified in Definition~\ref{def:Ax}.
    \item Define $\Sigma(x) = I_n \circ \sigma(x) \in \mathbb{R}^{n \times n}$ as described in Definition~\ref{def:sigma}.
\end{itemize}
Then, we can compute $\Sigma(x)$ in $O( nd^2 + d^{\omega})$ time.
\end{lemma}
\begin{proof}
Recall that 
\begin{align*}
    \sigma(x)_{i,i} = a(x)_i^\top \underbrace{ (A(x)^\top A(x))^{-1} }_{d \times d} a(x)_i
\end{align*}
First, we compute $\underbrace{ A(x)^\top}_{d \times n} \underbrace{ A(x) }_{n \times d}$, this takes $\Tmat(d,n,d)$ time.

Second we can compute $( \underbrace{ A(x)^\top A(x) }_{d \times d} )^{-1}$, this takes $d^{\omega}$ time.

Third, we can compute $a(x)_i^\top \cdot \underbrace{ (A(x)^\top A(x))^{-1} }_{d \times d} a(x)_i^\top$, this takes $d^2$ time. Since we have $n$ different $i$s, so that overall, it takes $nd^2$.

The total time is
\begin{align*}
    O(\Tmat(d,n,d) + d^{\omega} + nd^2) = O(nd^2 + d^{\omega})
\end{align*}
\end{proof}

\subsection{Computation of Leverage Score Matrix}\label{sec:computation:leverage_score_matrix}

\begin{lemma}[Computing $\sigma(x) \in \R^{n \times n}$]\label{lem:compute_sigma_x}
If we have:
\begin{itemize}
    \item Define $A(x) \in \mathbb{R}^{n \times d}$ according to Definition~\ref{def:Ax}.
    \item Define $\sigma(x) \in \mathbb{R}^{n \times n}$ as described in Definition~\ref{def:sigma}.
\end{itemize}
Then, we can compute $\sigma(x)$ in $O( \Tmat(n,n,d) + \Tmat(d,d,d) )$ time.
\end{lemma}
\begin{proof}
Recall that 
\begin{align*}
    \sigma(x) = \underbrace{ A(x) }_{n \times d} \underbrace{ (A(x)^\top A(x))^{-1} }_{d \times d} \underbrace{ A(x)^\top }_{d \times n}
\end{align*}
First, we compute $\underbrace{ A(x)^\top}_{d \times n} \underbrace{ A(x) }_{n \times d}$, this takes $\Tmat(d,n,d)$ time.

Second we can compute $( \underbrace{ A(x)^\top A(x) }_{d \times d} )^{-1}$, this takes $\Tmat(d,d,d)$ time. 

Third, we can compute $\underbrace{ A(x) }_{n \times d} \cdot \underbrace{ (A(x)^\top A(x))^{-1} }_{d \times d}$, this takes $\Tmat(n,d,d)$ time.

Forth, we compute $\underbrace{ ( A(x) (A(x)^\top A(x))^{-1} ) }_{n \times d} \cdot \underbrace{ A(x)^\top }_{d \times n}$, this takes $\Tmat(n,d,n)$ time.

The overall time is
\begin{align*}
    & ~ \Tmat(n,n,d) + \Tmat(n,d,d) + \Tmat(d,d,d) \\
    = & ~ O( \Tmat(n,n,d) + \Tmat(d,d,d) )
\end{align*}
Where it follows from $\Tmat(n,d,d) \leq O( \Tmat(n,n,d) + \Tmat(d,d,d) )$(the proof is simple, we can just consider two cases, one is $n > d$, the other $d< n$).
\end{proof}

\subsection{Computation of Gradient}\label{sec:computation:gradient}
\begin{lemma}\label{lem:compute_gradient}
If we have:
\begin{itemize}
    \item Let $\frac{\d \sigma(x)_{i,l}}{\d x_j} \in \R^n$ be defined as Lemma~\ref{lem:gradient_leverage}.
\end{itemize}
Then, we can compute $\frac{ \d \sigma(x)_{i,i}  }{ \d x } $ in $O(nd)$ time.

\end{lemma}

\begin{proof}

By Lemma~\ref{lem:gradient_leverage}, we have 
\begin{align*}
    \frac{ \d \sigma(x)_{i,l}  }{ \d x_j } 
    = & ~ 2  \langle \sigma(x)_{*,i} \circ \sigma(x)_{*,l} , A(x)_{*,j} \rangle - \sigma(x)_{i,l} \cdot (A(x)_{i,j} + A(x)_{l,j} ).
\end{align*}
By choosing the $i=l$ in the above equation, we can get the following,
\begin{align*}
    \frac{ \d \sigma(x)_{i,i}  }{ \d x_j } = 2  \langle \underbrace{\sigma(x)_{*,i}}_{n \times 1} \circ \underbrace{\sigma(x)_{*,i}}_{n \times 1} , \underbrace{A(x)_{*,j}}_{n \times 1} \rangle - 2 \sigma(x)_{i,i} \cdot A(x)_{i,j} .
\end{align*}

Thus, we have
\begin{align*}
    \frac{ \d \sigma(x)_{i,i}  }{ \d x } 
    = & ~ \begin{bmatrix}
        \frac{ \d \sigma(x)_{i,i}  }{ \d x_1 } \\
        \frac{ \d \sigma(x)_{i,i}  }{ \d x_2 } \\
        \vdots \\
        \frac{ \d \sigma(x)_{i,i}  }{ \d x_d }
    \end{bmatrix}\\
    = & ~ 2 \underbrace{A(x)^\top}_{d \times n} \underbrace{\sigma(x)_{*,i}^{\circ 2}}_{n \times 1}  - 2 \sigma(x)_{i,i} \cdot \underbrace{A(x)_{i, *}}_{d \times 1}
\end{align*}

Computing $\sigma(x)_{*,i}^{\circ 2}$ takes $O(n)$ times.

Computing $A(x)^\top$ which is a $d \times n$ matrix multiplying with $\sigma(x)_{*,i}^{\circ 2}$ which is an $n$-dimensional vector takes $O(\Tmat(d, n, 1))$ time, which is equal to $O(nd)$.

Therefore, computing $2 A(x)^\top \sigma(x)_{*,i}^{\circ 2}  - 2 \sigma(x)_{i,i} \cdot A(x)_{i, *}$ takes $O(nd)$ time.
\end{proof}

\subsection{Computation of Hessian}\label{sec:computation:hessian}
\begin{lemma}\label{lem:compute_hessian}

    Let $H_i \in \R^{d \times d}$ be defined as Lemma~\ref{lem:hessian_decompose} 
        \begin{align*}
   H_i  = D_{i,1} + D_{i,2} + D_{i,3} + D_{i,4} + D_{i,5} + D_{i,6} 
\end{align*}
where 
\begin{itemize}
    \item $D_{i,1} = 4\underbrace{ A(x)^\top }_{d \times n} \underbrace{ \sigma(x)_{*,i}^{\circ 2} }_{n \times 1} \cdot \underbrace{ (\sigma(x)_{*,i}^{\circ 2})^\top }_{1 \times n} \underbrace{ A(x) }_{n \times d} $ 
    \item $D_{i,2} = -8 \underbrace{ \sigma(x)_{i,i} }_{ \mathrm{scalar} } \underbrace{ ( A(x)_{i,*} )^\top }_{ d \times 1 } \underbrace{ (\sigma(x)_{*,i}^{\circ 2})^\top }_{1 \times n} \underbrace{A(x)}_{n \times d}$
    \item $D_{i,3} = -8 \underbrace{ ( A(x)_{i,*} )^\top }_{d \times 1} \underbrace{ (\sigma(x)_{*,i}^{\circ 2})^\top}_{1 \times n} \sigma(x)_{i,i} \underbrace{ A(x) }_{n \times d}$ 
    \item $D_{i,4} = 10  \underbrace{ ( A(x)_{i,*} )^\top }_{d \times 1} \sigma(x)_{i,i}^2 \underbrace{A(x)_{i,*}}_{1 \times d}$ 
    \item $D_{i,5} = 8 \sigma(x)_{i,i} \underbrace{A(x)^\top}_{d \times n} \underbrace{\diag( \sigma_{*,i}(x) )}_{n \times n} \underbrace{\sigma(x)}_{n \times n} \underbrace{\diag( \sigma_{*,i}(x) )}_{n \times n} \underbrace{A(x)}_{n \times d}$
    \item $D_{i,6} = -6 \underbrace{A(x)^{\top}}_{d \times n} \sigma(x)_{i,i}  \underbrace{\diag( \sigma(x)_{*,i}^{\circ 2})}_{n \times n} \underbrace{A(x)}_{n \times d}$  
\end{itemize}

    Then, we can compute $H_i$ in $O( \Tmat(d, n, n) + \Tmat(d, n, d))$

\end{lemma}
\begin{proof}
{\bf Proof of $D_{i,1}$}

Suppose $\sigma(x) \in \R^{n \times n}$ is known, then we compute $\sigma_{*,i}^{\circ 2}(x) \in \R^n$ in $O(n)$ time.

Then we can compute $\underbrace{A(x)^\top}_{d \times n} \underbrace{\sigma(x)_{*,i}^{\circ 2}}_{n \times 1}$ in $\Tmat(d,n,1) = O(nd)$ time.

Finally, we can compute $A(x)^\top \sigma(x)_{*,i}^{\circ 2}$ multiply with the transpose of $A(x)^\top \sigma(x)_{*,i}^{\circ 2}$, which takes $\Tmat(d, 1, d) = O(d^2)$.

Thus, the total running time of this step is $O(nd + d^2)$

{\bf Proof of $D_{i,2}$}

Suppose $\sigma(x)$ is known, then we can construct $(\sigma(x)_{*,i}^{\circ 2})^\top$ in $O(n)$ time.

Then we can compute $\underbrace{(\sigma(x)_{*,i}^{\circ 2})^\top}_{1 \times n} \cdot \underbrace{A(x)}_{n \times d}$ in $\Tmat(1, n, d) = O(nd)$ time.

Finally, the computation of $\underbrace{A(x)_{i,*}^\top}_{d \times 1} \underbrace{(\sigma(x)_{*,i}^{\circ 2})^\top A(x)}_{1 \times d}$ in $O(d^2)$ time.

Thus, the total running time for this step is $O(nd + d^2)$

{\bf Proof of $D_{i,3}$}

Suppose $\sigma(x)$ is known, then we can construct $(\sigma(x)_{*,i}^{\circ 2})^\top$ in $O(n)$ time.

Then we can compute $\underbrace{(\sigma(x)_{*,i}^{\circ 2})^\top}_{1 \times n} \cdot \underbrace{A(x)}_{n \times d}$ in $\Tmat(1, n, d) = O(nd)$ time.

Finally, the computation of $\underbrace{A(x)_{i,*}^\top}_{d \times 1} \cdot  \underbrace{(\sigma(x)_{*,i}^{\circ 2})^\top A(x)}_{1 \times d}$ in $O(d^2)$ time.

Thus, the total running time for this step is $O(nd + d^2)$

{\bf Proof of $D_{i,4}$}

$D_4$ can be viewed as a multiplication of a $d \times 1$ vector and a $1 \times d$ vector, thus it's trivial that it takes $O(d^2)$ time

{\bf Proof of $D_{i,5}$}

First, compute $\underbrace{\diag(\sigma_{*,i}(x))}_{n \times n} \cdot  \underbrace{\sigma(x)}_{n \times n}$ takes $O(n^2)$ time due to the property of diagonal matrix.

Then, compute $\underbrace{\diag(\sigma_{*,i}(x))\sigma(x)}_{n \times n} \cdot  \underbrace{\diag(\sigma_{*,i}(x))}_{n \times n}$ takes $O(n^2)$ time due to the property of diagonal matrix.

After that, we can compute $\underbrace{A(x)^\top}_{d \times n} \cdot \underbrace{\diag(\sigma_{*,i}(x))\sigma(x)\diag(\sigma_{*,i}(x))}_{n \times n}$ in $\Tmat(d, n, n)$ time.

Finally, we can compute $\underbrace{A(x)^\top \diag(\sigma_{*,i}(x))\sigma(x)\diag(\sigma_{*,i}(x))}_{d \times n} \cdot \underbrace{A(x)}_{n \times d}$ in $\Tmat(d, n, d)$ time.

Thus, the total running time would be $O(n^2 + n^2 + \Tmat(d, n, n) + \Tmat(d, n, d)) = \Tmat(d, n, n) + \Tmat(d, n, d)$.

{\bf Proof of $D_{i,6}$}

First, computing $\underbrace{\diag( \sigma(x)_{*,i}^{\circ 2})}_{n \times n} \cdot  \underbrace{A(x)}_{n \times d}$ takes $O(nd)$ time due to the property of diagonal matrix.

Then, we can compute $\underbrace{A(x)^\top}_{d \times n} \cdot \underbrace{\diag( \sigma(x)_{*,i}^{\circ 2}) A(x)}_{n \times d}$ in $\Tmat(d, n, d)$ time.

In total, it takes $O(nd + \Tmat(d, n, d)) = \Tmat(d, n, d)$ times.

{\bf Combination}
Then the computation time for $H_{k,l}$ is 
\begin{align*}
    & ~ O(\underbrace{nd + d^2}_{D_{i,1}} + \underbrace{nd + d^2}_{D_{i,2}} + \underbrace{nd + d^2}_{D_{i,3}} + \underbrace{d^2}_{D_{i,4}} + \underbrace{\Tmat(d, n, n) + \Tmat(d, n, d)}_{D_{i,5}} + \underbrace{\Tmat(d, n, d)}_{D_{i,6}}) \\
    = & ~ O(nd + d^2 + \Tmat(d, n, n) + \Tmat(d, n, d)) \\
    = & ~ O( \Tmat(d, n, n) + \Tmat(d, n, d))
\end{align*} 
where the 1st step is due to Fact~\ref{fac:Tmat} and the 2nd step comes from Fact~\ref{fac:basic_Tmat}.
\end{proof}

\section{Main Theorems}\label{sec:main_result}

In this section, we present the main result of our paper in an formal way. Theorem~\ref{thm:gradient_formal} states that we are able to solve the inverting leverage score problem through a first order method. Theorem~\ref{thm:newton_formal} states that we are able to solve the inverting leverage score problem through a second order method. Both method are able to solve the problem with high accuracy and acceptable running time.

\begin{theorem}[Formal version of Theorem~\ref{thm:gradient_informal}]\label{thm:gradient_formal}

    If we have:
    \begin{itemize}
        \item Let $A \in \R^{n \times d}$.
        \item Let $b \in \R^n$.
        \item Let $\sigma \in \R^n$.
        \item Let $s(x) \in \R^n$.
        \item Let $S(x) := \diag(s(x)) \in \R^{n \times n}$.
        \item Let $A(x):= S(x)^{-1} A \in \R^{n \times d}$. 
        \item Let $\epsilon > 0$.
    \end{itemize}
    
Then, there exists an algorithm (see Algorithm~\ref{alg:first_order}), which uses the first order iteration and runs in $O( \epsilon^{-1} \poly(nd, \exp(R^2)) ) $ iterations and spends $O(n^2d + d^{\omega})$
    time per iteration and solves the inverting leverage score problem (defined in Definition~\ref{def:inverting_leverage_score}):
    \begin{align*}
        \min_{x \in \R^d } \| \diag( \sigma ) - I_n \circ (A(x) (A(x)^\top A(x) )^{-1} A(x)^\top )  \|_F.
    \end{align*}
    and finally outputs $x_T$ such that
    \begin{align*}
        L(x_T) - L(x^*) < \epsilon.
    \end{align*}
    with probability $1 - \poly(n)$.

\end{theorem}
\begin{proof}
    {\bf Proof of the number of iterations.}

    From Lemma~\ref{lem:gradient_convergence}, we know the iterations bound is $O(\epsilon^{-1} L^2/\alpha)$.

    Using Lemma~\ref{lem:lipschitz_formal} and Lemma~\ref{lem:convex}, we know that $L^2/\alpha = \poly(nd, \exp(R^2))$.

    {\bf Proof of gradient computation.}

    Using Lemma~\ref{lem:compute_A_x}, we can compute $A(x)$ in $O(nd)$ time.

    Using Lemma~\ref{lem:compute_sigma_x}, we can compute $\sigma(x)$ in $\Tmat(n,n,d) + d^{\omega}$ time

    Using Lemma~\ref{lem:compute_gradient}, for each $i \in [n]$, we can compute $\frac{\d \sigma_{i,i}}{\d  x}$ in $O(nd)$ time. 
    
    Since there are $n$ coordinates, thus, this step takes $n \cdot O(nd) = O(n^2 d)$ time.

    Therefore, each iteration of gradient step takes $O(n^2 d + d^{\omega })$ time

    {\bf Proof of Hessian Being Positive Definite.}

    This follows from Lemma~\ref{lem:convex}.

    {\bf Proof of Hessian Being Lipschitz.}

    This follows from Lemma~\ref{lem:lipschitz_formal}.

\end{proof}

\begin{theorem}[Formal version of Theorem~\ref{thm:newton_informal}]\label{thm:newton_formal}
    If we have:
    \begin{itemize}
        \item Let $A \in \R^{n \times d}$.
        \item Let $b \in \R^n$.
        \item Let $\sigma \in \R^n$.
        \item Let $s(x) \in \R^n$.
        \item Let $S(x) := \diag(s(x)) \in \R^{n \times n}$.
        \item Let $A(x):= S(x)^{-1} A \in \R^{n \times d}$. 
        \item Let $\epsilon > 0$.
    \end{itemize}
    
     Then, there exists an algorithm (see Algorithm~\ref{alg:second_order}), which uses the second order iteration and runs in $\log(r_0/\epsilon)$ iterations and spends 
    \begin{align*}
        n \cdot  O( \Tmat(d, n, n) + \Tmat(d, n, d)) + d^{\omega}
    \end{align*} 
    time per iteration and solves the inverting leverage score problem (defined in Definition~\ref{def:inverting_leverage_score}):
    \begin{align*}
        \min_{x \in \R^d } \| \diag( \sigma ) - I_n \circ (A(x) (A(x)^\top A(x) )^{-1} A(x)^\top )  \|_F.
    \end{align*}
    and finally outputs $x_T$ such that
    \begin{align*}
        \|x_T - x^*\|_2 < \epsilon
    \end{align*}
    with probability $1 - \poly(n)$.

\end{theorem}
\begin{proof}
    {\bf Proof of the number of iterations.}

    It follows from Lemma~\ref{lem:newton_induction}.

    {\bf Proof of gradient computation.}

    This is the same as Part 1, which is $O(n^2 d + d^{\omega})$

    {\bf Proof of Hessian computation.}
    
    The Hessian computation follows from Lemma~\ref{lem:hessian_decompose}.

    By Lemma~\ref{lem:compute_hessian}, we know that for each $i\in [n]$, the running time is
    \begin{align*}
        O( \Tmat(d, n, n) + \Tmat(d, n, d))
    \end{align*}
    Then for all $i$, the time is 
     \begin{align*}
        n \cdot \cdot O( \Tmat(d, n, n) + \Tmat(d, n, d)) + d^{\omega}
    \end{align*}
    We also need to compute the inverse of Hessian which takes $d^{\omega}$ time.

    Finally we need to do Hessian inverse multiply with gradient vector, this takes $O(d^2)$ time.

    Thus, overall the running time is
    \begin{align*}
       & ~ ( n^2d + d^{\omega} ) + n \cdot O( \Tmat(d, n, n) + \Tmat(d, n, d)) + d^{\omega} + d^2 \\
       = & ~ n \cdot  O( \Tmat(d, n, n) + \Tmat(d, n, d)) + d^{\omega}
    \end{align*}

    {\bf Proof of Hessian Being Positive Definite.}

    This follows from Lemma~\ref{lem:convex}.

    {\bf Proof of Hessian Being Lipschitz.}

    This follows from Lemma~\ref{lem:lipschitz_formal}.
\end{proof}


\ifdefined\isarxiv
\bibliographystyle{alpha}
\bibliography{ref}
\else

\fi



\end{document}